\documentclass[12pt]{article}
\usepackage{algorithm, algorithmic, amssymb, amsthm, bm, comment, mathtools, setspace, upquote}
\newtheorem{theorem}{Theorem}[section]
\newtheorem{lemma}[theorem]{Lemma}
\newtheorem{corollary}[theorem]{Corollary}
\DeclareMathOperator{\E}{\mathbb{E}}
\DeclareMathOperator{\Var}{\mathrm{Var}}

\usepackage[colorlinks=true, linkcolor=blue, citecolor=blue, urlcolor=blue]{hyperref}
\usepackage{longtable}
\usepackage{booktabs}
\usepackage{caption}
\counterwithin{table}{section}

\usepackage{amsmath}
{
	\theoremstyle{plain}
	\newtheorem{assumption}{Assumption}[section]
}

\newtheorem{manuallemmainner}{Lemma}
\newenvironment{manuallemma}[1]{%
  \renewcommand\themanuallemmainner{#1}%
  \manuallemmainner
}{\endmanuallemmainner}

\newtheorem{manualtheoreminner}{Theorem}
\newenvironment{manualtheorem}[1]{%
  \renewcommand\themanualtheoreminner{#1}%
  \manualtheoreminner
}{\endmanualtheoreminner}

\newtheorem{manualcorollaryinner}{Corollary}
\newenvironment{manualcorollary}[1]{%
  \renewcommand\themanualcorollaryinner{#1}%
  \manualcorollaryinner
}{\endmanualcorollaryinner}

\begin{document}

\baselineskip 16pt
\let\oldbibliography\thebibliography
\renewcommand{\thebibliography}[1]{
\oldbibliography{#1}
\setlength{\itemsep}{2pt}
\setlength{\baselineskip}{12pt}
\setlength{\lineskiplimit}{-\maxdimen}}

\begin{center}

\begin{spacing}{1.5}
\textbf{\Large Non-stationary and Varying-discounting Markov Decision Processes for Reinforcement Learning\footnote{Code: \url{https://github.com/zhizuo-chen/jupyter-notebook/tree/main/NVMDP}}}
\end{spacing}

\vspace{0.8cc}
{Zhizuo Chen$^{1}$, Theodore T. Allen$^{2}$}\\

\vspace{0.4cc}

{\footnotesize
$^{1}$Dept. of ISE, the Ohio State University, \texttt{chen.13619@osu.edu, zhizuo.chen@outlook.com} \\
$^{2}$Dept. of ISE, the Ohio State University, \texttt{allen.515@osu.edu}}
\end{center}


\begin{abstract}

\noindent Algorithms developed under stationary Markov Decision Processes (MDPs) often face challenges in non-stationary environments, and infinite-horizon formulations may not directly apply to finite-horizon tasks.
To address these limitations, we introduce the Non-stationary and Varying-discounting MDP (NVMDP) framework, which naturally accommodates non-stationarity and allows discount rates to vary with time and transitions.
Infinite-horizon, stationary MDPs emerge as special cases of NVMDPs for identifying an optimal policy, and finite-horizon MDPs are also subsumed within the NVMDP formulations.
Moreover, NVMDPs provide a flexible mechanism to shape optimal policies, without altering the state space, action space, or the reward structure.
We establish the theoretical foundations of NVMDPs, including assumptions, state- and action-value formulation and recursion, matrix representation, optimality conditions, and policy improvement under finite state and action spaces.
Building on these results, we adapt dynamic programming and generalized Q-learning algorithms to NVMDPs, along with formal convergence proofs.
For problems requiring function approximation, we extend the Policy Gradient Theorem and the policy improvement bound in Trust Region Policy Optimization (TRPO), offering proofs in both scalar and matrix forms.
Empirical evaluations in a non-stationary gridworld environment demonstrate that NVMDP-based algorithms successfully recover optimal trajectories under multiple reward and discounting schemes, whereas original Q-learning fails.
These results collectively show that NVMDPs provide a theoretically sound and practically effective framework for reinforcement learning, requiring only minor algorithmic modifications while enabling robust handling of non-stationarity and explicit optimal policy shaping.

\vspace{0.95cc}
\parbox{24cc}{{\it Key words}: Markov Decision Processes, non-stationarity, varying discounting, generalized Q-learning, TRPO}
\end{abstract}

\section{Introduction}\label{sec:intro}

According to Sutton et al. \cite{Sutton-RL-2018}, Reinforcement Learning (RL) is a computational framework in which a learner improves its performance through interactions with an environment. In the literature, the learner is commonly referred to as the \textit{agent}. The primary objective of RL is to learn a policy that maximizes the long-term cumulative reward obtained during the agent-environment interaction.

Grounded in the Markov Decision Process (MDP) framework, RL has achieved remarkable success across a wide range of domains, from strategic games to natural language processing.
Notable examples include AlphaGo, which combined deep neural networks with Monte Carlo Tree Search to defeat one of the strongest human Go players \cite{AlphaGo-2018}, and ChatGPT, which utilizes Reinforcement Learning from Human Feedback (RLHF) to align large language models with human preferences \cite{ChatGPT-2023}.

Generally, MDPs model sequential decision-making in discrete time.
At each time step $t \in \mathbb{Z}_{\ge 0}$, the agent observes a state $s_t$, selects an action $a_t$, receives a reward $r_t$, and transitions to a new state $s_{t+1}$.
The resulting sequence $\tau_t = (s_t, a_t, r_t, s_{t+1}, a_{t+1}, r_{t+1}, ...)$ is called a \textit{trajectory}, and a complete interaction from an initial state $s_0$ until termination is referred to as an \textit{episode}.
MDPs rely on the \textit{Markov property}, which requires that each state $s_t$ serves as a sufficient statistic for the future, containing all information necessary for current decision-making (i.e., action selection).
In other words, selecting an action $a_t$ based solely on $s_t$ is just as effective as relying on $s_t$ together with the entire preceding trajectory.

MDPs can be classified according to their state, action, and time structures.
When both the state space $S$ and the action space $A$ are finite, the model is a \textbf{Finite MDP (FMDP)}.
If the agent makes decisions over a limited number of time steps, the model is a \textbf{finite-horizon MDP}; otherwise, it is an \textbf{infinite-horizon MDP}.
These cases are typically analyzed differently in the literature.

Given a state–action pair $(s_t, a_t)$ at time $t$, the next state $s_{t+1}$ and reward $r_t$ are drawn from a transition–reward distribution $p(s_{t+1}, r_t | s_t, a_t)$, referred to as the \textit{dynamics} of the environment.
If this distribution remains invariant over time, the MDP is \textbf{stationary}; otherwise, it is \textbf{non-stationary}.
A policy $\pi(a | s)$ specifies the probability of selecting action $a$ in state $s$, which is typically assumed to be time-invariant under stationary MDPs.

In this work, we use infinite-horizon, non-stationary FMDPs as the ground setting.
We demonstrate that both stationary MDPs and finite-horizon MDPs can be treated as special cases within our framework presented later.

\subsection{Classic MDPs}

A substantial portion of RL research assumes an infinite-horizon, stationary FMDP with a fixed discount rate, which we refer to as a \textbf{classic MDP} in this work.
In a classic MDP, the return of a trajectory $\tau_t = (s_t, a_t, r_t, s_{t+1}, ...)$ starting from time $t$ is defined as
\begin{equation*}
G(\tau_t) = \sum_{i = t}^{\infty} \gamma^{i - t} r_i \ ,
\end{equation*}
where $\gamma \in [0, 1)$ is the discount rate, typically chosen by the experimenter to balance current and future rewards.

For a policy $\pi$, the state-value of state $s$ is
\begin{equation}\phantomsection\label{sv-mdp}
V^{\pi}(s) = \E^{\tau_t \sim \pi}[G(\tau_t) \mid s_t = s] \ ,
\end{equation}
and the action-value of selecting action $a$ in state $s$ is
\begin{equation}\phantomsection\label{av-mdp}
Q^{\pi}(s, a) = \E^{\tau_t \sim \pi}[G(\tau_t) \mid s_t = s, a_t = a] \ ,
\end{equation}
where $\E^{\tau_t \sim \pi}$ denotes that the expectation is over trajectories following $\pi$.
For brevity, we write $\E^\pi$ whenever unambiguous.
State-values and action-values are related by
\begin{equation*}
\begin{aligned}
V^{\pi}(s) &= \sum_{a} \pi(a | s) Q^{\pi}(s, a) \ , \\
Q^{\pi}(s, a) &= \sum_{s', r} p(s', r | s, a) \big(r + \gamma V^{\pi}(s')\big) \ ,
\end{aligned}
\end{equation*}
where $s, a, r$ denote the current state, action, and reward, and $s'$ is the next state.

A typical RL objective is to find a policy $\pi$ that maximizes $V^{\pi}(s_0)$ for a start state $s_0$ drawn from some distribution.
Many RL algorithms, designed under the framework of classic MDPs, aim to find a stationary (i.e., time-invariant) policy for this objective.
This approach relies on the stationarity assumption, which posits that if a policy $\pi$ is optimal, its optimality is time-consistent---for every future state (reachable under $\pi$), continuing to follow the same policy $\pi$ for action selection yields optimal state- and action-values.

In practice, methods based on the stationarity assumption may fail in non-stationary environments.
Experiments in Section \ref{sec:expt} illustrate such failures.
Another challenge arises in episodic RL tasks, where each episode must terminate in a finite number of steps.
When truncating the infinite-horizon of a classic MDP into a finite-horizon, the time consistency of optimal policies may no longer hold.
As the horizon approaches, optimality may require maximizing rewards within the few remaining steps, which can lead to a decision-making logic different from that of an infinite-horizon setting.
This issue is discussed in depth by Pardo et al. \cite{Time-Limit-RL-2018}.


\subsection{NSMDPs}

When the transition–reward distribution $p(s_{t + 1}, r_t | s_t, a_t)$ changes over time, the MDP becomes non-stationary.
The dynamics can then be written as $p_t(s_{t + 1}, r_t | s_t, a_t)$, where the subscript $t$ under $p$ indicates explicit time dependence.
Equivalently, one can define time-dependent transitions $p_t(s' | s, a)$ and rewards $r_t(s, a, s')$, both of which can be derived from $p_t(s_{t + 1}, r_t | s_t, a_t)$.
Such a model is referred to as a \textbf{NSMDP} (Non-stationary MDP) by Lecarpentier et al. \cite{NSMDP-wcapp-2019}, a term we adopt in this work.

For an NSMDP, the state-value and action-value functions in Equation \eqref{sv-mdp} and Equation \eqref{av-mdp} are generalized to
\begin{equation}\phantomsection\label{sv-av-nsmdp}
\begin{aligned}
V^{\pi}_t(s) &= \E^\pi[\sum_{i = t}^{\infty} \gamma^{i - t} r_i \mid s_t = s] \ , \\
Q^{\pi}_t(s, a) &= \E^\pi[\sum_{i = t}^{\infty} \gamma^{i - t} r_i \mid s_t = s, a_t = a] \ ,
\end{aligned}
\end{equation}
where the subscript $t$ in $V^{\pi}_t(s)$ and $Q^{\pi}_t(s, a)$ reflects the time-dependent dynamics.
In addition, the relationship between state-values and action-values becomes
\begin{equation}\phantomsection\label{sv-av-iter-nsmdp}
\begin{aligned}
V^{\pi}_t(s) &= \sum_{a} \pi_t(a | s) Q^{\pi}_t(s, a) \ , \\
Q^{\pi}_t(s, a) &= \sum_{s', r} p_t(s', r | s, a) (r + \gamma V^{\pi}_{t + 1}(s')) \ .
\end{aligned}
\end{equation}

A fundamental difference from classic MDPs is that a global policy $\pi$ in an NSMDP is a collection of \textbf{time-specific sub-policies}, i.e., $\pi = (\pi_0, \pi_1, ..., \pi_t, ...)$.
Time consistency is no longer assumed in NSMDPs, meaning that the optimal action for the same state may vary across different time steps.
For convenience, we omit the subscript of $\pi$ (and similarly in our subsequent NVMDP framework) whenever it is unambiguous.

The NSMDP framework relaxes the fundamental assumption of stationarity in classic MDPs, allowing for non-stationarity, while still maintaining a fixed discount rate $\gamma \in [0, 1)$.
If we interpret the rewards as cash flows from the agent's bank account and the discount rate analogously to its counterpart in finance, then the state- and action-values defined by Equation \eqref{sv-av-nsmdp} correspond to the net present values of the agent's investment under a specific policy $\pi$.
This interpretation naturally raises the following questions: should discount rates necessarily remain fixed across all time steps, and must they always be constrained to values not exceeding 1?

Insights from economics and finance can help address these questions.
For example, when calculating bond prices, time-dependent spot rates are often employed instead of a fixed discount rate (derived from the global yield) to discount future cash flows (Lyuu \cite{Lyuu-FinEng-2001}).
Equivalently, such spot rates correspond to time-dependent discount rates in Equation \eqref{sv-av-nsmdp}.
Moreover, Japan implemented negative interest rates for eight years starting in February 2016, which can be interpreted as having $\gamma > 1$ for some time steps if the agent is viewed as the aggregate of Japan's bank accounts and the starting time precedes February 2016.
These two cases suggest that the discounting-sum structure in Equation \eqref{sv-av-nsmdp} can be further generalized to allow time-varying discount rates, including values that exceed 1.


\subsection{Our Contributions}

In this work, we propose the NVMDP framework, which extends the NSMDP framework with varying discount rates.  
For readability, \textbf{all proofs of lemmas, theorems, and corollaries are provided in Appendix \ref{sec:apndx-proofs}}.  
Furthermore, we adopt commonly used symbols from the RL literature while avoiding excessive notational abuse.  
Consequently, we do not strictly follow the statistical convention where uppercase letters denote random variables and lowercase letters denote their realizations.  
Some minor notational overload is unavoidable; for instance, $A$ denotes the action space when used without decorations, and the advantage function when superscripts or subscripts are present.  
We trust that readers can infer the intended meaning from the context.  

Before presenting the formal NVMDP formulation, we summarize our contributions and indicate the corresponding sections for ease of reference.  
To the best of our knowledge, our contributions include:

\begin{itemize}
\setlength{\itemsep}{0pt}
\item[$\square$] We introduce the NVMDP framework, which extends NSMDPs with both time- and transition-based discount rates (Section \ref{sec:nvmdp-fda}).
To ensure the framework is well-defined, we present Assumptions \ref{asm:nonneg-gamma-nvmdp}, \ref{asm:rwd-absbnd-nvmdp}, \ref{asm:Gamma-sum-absbnd-nvmdp}, and \ref{asm:gamma-bellman-cond-nvmdp}, which allow some discount rates to exceed 1.
We formalize the dynamic programming recursion (Lemma \ref{thm:sv-av-nvmdp}) and establish a value-consistency guarantee (Lemma \ref{thm:val-cns-nvmdp}).

\item[$\square$] We extend the matrix formulations of classic MDPs by Agarwal et al. \cite{RLTA-2022} to NVMDPs, making $\bm{V}_t^\pi$ and $\bm{Q}_t^\pi$ explicitly time-dependent and introducing matrices $\bm{\Pi}^\pi_t$ and $\bm{U}$ for policy operations, as well as $\bm{J}_t$, $\bm{K}_t$, and $\bm{L}_t$ for discounting.
The matrix representation is detailed in Section \ref{sec:nvmdp-mat-rep}.

\item[$\square$] Analogous to classic MDPs, we formalize the existence of a deterministic optimal policy in every NVMDP (Theorem \ref{thm:optim-pol-nvmdp}) and extend the Bellman optimality equation to the NVMDP setting (Theorem \ref{thm:bellman-nvmdp}).
Additionally, we adapt the reward shaping technique proposed by Ng et al. \cite{Rwd-Shaping-1999} to NVMDPs.
These results are discussed in Section \ref{sec:nvmdp-opt}, while the policy improvement theorem (Theorem \ref{thm:pol-imprv-nvmdp}) is adapted in Section \ref{sec:pol-imprv}.

\item[$\square$] We prove that classic MDPs are special cases of the NVMDP framework when the objective is to find an optimal policy, and demonstrate that finite-horizon MDPs are subsumed by NVMDPs, with discount rate requirements relaxed to merely non-negativity (Section \ref{sec:spcl-cases-nvmdp}).

\item[$\square$] We discuss time treatment in tabular solutions for NVMDPs (Section \ref{sec:time-trmt}) and extend two dynamic programming algorithms---policy evaluation and value iteration--- to NVMDPs along with convergence guarantees (Section \ref{sec:dp-nvmdp}).

\item[$\square$] We adapt the generalized Q-learning framework for classic MDPs by Lan et al. \cite{Maxmin-q-learn-2020} to NVMDPs, building upon extension of Tsitsiklis \cite{stoch-approx-ql-1994} (Lemma \ref{thm:stoapp-xk-absbnd} and Lemma \ref{thm:stoapp-xk-convg}).
Convergence is guaranteed by Theorem \ref{thm:geq-learn-nvmdp-convg}, which covers both the extended Q-learning in Section \ref{sec:q-learn} and four additional variants discussed in Section \ref{sec:geq-learn}.

\item[$\square$] For methods employing function approximation, we extend key theoretical results from classic MDPs to NVMDPs.
This includes the policy gradient theorem (Theorem \ref{thm:pol-grad-nvmdp}), the performance difference lemma (Lemma \ref{thm:prfm-diff-lemma-nvmdp}), and the policy improvement bound in TRPO (Theorem \ref{thm:prfm-diff-absbnd}), all discussed in Section \ref{sec:func-approx}.
For each result, we provide two proofs: one in the traditional scalar form and another using our matrix representation.

\item[$\square$] We introduce Tricky Gridworld, a gridworld environment with non-stationary dynamics (Section \ref{sec:expt}).
We demonstrate that original Q-learning fails to learn optimal policies in this environment, whereas NVMDP-based algorithms succeed.
Furthermore, we demonstrate that different discount rate settings can shape the optimal trajectories learned by the algorithms.
Detailed empirical results are provided in Appendix \ref{sec:apndx-long-table}.
\end{itemize}


\section{The NVMDP framework}\label{sec:nvmdp}

The NVMDP framework essentially integrates time- and transition-based discount rates into the NSMDP framework.
In the literature, both NSMDPs and varying discounting are widely studied.
As early as the 1970s, Hinderer \cite{Hinderer-1970} establishes dynamic programming under non-stationary conditions.
More recent research on non-stationarity includes Pardo et al. \cite{Time-Limit-RL-2018} in 2018 and Lecarpentier et al. \cite{NSMDP-wcapp-2019} in 2019.
The former proposes an implicit non-stationary treatment by replacing the state $s_t$ with $(t, s_t)$ in finite-horizon tasks, which improves the performance of existing RL algorithms in their empirical evaluations.
The latter models MDPs with evolving dynamics and plans using a risk-averse approach, and we adopt the term NSMDP from their work.

On the other hand, both time-dependent and transition-based discount rates are also widely discussed.
In 2017, White \cite{Trans-Based-Gamma-2017} introduces transition-based discount rates $\gamma(s, a, s')$ in RL, enabling adjustments for scenarios where the importance of future rewards varies with the environment.
Also in 2017, Ilhuicatzi-Roldán et al. \cite{MDP-tvd-rhz-2017} introduce dynamic programming with time-varying, state- and action-dependent discount rates $\gamma_t(s_t, a_t)$ in MDPs with random horizons.
Moreover, in 2023 we \cite{CFBF-2023} demonstrate that the Proximal Policy Optimization (PPO) algorithm \cite{PPO-2017}---a variant of the Trust Region Policy Optimization (TRPO) algorithm \cite{TRPO-2015}---is compatible with time-varying discount rates in portfolio optimization problems.

\subsection{Fundamentals}\label{sec:nvmdp-fda}

Consider both time- and transition-based discount rates $\gamma_{t + 1}(s_t, a_t, s_{t + 1})$ in an infinite-horizon, non-stationary FMDP that satisfy:
\begin{assumption}\label{asm:nonneg-gamma-nvmdp}
$\ \gamma_0 = 1, \ \gamma_{t + 1}(s_t, a_t, s_{t + 1}) \ge 0 \ (\forall t \ge 0) \ .$
\end{assumption}
\noindent Here, the subscript $t + 1$ in $\gamma_{t + 1}(s_t, a_t, s_{t + 1})$ indicates that the rate discounts the rewards received at time $t + 1$ and thereafter back to time $t$.
The terms $s_t, a_t, s_{t + 1}$ inside the parentheses of $\gamma_{t + 1}(s_t, a_t, s_{t + 1})$ specify that the rate is associated with the transition $(s_t, a_t) \rightarrow s_{t + 1}$.
An infinite-horizon, non-stationary FMDP with such discounting structures is referred to as a \textbf{Non-stationary and Varying-discounting Markov Decision Process (NVMDP)} in this work.

For an NVMDP, the discounted return of the trajectory $\tau_t = (s_t, a_t, r_t, s_{t + 1}, ...)$ at time step $t$ is:
\begin{equation}\phantomsection\label{return-nvmdp}
G_t(\tau_t) = \sum_{i = t}^{\infty}\Gamma^\tau_{t, i}r_i(s_i, a_i, s_{i + 1}) \ ,
\end{equation}
where $\Gamma^\tau_{t, i}$ is the product of successive discount rates:
\begin{equation}\phantomsection\label{Gamma-def}
\Gamma^\tau_{t, i} = \prod_{j = t + 1}^{i}\gamma_j(s_{j - 1}, a_{j - 1}, s_j) \ .
\end{equation}
($\prod_{j = t + 1}^{t}(\cdot)$ is defined as 1 throughout this work, therefore $\Gamma^\tau_{t, t} \equiv 1$.)
The state-values and action-values in NVMDPs generalize to:
\begin{equation}\phantomsection\label{sv-nvmdp}
V^\pi_t(s) = \E^\pi[G_t(\tau_t) \mid s_t = s] \ ,
\end{equation}
\begin{equation}\phantomsection\label{av-nvmdp}
Q^\pi_t(s, a) = \E^\pi[G_t(\tau_t) \mid s_t = s, a_t = a] \ .
\end{equation}
Importantly, the policy $\pi$ in an NVMDP is also a collection of time-specific sub-policies, as NVMDPs are more general forms of NSMDPs.

Another commonly used measure is the advantage function, which quantifies how much better or worse an action is compared to the expected return at one state.
In NVMDPs, the advantage function is defined as:
\begin{equation}\phantomsection\label{adv-nvmdp}
A^\pi_t(s, a) = Q^\pi_t(s, a) - V^\pi_t(s) \ ,
\end{equation}
which is mainly used in Section \ref{sec:func-approx}.

For an infinite-horizon MDP, a primary concern is the convergence of state and action values.
In the classic MDP setting, convergence is typically ensured by assuming that rewards are uniformly bounded in absolute value by some constant $R_B$ and that the fixed discount rate satisfies $\gamma < 1$.
Under these assumptions, all state- and action-values are bounded in absolute value by $R_B / (1 - \gamma)$, since
\begin{equation*}
\big|G(\tau_t)\big| = \big|\sum_{i = t}^{\infty} \gamma^{i - t} r_i\big| \le R_B \big|\sum_{i = t}^{\infty} \gamma^{i - t}\big| = {R_B \over {1 - \gamma}} \ .
\end{equation*}

In NVMDPs, the assumption of bounded rewards is adapted to:
\begin{assumption}\label{asm:rwd-absbnd-nvmdp}
There exists $R_B > 0$ such that
\begin{equation*}
|r_t(s, a, s')| \le R_B \quad(\forall t \ge 0, s, s' \in S, a \in A) \ .
\end{equation*}
\end{assumption}
\noindent And the requirement that the accumulated discounting sum be finite is generalized to:
\begin{assumption}\label{asm:Gamma-sum-absbnd-nvmdp}
There exists $\Gamma_B > 0$ such that for any trajectory starting from any time step $t$,
\begin{equation*}
\sum^\infty_{i = t} \Gamma^\tau_{t, i} \le \Gamma_B \ .
\end{equation*}
\end{assumption}

By Assumption \ref{asm:rwd-absbnd-nvmdp} and Assumption \ref{asm:Gamma-sum-absbnd-nvmdp}, all state- and action-values are bounded in absolute value by
\begin{equation}\phantomsection\label{sv-av-bnd}
\big(\sum^\infty_{i = t} \Gamma^\tau_{t, i} \big) R_B \le \Gamma_B R_B = V_B \ .
\end{equation}
While classic MDPs and NSMDPs require a fixed discount rate $\gamma < 1$ to guarantee the convergence of state- and action-values, the NVMDP framework permits some discount rates greater than $1$, provided that Assumption \ref{asm:Gamma-sum-absbnd-nvmdp} is satisfied.
This additional flexibility in setting discount rates can be leveraged to shape the trajectories of the optimal policies, as demonstrated in Section \ref{sec:expt}.

With convergence ensured, the following lemma establishes the recursive relationship between the state- and action-values in NVMDPs, analogous to Equation \eqref{sv-av-iter-nsmdp} in NSMDPs.
\begin{lemma}\label{thm:sv-av-nvmdp}
State-values and action-values in an NVMDP satisfy:
\begin{equation}\phantomsection\label{sv-av-nvmdp}
V^{\pi}_t(s_t) = \sum_{a_t} \pi_t(a_t | s_t) Q^{\pi}_t(s_t, a_t) \ ,
\end{equation}
\begin{equation}\phantomsection\label{av-iter-nvmdp}
Q^{\pi}_t(s_t, a_t) = \sum_{s_{t + 1}, r_t} p_t(s_{t + 1}, r_t | s_t, a_t) (r_t(s_t, a_t, s_{t + 1}) + \gamma_{t + 1}(s_t, a_t, s_{t + 1}) V^{\pi}_{t + 1}(s_{t + 1})) \ .
\end{equation}
\end{lemma}

By Lemma \ref{thm:sv-av-nvmdp},
\begin{equation*}
\begin{aligned}
Q^\pi_t(s, a) &= \E^{s', r_t \sim p_t(\cdot | s, a)}[r_t(s, a, s') + \gamma_{t + 1}(s, a, s') V^{\pi}_{t + 1}(s')] \\
&= \E^{s' \sim p_t(\cdot | s, a)}[r_t(s, a, s')] + \E^{s' \sim p_t(\cdot | s, a)}[\gamma_{t + 1}(s, a, s') V^{\pi}_{t + 1}(s')] \ .
\end{aligned}
\end{equation*}
With the average reward for selecting action $a$ in state $s$ at time $t$ denoted by:
\begin{equation}\phantomsection\label{avg-rwd-nvmdp}
\bar{r}_t(s, a) = \E^{s' \sim p_t(\cdot | s, a)}[r_t(s, a, s')] \ ,
\end{equation}
Equation \eqref{av-iter-nvmdp} can be rewritten as:
\begin{equation}\phantomsection\label{av-iter-nvmdp-e}
Q^{\pi}_t(s, a) = \bar{r}_t(s, a) + \E^{s' \sim p_t(\cdot | s, a)}[\gamma_{t + 1}(s, a, s') V^{\pi}_{t + 1}(s')] \ .
\end{equation}
Moreover, the definition of state- and action-values in Equations \eqref{sv-nvmdp} and \eqref{av-nvmdp} admit alternative forms, which are formalized below.
\begin{lemma}\label{thm:sv-av-avgrwd-nvmdp}
With $\bar{r}_t(s, a)$ defined by Equation \eqref{avg-rwd-nvmdp},
\begin{equation*}
\begin{aligned}
V^\pi_t(s) &= \E^\pi [\sum_{i = t}^{\infty}\Gamma^\tau_{t, i} \bar{r}_i(s_i, a_i) \mid s_t = s] \\
Q^\pi_t(s, a) &= \E^\pi [\sum_{i = t}^{\infty}\Gamma^\tau_{t, i} \bar{r}_i(s_i, a_i) \mid s_t = s, a_t = a] \\
\end{aligned}
\end{equation*}
\end{lemma}

Under Assumptions \ref{asm:nonneg-gamma-nvmdp}, \ref{asm:rwd-absbnd-nvmdp}, and \ref{asm:Gamma-sum-absbnd-nvmdp}, the state- and action-values are guaranteed to converge and remain bounded in absolute value.
An interesting question then arises: if two policies are identical from time step $n$ onward, are their corresponding state- and action-values also guaranteed to be identical for all $t \ge n$?
With a further assumption stated below:
\begin{assumption}\label{asm:gamma-bellman-cond-nvmdp}
For any time step $t$,
\begin{equation*}
\lim_{m \rightarrow \infty} \prod^m_{j = t + 1} \max_{s, a, s'} \gamma_j(s, a, s') = 0 \ .
\end{equation*}
\end{assumption}
\noindent the following lemma provides an affirmative answer:
\begin{lemma}\label{thm:val-cns-nvmdp}
Policy $\pi$ and $\pi'$ are identical for all time $t \ge n$, then
\begin{equation*}
\begin{aligned}
V^{\pi'}_t(s) &= V^\pi_t(s) \quad(\forall t \ge n, s \in S) \\
Q^{\pi'}_t(s, a) &= Q^\pi_t(s, a) \quad(\forall t \ge n - 1, s \in S, a \in A)
\end{aligned}
\end{equation*}
\end{lemma}

Lemma \ref{thm:val-cns-nvmdp} expresses a form of value-consistency: if two policies behave identically from a certain time step onward, their future value assessments also coincide.
In other words, the evaluation of expected future returns is determined solely by the sub-policies from that point onward.

The remainder of this work builds upon Assumption \ref{asm:nonneg-gamma-nvmdp}, \ref{asm:rwd-absbnd-nvmdp}, \ref{asm:Gamma-sum-absbnd-nvmdp}, and \ref{asm:gamma-bellman-cond-nvmdp}.


\subsection{Matrix Representations}\label{sec:nvmdp-mat-rep}

In the study of Markov chains, matrix representation provides a convenient framework for analyzing state transitions and streamline many proofs.
Since MDPs extend Markov chains by incorporating decision-making through action selection, the RL community has similarly adopted matrix-based formulations to achieve more compact expressions and facilitate theoretical analysis.

In this section, we adapt the matrix formulations of classic MDPs from Agarwal et al. \cite{RLTA-2022} to the NVMDP setting.
Bold symbols denote vectors or matrices to distinguish them from scalars, and we add time subscripts to value functions, making $\bm{V}_t^\pi$ and $\bm{Q}_t^\pi$ explicitly time-dependent.
We also introduce matrices $\bm{\Pi}^\pi_t$ and $\bm{U}$ for policy operations, and $\bm{J}_t$, $\bm{K}_t$, and $\bm{L}_t$ for discounting.

Consider the state space $S$ and action space $A$.
$\bm{V}^{\pi}_t$ denotes the vector of all state-values $V^{\pi}_t(s)$ at time $t$, and $\bm{Q}^{\pi}_t$ denotes the vector of all action-values $Q^{\pi}_t(s, a)$ at time $t$.
Their lengths are $|S|$ and $|S \times A|$, respectively.
The vector $\bm{r}_t$ contains the average rewards $\bar{r}_t(s, a)$ defined by Equation \eqref{avg-rwd-nvmdp}, with length $|S \times A|$.
For the state transition $p_t(s' | s, a) = \sum_r p_t(s', r | s, a)$ between times $t$ and $t + 1$, the corresponding transition matrix $\bm{P}_t \in \mathbb{R}^{|S \times A| \times |S|}$ is defined as
\begin{equation*}
(\bm{P}_t)_{(s, a), s'} = p_t(s' | s, a) \ .
\end{equation*}
The matrix of sub-policy $\pi_t$ is $\bm{\Pi}^\pi_t \in \mathbb{R}^{|S| \times |S \times A|}$ that satisfies
\begin{equation}\phantomsection\label{pi-def-nvmdp-mat}
\big(\bm{\Pi}^\pi_t\big)_{s, (\bar{s}, \bar{a})} =
\begin{cases}
0 & \ s \ne \bar{s} \\
\pi_t(\bar{a} | s) & \ s = \bar{s}
\end{cases} \ .
\end{equation}
In the above matrix formulations, the ordering of state–action pairs in $\bm{Q}^{\pi}_t$, $\bm{r}_t$, the rows of $\bm{P}_t$, and the columns of $\bm{\Pi}^\pi_t$ is consistent.
Similarly, the ordering of states in $\bm{V}^{\pi}_t$, the columns of $\bm{P}_t$, and the rows of $\bm{\Pi}^\pi_t$ is the same.

Under the matrix representation, Equation \eqref{sv-av-nvmdp} becomes:
\begin{equation}\phantomsection\label{sv-av-nvmdp-mat}
\bm{V}^{\pi}_t = \bm{\Pi}^\pi_t \bm{Q}^{\pi}_t \ .
\end{equation}
Equation \eqref{av-iter-nvmdp} and Equation \eqref{av-iter-nvmdp-e} become:
\begin{equation}\phantomsection\label{av-iter-nvmdp-mat}
\bm{Q}^{\pi}_t = \bm{r}_t + \bm{J}_{t + 1} \bm{V}^{\pi}_{t + 1} \ ,
\end{equation}
where $\bm{J}_{t + 1}$ is a $|S \times A| \times |S|$ matrix that:
\begin{equation}\phantomsection\label{j-def-nvmdp-mat}
\bm{J}_{t + 1} = \bm{W}_{t + 1} \odot \bm{P}_t \ .
\end{equation}
In the above equation, 
\begin{equation*}
(\bm{W}_{t + 1})_{(s, a), s'} = \gamma_{t + 1}(s, a, s') \ ,
\end{equation*}
``$\odot$'' is the element-wise multiplier for matrices.

By multiplying $\bm{\Pi}^\pi_t$ to both sides of Equation \eqref{av-iter-nvmdp-mat},
\begin{equation}\phantomsection\label{sv-selfiter-nvmdp-mat}
\bm{V}^{\pi}_t = \bm{\Pi}^\pi_t \bm{r}_t + \bm{L}_{t + 1}^\pi \bm{V}^{\pi}_{t + 1} \ ,
\end{equation}
where the L.H.S is by Equation \eqref{sv-av-nvmdp-mat} and in the R.H.S,
\begin{equation}\phantomsection\label{l-def-nvmdp-mat}
\bm{L}_{t + 1}^\pi = \bm{\Pi}^\pi_t \bm{J}_{t + 1} \ .
\end{equation}
Recursively applying Equation \eqref{sv-selfiter-nvmdp-mat} results in:
\begin{equation}\phantomsection\label{sv-recurs-nvmdp-mat}
\begin{aligned}
\bm{V}^{\pi}_t &= \Big(\sum_{i = t}^{m - 1} \big(\prod_{j = t + 1}^i \bm{L}_j^\pi\big) {\bm{\Pi}^\pi_i \bm{r}_i}\Big) + \Big(\big(\prod_{j = t + 1}^m \bm{L}_j^\pi \big) \bm{V}_m^\pi \Big) \quad(\forall m \ge t + 1) \\
&= \sum_{i = t}^\infty \big(\prod_{j = t + 1}^i \bm{L}_j^\pi\big) {\bm{\Pi}^\pi_i \bm{r}_i} \ ,
\end{aligned}
\end{equation}
where the $\big(\prod_{j = t + 1}^m \bm{L}_j^\pi \big) \bm{V}_m^\pi$ term in the R.H.S of the first line disappears in the second line as $m \rightarrow \infty$ by Assumption \ref{asm:gamma-bellman-cond-nvmdp}.

Equation \eqref{av-iter-nvmdp-mat} can also be rewritten as:
\begin{equation}\phantomsection\label{av-selfiter-nvmdp-mat}
\bm{Q}^{\pi}_t = \bm{r}_t + \bm{K}^\pi_{t + 1} \bm{Q}^{\pi}_{t + 1} \ ,
\end{equation}
where $\bm{K}^\pi_{t + 1}$ is a $|S \times A| \times |S \times A|$ matrix that:
\begin{equation}\phantomsection\label{k-def-nvmdp-mat}
\bm{K}^\pi_{t + 1} = \bm{M}_{t + 1} \odot \bm{P}^\pi_t \ .
\end{equation}
Here,
\begin{equation*}
(\bm{M}_{t + 1})_{(s, a), (s', a')} = \gamma_{t + 1}(s, a, s') \ ,
\end{equation*}
and
\begin{equation*}
\bm{P}^\pi_t = \bm{P}_t \bm{\Pi}^\pi_{t + 1} \ .
\end{equation*}
Apply Equation \eqref {av-selfiter-nvmdp-mat} recursively:
\begin{equation}\phantomsection\label{av-recurs-nvmdp-mat}
\begin{aligned}
\bm{Q}^{\pi}_t &= \Big(\sum_{i = t}^{m - 1} \big(\prod_{j = t + 1}^i \bm{K}_j^\pi\big) {\bm{r}_i}\Big) + \Big(\big(\prod_{j = t + 1}^m \bm{K}_j^\pi \big) \bm{Q}_m^\pi \Big) \quad(\forall m \ge t + 1) \\
&= \sum_{i = t}^\infty \big(\prod_{j = t + 1}^i \bm{K}_j^\pi\big) {\bm{r}_i} \ ,
\end{aligned}
\end{equation}
where the $\big(\prod_{j = t + 1}^m \bm{K}_j^\pi \big) \bm{Q}_m^\pi$ term in the R.H.S of the first line disappears in the second line as $m \rightarrow \infty$ by Assumption \ref{asm:gamma-bellman-cond-nvmdp}, similar to Equation \eqref{sv-recurs-nvmdp-mat}.

The $\bm{J}_{t + 1}$ matrix in \eqref{j-def-nvmdp-mat} is related to the $\bm{K}_{t + 1}$ matrix in \eqref{k-def-nvmdp-mat} by:
\begin{equation}\phantomsection\label{jk-relp-nvmdp-mat}
\bm{J}_{t + 1} \bm{\Pi}^\pi_{t + 1} = \bm{K}^\pi_{t + 1}\ ,
\end{equation}
since
\begin{equation*}
\begin{aligned}
(\bm{J}_{t + 1} \bm{\Pi}^\pi_{t + 1})_{(s, a), (s', a')} &= \sum_{\bar{s}} (\bm{J}_{t + 1})_{(s, a), \bar{s}} (\bm{\Pi}^\pi_{t + 1})_{\bar{s}, (s', a')} \\
&= \sum_{\bar{s}} \gamma_{t + 1}(s, a, \bar{s}) p_t(\bar{s} | s, a) (\bm{\Pi}^\pi_{t + 1})_{\bar{s}, (s', a')} \\
&= \gamma_{t + 1}(s, a, s') p_t(s'| s, a) \pi_{t + 1}(a' | s') \\
&= (\bm{K}^\pi_{t + 1})_{(s, a), (s', a')} \ .
\end{aligned}
\end{equation*}

Finally, we introduce a $|S \times A| \times |S|$ matrix $\bm{U}$:
\begin{equation}\phantomsection\label{u-def-nvmdp-mat}
\big(\bm{U})_{(\bar{s}, \bar{a}), s} =
\begin{cases}
0 &\ s \ne \bar{s} \\
1 &\ s = \bar{s}
\end{cases} \ .
\end{equation}
With matrix $\bm{U}$, Equation \eqref{adv-nvmdp} (which defines the advantage function) is transformed into vector formats as follows:
\begin{equation}\phantomsection\label{adv-nvmdp-mat}
\bm{A}_t^\pi = \bm{Q}_t^\pi - \bm{U} \bm{V}_t^\pi \ .
\end{equation}

For any policy matrix $\bm{\Pi}^\pi_t$, it satisifies
\begin{equation}\phantomsection\label{piu-relp-nvmdp-mat}
\bm{\Pi}^\pi_t \bm{U} = \bm{I} \ ,
\end{equation}
since
\begin{equation*}
\begin{aligned}
(\bm{\Pi}^\pi_t \bm{U})_{s, \bar{s}} &= \sum_{(\widetilde s, \widetilde a)} \big(\bm{\Pi}^\pi_t\big)_{s, (\widetilde s, \widetilde a)} \big(\bm{U}\big)_{(\widetilde s, \widetilde a), \bar s} \\
&= \begin{cases}
0 &\ s \ne \bar{s} \\
\sum_{\widetilde a} \pi_t(\widetilde a | s) = 1 &\ s = \bar{s}
\end{cases} \ .
\end{aligned}
\end{equation*}
However, matrix $\bm{U}$ is not the inverse matrix of $\bm{\Pi}^\pi_t$ since it is not a square matrix.


\subsection{Optimality in NVMDPs}\label{sec:nvmdp-opt}

A cornerstone of classic MDP theory is the existence of globally optimal state- and action-values, which can be achieved by a deterministic policy.
This theoretical result has been extensively discussed in the literature, including by Puterman \cite{Puterman-MDP-2014} and Agarwal et al. \cite{RLTA-2022}.
NVMDPs exhibit an analogous property, and the following theorem formalizes this claim.
Notably, when the time subscripts in the state- and action-values are omitted, the result reduces to its counterpart in classic MDPs.

\begin{theorem}\label{thm:optim-pol-nvmdp}
Consider $\Pi = \{(\pi_0, \pi_1, \pi_2, ...) \mid \pi_i: S \rightarrow \mathrm{Dist}(A) \ , \forall i \ge 0\}$.
Denote the optimal state-values and action-values at time $t$ as:
\begin{equation*}
\begin{aligned}
V^*_t(s) &= \sup_{\pi \in \Pi} V^{\pi}_t(s) \ , \\
Q^*_t(s, a) &= \sup_{\pi \in \Pi} Q^{\pi}_t(s, a) \ .
\end{aligned}
\end{equation*}
Then there exists a deterministic policy $\pi^* \in \Pi$ which is optimal:
\begin{equation*}
\begin{aligned}
V^{\pi^*}_t(s) &= V^*_t(s) \quad(\forall t \ge 0), \\
Q^{\pi^*}_t(s, a) &= Q^*_t(s, a) \quad(\forall t \ge 0).
\end{aligned}
\end{equation*}
(A deterministic policy selects one action w.p. 1 in every state at every time.)
\end{theorem}

The corollary below directly follows from Theorem \ref{thm:optim-pol-nvmdp}:
\begin{corollary}\label{thm:optim-sv-av-relp}
For all $t \ge 0, s \in S$,
\begin{equation}\phantomsection\label{optim-sv-av-relp}
V^*_t(s) = \max_a Q^*_t(s, a) \ .
\end{equation}
\end{corollary}

By Theorem \ref{thm:optim-pol-nvmdp}, the optimal state- and action-values in any NVMDP are unique, and there exists a deterministic policy that attains these values.
Although the optimal state- and action-values are unique, multiple policies may yield these same optimal values.
In the remainder of this work, any policy that corresponds to the optimal state- and action-values will be referred to as an optimal policy.

While Theorem \ref{thm:optim-pol-nvmdp} guarantees the existence of globally optimal state- and action-values and a corresponding optimal policy, it does not provide a constructive method to obtain them.
Since an NVMDP is defined over an infinite horizon, enumerating all states and actions across all time steps is theoretically infeasible.
Instead, a common approach to finding an optimal policy is to iteratively compute optimal action-values via the \textbf{Bellman optimality equation}, a fundamental tool for verifying the optimality of action-values in classic MDPs.

According to Agarwal et al. \cite{RLTA-2022}, the Bellman optimality equation in classic MDPs is expressed as:
\begin{equation*}
Q(s, a) = \bar{r}(s, a) + \gamma \E^{s' \sim p(\cdot \mid s, a)}[\max_{a'} Q(s', a')] \ ,
\end{equation*}
and for finite-horizon tasks with discount rate $\gamma = 1$:
\begin{equation*}
Q_t(s, a) = \bar{r}_t(s, a) + \E^{s' \sim p_t(\cdot \mid s, a)}[\max_{a'} Q_{t + 1}(s', a')] \ .
\end{equation*}
In NVMDPs, since a discount rate depends on both the time and the transition, the rate effectively ``moves'' inside the expectation operator in the adapted form of the Bellman optimality equation, as formalized in the following theorem:
\begin{theorem}\label{thm:bellman-nvmdp}
For the following Bellman optimality equation in an NVMDP:
\begin{equation}\phantomsection\label{bellman-equ-nvmdp}
Q^\pi_t(s, a) = \bar{r}_t(s, a) + \E^{s' \sim p_t(\cdot \mid s, a)}[\gamma_{t + 1}(s, a, s') \max_{a'} Q^{\pi}_{t + 1}(s', a')] \ ,
\end{equation}
where $\bar{r}_t(s, a)$ is defined by Equation \eqref{avg-rwd-nvmdp}: \\
(1) Equation \eqref{bellman-equ-nvmdp} holds for any policy $\pi$ that is optimal. \\
(2) Action-values satisfy Equation \eqref{bellman-equ-nvmdp} are optimal.
\end{theorem}

A direct consequence of Theorem \ref{thm:bellman-nvmdp} is that augmenting the reward function with additional terms in certain differential forms preserves the optimal policies, a property known as reward shaping.
The original result, established by Ng et al. \cite{Rwd-Shaping-1999} in classic MDPs, demonstrates that appropriate reward shaping can, in some cases, significantly accelerate the convergence of RL algorithms.
We extend this result to NVMDPs in the following corollary:
\begin{corollary}\label{thm:reward-shaping-nvmdp}
For an NVMDP and functions $\Phi_t: S \rightarrow \mathbb{R} \ (\forall t \ge 0)$, if
\begin{equation*}
\widetilde{r}_t(s, a, s') = r_t(s, a, s') + \gamma_{t + 1}(s, a, s') \Phi_{t + 1}(s') - \Phi_t(s) \quad(\forall t \ge 0, s, s' \in S, a \in A)
\end{equation*}
is bounded in absolute-value, then an optimal policy under the reward $r_t(s, a, s')$ is also optimal under the reward $\widetilde{r}_t(s, a, s')$, and vice versa.
\end{corollary}


\subsection{Policy Improvement}\label{sec:pol-imprv}

Theorem \ref{thm:bellman-nvmdp} provides one approach for finding an optimal policy: first, identify the optimal action-values that satisfy Equation \eqref{bellman-equ-nvmdp}, and then derive the corresponding deterministic policy from these optimal action-values.
An alternative approach is policy improvement, in which the agent iteratively refines a policy based on its current state-values, rather than obtaining an optimal policy from the optimal action-values.
For this method, the policy improvement theorem provides a theoretical guarantee, ensuring that state-values increase monotonically with each policy update.
Sutton et al. \cite{Sutton-RL-2018} present the policy improvement theorem in classic MDPs.
Here, we extend this result and present a generalized version of the policy improvement theorem in NVMDPs.

\begin{theorem}\label{thm:pol-imprv-nvmdp}
For a given time $n$, policy $\pi$ and $\pi'$ are identical for all time $t \ge n$.
Furthermore,
\begin{equation*}
\sum_{\bar{a} \in A} \pi'_t(\bar{a} | s) Q^\pi_t(s, \bar{a}) \ge V^\pi_t(s) \quad\forall s \in S, t < n \ .
\end{equation*}
Then: \\
(1) for any $s \in S$, $V^{\pi'}_t(s) = V^\pi_t(s) \ (\forall t \ge n)$ and $V^{\pi'}_t(s) \ge V^\pi_t(s) \ (\forall t \le n - 1)$. \\
(2) for any $(s, a) \in S \times A$, $Q^{\pi'}_t(s, a) = Q^\pi_t(s, a) \ (\forall t \ge n - 1)$ and $Q^{\pi'}_t(s, a) \ge Q^\pi_t(s, a) \ (\forall t < n - 1)$.
\end{theorem}

A key distinction between the classic policy improvement theorem by Sutton et al. \cite{Sutton-RL-2018} and Theorem \ref{thm:pol-imprv-nvmdp} lies in the treatment of recurrence.
In an irreducible, classic MDP with policy $\pi$ and recurrent state $\bar{s}$,
the classic policy improvement theorem claims that if a deterministic policy $\pi'$ satisfies $Q^\pi(\bar{s}, \pi'(\bar{s})) \ge V^\pi(\bar{s})$ and agrees with $\pi$ elsewhere, then $\pi'$ improves upon $\pi$.
Since $\bar{s}$ is visited infinitely often, this local modification yields an infinite cumulative improvement.

In contrast, in an irreducible NVMDP, Theorem \ref{thm:bellman-nvmdp} states that if
$\E^{a \sim \pi'}[Q^\pi_{\bar{t}}(\bar{s}, a)] \ge V^\pi_{\bar{t}}(\bar{s})$
for some $(\bar{t}, \bar{s})$ and $\pi'$ matches $\pi$ otherwise, then $\pi'$ improves $\pi$ only for values up to time $\bar{t}$, since each time-state pair $(\bar{t}, \bar{s})$ occurs once per episode.
Thus, Theorem \ref{thm:bellman-nvmdp} provides a finer, time-dependent view of policy improvement compared to the classic policy improvement theorem.

Although the policy improvement theorem provides a fundamental theoretical guarantee 
that a refined policy (e.g., a greedy policy) derived from the current action-values yields performance no worse than the current policy, the theorem itself does not provide 
concrete algorithmic procedures.
Instead, it serves as a conceptual foundation: many RL algorithms---such as policy iteration, actor-critic methods, and TRPO---are built upon the principle of iteratively refining an existing policy using current value-based information.
Thus, while the theorem guides algorithm development, it is not an algorithmic recipe by itself.


\subsection{Special Cases: Classic MDPs and Finite-horizon MDPs}\label{sec:spcl-cases-nvmdp}

In this subsection, we demonstrate how two commonly used MDP formulations in RL---the classic MDP and the finite-horizon MDP---arise as special cases of our NVMDP framework.
We begin with the classic MDP.

Consider a generalization of classic MDPs where the dynamics are stationary and the discount rates depend only on transitions, i.e., of the form $\gamma(s, a, s')$, with all rates lying in $[0, 1)$.
As a result, the optimal state- and action-values computed within the NVMDP framework become time-independent.
Furthermore, if $\pi^*_{\bar k} : S \rightarrow \mathrm{Dist}(A)$ denotes a sub-policy of an optimal policy at some time $\bar k$, then applying $\pi^*_{\bar k}$ uniformly across all time steps also yields an optimal policy.
The following theorem formalizes these statements.

\begin{theorem}\label{thm:station-nvmdp-is-classic}
If a NVMDP satisfies $r_t(s, a, s') = r(s, a, s')$ and $p_t(s', r | s, a) = p(s', r | s, a)$ for all $t \ge 0, s \in S, a \in A$ and all discount rates $\gamma_{t + 1}(s, a, s') = \gamma(s, a, s') \in [0, 1)$, then: \\
(1) $V^*_t(s) = V^*_k(s)$ and $Q^*_t(s, a) = Q^*_k(s, a)$ for all $t, k \ge 0, s \in S, a \in A$. \\
(2) for optimal policy $\pi^* = (\pi^*_0, \pi^*_1, \pi^*_2, ...)$ and any integer ${\bar k} \ge 0$, policy $(\pi^*_{\bar k}, \pi^*_{\bar k}, \pi^*_{\bar k}, ...)$ is also optimal.
\end{theorem}

By Theorem \ref{thm:optim-pol-nvmdp}, every NVMDP has an optimal policy.
In particular, if an NVMDP is also a classic MDP, then by Theorem \ref{thm:station-nvmdp-is-classic} this MDP always admits an optimal, stationary policy.
Therefore, when the primary objective in studying an MDP is to identify one optimal policy, it suffices to restrict attention to stationary policies, which reduces memory costs.
Hence, classic MDPs can be seen as special cases of NVMDPs, where the objective is to identify a single optimal policy, as in typical RL tasks.

The finite-horizon MDP is another widely used formulation in episodic RL tasks, as all episodes must end within a finite number of steps in practice. 
In episodic tasks, each episode can be treated as having a finite horizon $H$, such that the episode terminates before the time step $t$ reaches $H$.
For state- and action-values defined in the infinite-horizon setting, a common approach under NSMDPs is to view $r_t(\cdot) = 0$ for all $t \ge H$.
Alternatively, within the NVMDP framework, the same effect can be achieved by viewing $\gamma_{t+1}(\cdot) = 0$ for all $t \ge H-1$, without modifying the reward structure.

Moreover, since the setting of $\gamma_{t + 1}(\cdot) = 0 \ (\forall t \ge H - 1)$ naturally satisfies Assumptions \ref{asm:Gamma-sum-absbnd-nvmdp} and \ref{asm:gamma-bellman-cond-nvmdp}, this formulation allows flexibility in assigning any non-negative values to $\gamma_{t + 1}(s, a, s')$ for $0 \le t < H - 1$ in finite-horizon tasks, provided that the resulting returns remain within the numerical limits of the computing environment.
Experiments presented in Section \ref{sec:expt} further demonstrate that allowing certain discount rates to exceed 1 still yields effective learning, though at the cost of increased variance in the action-value estimates.

The Stochastic Shortest Path (SSP) problem represents another class of MDPs that shares conceptual similarities with finite-horizon MDPs.
Although SSPs do not impose an explicit time limit, each trajectory terminates almost surely upon reaching an absorbing (terminal) state $T$.
Analogous to the finite-horizon case, one can view $r_t(T, a, T) = 0 \ (\forall a \in A)$ in NSMDPs, or equivalently, $\gamma_{t + 1}(s, a, T) = 0 \ (\forall s \in S, a \in A)$ in NVMDPs.


\section{Tabular Solution Methods}\label{sec:tab-solu}

When the state space $S$ and the action space $A$ are finite and relatively small, the state-value function, the action-value function, and the policy can all be stored in tables.
RL algorithms based on such tabular representation are generally referred to as \textit{tabular methods} or \textit{tabular solution methods}.

Although tabular methods address RL problems at smaller scales, research on tabular methods in RL remains active.
Fundamentally, tabular methods offer two key advantages for RL research.
First, they often serve as algorithmic prototypes for large-scale problems.
The Deep Q-Networks (DQN) algorithm proposed by Mnih et al.\ \cite{Mnih-DQN-2013} is a typical example: as a widely used deep RL algorithm, DQN extends Q-learning (Watkins \cite{q-learn-1992}) by replacing tabular action-value estimates with deep neural networks.

Second, many tabular methods come with explicit convergence guarantees, whereas algorithms combined with function approximation---required for larger problems---typically have weaker or less general convergence results.
Sutton et al.\ \cite{Sutton-RL-2018} provide a thorough discussion of such tabular methods in classic MDPs.

In this section, we first discuss the treatment of time, and then present the adaptation of dynamic programming and (generalized) Q-learning to NVMDPs, together with their associated convergence guarantees.
Since tabular methods constitute a large body of work within RL, we can only adapt a small subset of them to NVMDPs.
Nevertheless, we believe that our selected algorithms sufficiently demonstrate that tabular methods developed for classic MDPs have strong potential for adaptation to NVMDPs.

\subsection{Treatment of Time}\label{sec:time-trmt}

The NVMDP framework addresses non-stationarity and varying discount rates through time-dependent state-values, action-values, and policies.
Unlike a classic MDP with stationary $V^\pi(s)$, $Q^\pi(s, a)$, and a global policy $\pi(s, a)$, an NVMDP employs time-indexed forms $V^\pi_t(s)$, $Q^\pi_t(s, a)$, and $\pi_t(s, a)$, enabling them to capture temporal variations in returns.

A common method to introduce time-awareness in RL is state-time augmentation, where each state $s_t$ becomes $(t, s_t)$.
Pardo et al. \cite{Time-Limit-RL-2018} demonstrate that this technique improves RL performance, as
\begin{equation*}
V^\pi((t, s_t)) = V^\pi_t(s_t) \ , \quad Q^\pi((t, s_t), a_t) = Q^\pi_t(s_t, a_t) \ , \quad \pi(a_t | (t, s_t)) = \pi_t(a_t | s_t) \ .
\end{equation*}

Alternatively, time can be treated as an additional index rather than embedded in states.
While in a classic MDP value functions and policies are stored in tables of size $|S|$ or $|S| \times |A|$, in an NVMDP with a finite horizon $H$ the corresponding tables store:
\begin{equation*}
V^\pi_t(\cdot): H \times |S| \ , \quad Q^\pi_t(\cdot), \pi_t(\cdot): H \times |S| \times |A| \ ,
\end{equation*}
thereby preserving the original state space.
Notably, both approaches increase memory requirements linearly with $H$.

In practice, imposing a finite horizon $H$ truncates returns for $t \ge H$, introducing approximation errors in value estimates near $H$.
However, this impact diminishes for earlier time steps, as demonstrated later in our adapted dynamic programming and (generalized) Q-learning algorithms.


\subsection{Dynamic Programming}\label{sec:dp-nvmdp}

Dynamic programming (DP) is a class of algorithms that solve complex optimization problems by decomposing them into simpler, recursive subproblems, as introduced by Bellman \cite{Bellman-DP-1957}.
In RL, DP methods such as policy evaluation and value iteration exploit this recursive structure when the environment dynamics are known.
Policy evaluation computes state- and action-values for a fixed policy $\pi$, whereas value iteration alternates between evaluation and improvement to estimate the optimal values, as described by Sutton et al. \cite{Sutton-RL-2018}.

\begin{algorithm}
\caption{DP Policy Evaluation in NVMDPs}\label{algo:dp-poleval-nvmdp}
\begin{algorithmic}
\STATE{Input: horizon $H$, policy $\pi$}
\STATE{Initialize ${\hat V}^\pi_t(s) = 0$ and ${\hat Q}^\pi_t(s, a) = 0$ for all $t \le H, s \in S, a \in A$}
\STATE{$t \leftarrow H - 1$}
\WHILE{$t \ge 0$}
	\FOR{$s \in S$}
		\FOR{$a \in A$}
			\FOR{$s', r$ and the corresponding probability $p$}
				\STATE{${\hat Q}^\pi_t(s, a) \text{ += } p \times (r + \gamma_{t + 1}(s, a, s') {\hat V}^\pi_{t + 1}(s'))$}
			\ENDFOR
			\STATE{${\hat V}^\pi_t(s) \text{ += } \pi_t(a|s){\hat Q}^\pi_t(s, a)$}
		\ENDFOR
	\ENDFOR
	\STATE{$t \leftarrow t - 1$}
\ENDWHILE
\RETURN{${\hat V}^\pi_t(\cdot), {\hat Q}^\pi_t(\cdot)$}
\end{algorithmic}
\end{algorithm}

The policy evaluation procedure in NVMDPs is presented in Algorithm \ref{algo:dp-poleval-nvmdp}, which is derived from Equation \eqref{sv-av-nvmdp} and Equation \eqref{av-iter-nvmdp} for estimating the state- and action-values, respectively.
The value iteration method in NVMDPs is given by Algorithm \ref{algo:dp-valiter-nvmdp}.
It builds on Equation \eqref{optim-sv-av-relp} and Equation \eqref{bellman-equ-nvmdp} to estimate the optimal state- and action-values, respectively.

\begin{algorithm}
\caption{DP Value Iteration in NVMDPs}\label{algo:dp-valiter-nvmdp}
\begin{algorithmic}
\STATE{Input: horizon $H$}
\STATE{Initialize ${\hat V}^*_t(s) = 0$ and ${\hat Q}^*_t(s, a) = 0$ for all $t \le H, s \in S, a \in A$}
\STATE{$t \leftarrow H - 1$}
\WHILE{$t \ge 0$}
	\FOR{$s \in S$}
		\FOR{$a \in A$}
			\FOR{$s', r$ and the corresponding probability $p$}
				\STATE{${\hat Q}^*_t(s, a) \text{ += } p \times (r + \gamma_{t + 1}(s, a, s') {\hat V}^*_{t + 1}(s'))$}
			\ENDFOR
		\ENDFOR
		\STATE{${\hat V}^*_t(s) \leftarrow \max_a {\hat Q}^*_t(s, a)$}
	\ENDFOR
	\STATE{$t \leftarrow t - 1$}
\ENDWHILE
\RETURN{${\hat V}^*_t(\cdot), {\hat Q}^*_t(\cdot)$}
\end{algorithmic}
\end{algorithm}

A key difference between the policy evaluation in Algorithm \ref{algo:dp-poleval-nvmdp} and the value iteration in Algorithm \ref{algo:dp-valiter-nvmdp}, compared with their counterparts in classic MDPs, is that the main loops in the NVMDP versions traverse backward in the time index~$t$, without revisiting the same state at the same time step.
In contrast, the classic MDP procedures update the value estimates of the same states multiple times until the gap between successive updates becomes sufficiently small.

As discussed in Section \ref{sec:time-trmt}, this updating scheme in Algorithm \ref{algo:dp-poleval-nvmdp} and Algorithm \ref{algo:dp-valiter-nvmdp} may introduce larger approximation errors in state- and action-value estimates near the time horizon $H$.
For a time window of interest $[0, n]$, a practical remedy is to select a horizon $H$ such that $H \gg n$.
Under this condition, the estimated state- and action-values in policy evaluation by Algorithm \ref{algo:dp-poleval-nvmdp} deviate only slightly from their theoretical values, as guaranteed below:
\begin{theorem}\label{thm:dp-poleval-nvmdp-limit0}
Consider a NVMDP and a given policy $\pi$.
For any $t \ge 0$, the estimated state-values $\hat{V}^\pi_t(\cdot)$ and action-values $\hat{Q}^\pi_t(\cdot)$ by Algorithm \ref{algo:dp-poleval-nvmdp} with horizon $H$ satisfy:
\begin{equation*}
\lim_{H \rightarrow \infty} \hat{V}^\pi_t(s) = V^\pi_t(s) \ , \quad \lim_{H \rightarrow \infty} \hat{Q}^\pi_t(s, a) = Q^\pi_t(s, a) \quad(\forall s \in S, a \in A) \ .
\end{equation*}
\end{theorem}
\noindent Similarly, value iteration by Algorithm \ref{algo:dp-valiter-nvmdp} is guaranteed by the following:
\begin{theorem}\label{thm:dp-valiter-nvmdp-limit0}
For a NVMDP and any $t \ge 0$, the estimated optimal state-values $\hat{V}^*_t(\cdot)$ and action-values $\hat{Q}^*_t(\cdot)$ by Algorithm \ref{algo:dp-valiter-nvmdp} with horizon $H$ satisfy:
\begin{equation*}
\lim_{H \rightarrow \infty} \hat{V}^*_t(s) = V^*_t(s) \ , \quad \lim_{H \rightarrow \infty} \hat{Q}^*_t(s, a) = Q^*_t(s, a) \quad(\forall s \in S, a \in A) \ .
\end{equation*}
\end{theorem}

Although DP algorithms can yield highly accurate state- and action-values, they are often computationally and memory intensive and require full knowledge of the environment dynamics.
In high-dimensional, finely discretized state spaces, storing state-values and computing transitions can become intractable, a problem further exacerbated by large action spaces, numerous outcomes, or incorporating the time index as an additional dimension.
Moreover, obtaining exact transition probabilities for all state-action pairs may be impractical, even in many simulated environments.

Model-free RL algorithms avoid this limitation by assuming the environment can be sampled infinitely often, allowing the agent to gather sufficient experience.
This assumption is adopted in this work for all algorithms except DP.


\subsection{Q-learning}\label{sec:q-learn}

Sutton's seminal work \cite{TD-1988} introduces temporal-difference (TD) learning, forming the foundation of RL by integrating Monte Carlo methods with DP.
Building on this, Watkins \cite{Watkins-PhD-1989} proposes Q-learning, a TD control algorithm estimating optimal action-values analogous to value iteration.
To distinguish it from later variants and our extensions, we refer to Watkins’s original algorithm for classical MDPs as \textbf{classic Q-learning} throughout this work.

In classic Q-learning, the estimated action-value $\hat Q(\cdot)$ is updated by
\begin{equation}\phantomsection\label{q-learn-clascmdp-upd}
{\hat Q}(s, a) \leftarrow {\hat Q}(s, a) + \alpha \Big(r + \gamma \big(\max_{a'}{\hat Q}(s', a')\big) - {\hat Q}(s, a)\Big) \ ,
\end{equation}
when the agent receives a reward $r$ by selecting action $a$ in state $s$ and transitions to the next state $s'$.
Here $\gamma \in (0,1)$ is the constant discount factor and $\alpha \in [0, 1]$ denotes the step size.
For the convergence of $\hat Q(\cdot)$ to the true optimals, the sequence of $\{\alpha\}$ is typically assumed to be a \textit{tapening step size}:
\begin{equation*}
\sum^\infty_{i = 1} \alpha_{n_i(s, a)} = \infty, \ \sum^\infty_{i = 1} \alpha^2_{n_i(s, a)} < \infty \ ,
\end{equation*}
where $n_i(s, a)$ denotes the index of the $i$-th visit to state-action pair $(s, a)$ across all episodes.

Convergence further requires that each reachable $(s,a)$ pair is visited infinitely often as the total number of training steps grows.
This is commonly ensured via $\varepsilon$-greedy exploration, where the agent selects the greedy action with probability $1 - \varepsilon$ and a random action with probability $\varepsilon$.
While greedy actions promote exploitation of current estimates, random choices ensure sufficient long-term exploration of all reachable state-action pairs.

Watkins \cite{q-learn-1992} established the convergence of classic Q-learning via action-replay processes (ARPs).
While elegant, this approach has limited applicability in more general settings, as observed by Tsitsiklis \cite{stoch-approx-ql-1994}.
To address broader scenarios, subsequent works have employed alternative theoretical tools, including stochastic approximation techniques \cite{stoch-approx-ql-1994} and ordinary differential equation (ODE) methods \cite{ODE-stochapprox-RL-2000}.
Building on this foundation, we introduce the NVMDP-Q-learning algorithm, the counterpart of classic Q-learning in NVMDPs.
As NVMDP-Q-learning (Algorithm \ref{algo:q-learn-nvmdp}) is a special case within a broader algorithmic framework, we defer the discussion of its convergence guarantees to the next subsection.

\begin{algorithm}
\caption{Q-learning in NVMDPs (NVMDP-Q-learning)}\label{algo:q-learn-nvmdp}
\begin{algorithmic}
\STATE{Input: horizon $H$, tapening step size $\alpha$}
\STATE{Initialize ${\hat Q}_t(s, a) = 0$ for all $t \le H, s \in S, a \in A$}
\FOR{each episode}
	\STATE{$t \leftarrow 0$, initialize $s$ to a start state}
	\FOR{each step in the current episode}
		\STATE{sample $a$, observe $s', r$}
		\STATE{${\hat Q}_t(s, a) \text{ += } \alpha \times (r + \gamma_{t + 1}(s, a, s') \max_{a'}{\hat Q}_{t + 1}(s', a') - {\hat Q}_t(s, a))$}
		\STATE{$t \leftarrow t + 1, s \leftarrow s'$}
	\ENDFOR
\ENDFOR
\RETURN{${\hat Q}_t(\cdot)$}
\end{algorithmic}
\end{algorithm}


\subsection{Generalized Q-learning}\label{sec:geq-learn}

Building on the theoretical work of Tsitsiklis \cite{stoch-approx-ql-1994}, Lan et al. \cite{Maxmin-q-learn-2020} introduced the generalized Q-learning framework.
In classic MDPs, their framework maintains $n$ estimation tracks $\bm{\hat Q}^{(1)}(\cdot), \bm{\hat Q}^{(2)}(\cdot), ..., \bm{\hat Q}^{(n)}(\cdot)$ and selects one track uniformly at random for updating at each time step.
Moreover, the framework retains $l$ most recent estimates for each track, denoted by ${\hat Q}^{(i, j)}(s, a)$ for the $j$-th most recent estimate of the optimal action-value $Q^*(s, a)$ in the $i$-th track $\bm{\hat Q}^{(i)}(\cdot)$.
Accordingly, the tensor $\bm{\hat Q}(s, \cdot) \in {\mathbb R}^{|A| \times n \times l}$ is defined as
\begin{equation*}
\bm{\hat Q}(s, a) = \begin{bmatrix}
{\hat Q}^{(1, 1)}(s, a) & {\hat Q}^{(1, 2)}(s, a) & \cdots & {\hat Q}^{(1, l)}(s, a) \\
{\hat Q}^{(2, 1)}(s, a) & {\hat Q}^{(2, 2)}(s, a) & \cdots & {\hat Q}^{(2, l)}(s, a) \\
\vdots & \vdots & \ddots & \vdots \\
{\hat Q}^{(n, 1)}(s, a) & {\hat Q}^{(n, 2)}(s, a) & \cdots & {\hat Q}^{(n, l)}(s, a)
\end{bmatrix} \quad(\forall s \in S, a \in A) \ .
\end{equation*}

The update rule in generalized Q-learning is given by
\begin{equation}\phantomsection\label{geq-learn-clascmdp-upd}
\textit{latest estimate of } {\hat Q}^{(i)}(s, a) \leftarrow {\hat Q}^{(i, 1)}(s, a) + \alpha \Big(r + \gamma f\big(\bm{\hat Q}(s', \cdot)\big) - {\hat Q}^{(i, 1)}(s, a)\Big) \ ,
\end{equation}
where the agent receives a reward $r$ and transitions to the next state $s'$ after selecting action $a$ in state $s$.
Here, the $i$-the estimation track is selected, and $\alpha \in [0, 1]$ denotes the step size, analogous to the update rule in Equation \eqref{q-learn-clascmdp-upd}.
The function $f: {\mathbb R}^{|A| \times n \times l} \rightarrow {\mathbb R}$ satisfies the following assumptions:
\begin{assumption}\label{asm:geq-f-maxq-cond}
For $\bm{Q}(s, \cdot) \in {\mathbb R}^{|A| \times n \times l}$, if
\begin{equation*}
Q^{(i, j)} (s, a) = Q^{(i', j')} (s, a) \quad(\forall 1 \le i, i' \le n, 1 \le j, j' \le l) \ ,
\end{equation*}
then $f\big(\bm{Q}(s, \cdot)\big) = \max_{a, i, j} Q^{(i, j)}(s, a)$.
\end{assumption}
\begin{assumption}\label{asm:geq-f-absbnd-cond}
For any $\bm{Q}(s, \cdot), \bm{Q'}(s, \cdot) \in {\mathbb R}^{|A| \times n \times l}$, function $f$ satisfies:
\begin{equation*}
\big|f\big(\bm{Q}(s, \cdot)\big) - f\big(\bm{Q'}(s, \cdot)\big)\big| \le \max_{a, i, j} \big| Q^{(i, j)}(s, a) - {Q'}^{(i, j)}(s, a) \big|
\end{equation*}
\end{assumption}

By Assumption \ref{asm:geq-f-maxq-cond}, function $f$ returns $\max_a Q^*(s', a)$ when its input is the tensor of optimal action-values (i.e., ${\hat Q}^{(i, j)}(s, a) = Q^*(s, a) \ (\forall 1 \le i \le n, 1 \le j \le l)$).
Notably, when $n = l = 1$ and $f\big(\bm{\hat Q}(s, \cdot)\big) = \max_a {\hat Q}(s, a)$, the generalized update rule \eqref{geq-learn-clascmdp-upd} reduces exactly to the classic Q-learning update rule \eqref{q-learn-clascmdp-upd}.

We adapt the generalized Q-learning framework by Lan et al. \cite{Maxmin-q-learn-2020} to NVMDPs as follows.
In NVMDPs, $n$ estimation tracks $\bm{\hat Q}^{(1)}_t(\cdot), \bm{\hat Q}^{(2)}_t(\cdot), ..., \bm{\hat Q}^{(n)}_t(\cdot)$ are maintained, with ${\hat Q}^{(i, j)}_t(s, a)$ denote the $j$-th most recent estimate of the optimal action-value $Q^*_t(s, a)$ in the $i$-th track $\bm{\hat Q}^{(i)}_t(\cdot)$.
Each estimation track contains $l$ most recent estimates, i.e., $j \le l$.
Moreover, tensor $\bm{\hat Q}_t(s', \cdot)$ is defined as
\begin{equation}\phantomsection\label{tensor-geq-def-nvmdp}
\bm{\hat Q}_t(s, a) = \begin{bmatrix}
{\hat Q}_t^{(1, 1)}(s, a) & {\hat Q}_t^{(1, 2)}(s, a) & \cdots & {\hat Q}_t^{(1, l)}(s, a) \\
{\hat Q}_t^{(2, 1)}(s, a) & {\hat Q}_t^{(2, 2)}(s, a) & \cdots & {\hat Q}_t^{(2, l)}(s, a) \\
\vdots & \vdots & \ddots & \vdots \\
{\hat Q}_t^{(n, 1)}(s, a) & {\hat Q}_t^{(n, 2)}(s, a) & \cdots & {\hat Q}_t^{(n, l)}(s, a)
\end{bmatrix} \quad(\forall s \in S, a \in A) \ .
\end{equation}
The settings is analogous to generalized Q-learning in classic MDPs, with the difference that action-value estimates in NVMDPs are time-dependent.
Alternatively, one could view states in NVMDPs as time-augmented states $(t, s_t)$.

\begin{algorithm}
\caption{Generalized Q-learning in NVMDPs}\label{algo:geq-learn-nvmdp}
\begin{algorithmic}
\STATE{Input: horizon $H$, tapening step size $\alpha$, functions $f_1, f_2, ..., f_n$ that satisfying Assumption \ref{asm:geq-f-maxq-cond} and Assumption \ref{asm:geq-f-absbnd-cond}}
\STATE{Initialize ${\hat Q}_t(s, a) = 0$ for all $t \le H, s \in S, a \in A$}
\FOR{each episode}
	\STATE{$t \leftarrow 0$, initialize $s$ to a start state}
	\FOR{each step in the current episode}
		\STATE{sample $a$, observe $s', r$}
		\STATE{select $i$ uniformly at random from $\{1, 2, ..., n\}$}
		\STATE{${\hat Q}_t^{(i, j)}(s, a) \leftarrow {\hat Q}_t^{(i, j - 1)}(s, a) \ (\text{for } j = l, l - 1, ..., 2 \text{ in turn})$}
		\STATE{${\hat Q}_t^{(i, 1)}(s, a) \text{ += } \alpha \Big(r + \gamma_{t + 1}(s, a, s') {f_i}\big(\bm{\hat Q}_{t + 1}(s', \cdot)\big) - {\hat Q}_t^{(i, 1)}(s, a)\Big)$}
		\STATE{$t \leftarrow t + 1, s \leftarrow s'$}
	\ENDFOR
\ENDFOR
\RETURN{$\bm{\hat Q}_t(\cdot)$}
\end{algorithmic}
\end{algorithm}

Algorithm \ref{algo:geq-learn-nvmdp} outlines the generalized Q-learning procedure for NVMDPs under the assumption of a finite horizon $H$ and that every reachable state–action pair at each time step $t \le H$ is visited infinitely often.
This condition can be ensured through the $\varepsilon$-greedy exploration strategy described in Section \ref{sec:q-learn}, and under the tapening step size introduced therein, the algorithm converges almost surely.
Theorem \ref{thm:geq-learn-nvmdp-convg} formally establishes the convergence result.
Although the theorem assumes that all state–action pairs are visited infinitely often for analytical convenience, this requirement applies only to the reachable pairs in practice.

\begin{theorem}\label{thm:geq-learn-nvmdp-convg}
Let $n_m(t, s, a)$ and $k$ denote, respectively, the index of the $m$-th visit to the state–action pair $(s, a)$ at time step $t$ and the total number of completed iterations of the inner for-loop in Algorithm \ref{algo:geq-learn-nvmdp} across all episodes.
If there exists constant $\bar C$ s.t.
\begin{equation*}
\sum^\infty_{m = 1} \alpha_{n_m(t, s, a)} = \infty, \ \sum^\infty_{m = 1} \alpha^2_{n_m(t, s, a)} \le \bar C \quad\textit{w.p. 1} \ (\forall 0 \le t < H, s \in S, a \in A)\ ,
\end{equation*}
and
\begin{equation*}
\lim_{k \rightarrow \infty} n_m(t, s, a) = \infty \quad\textit{w.p. 1} \ .
\end{equation*}
Then for all $1 \le i \le n, 1 \le j \le l, 0 \le t < H, s \in S, a \in A$: 
\begin{equation*}
\lim_{k \rightarrow \infty} {\hat Q}_t^{(i, j)}(s, a) = \hat{Q}^*_t(s, a) \quad\textit{w.p. 1},
\end{equation*}
where $\hat{Q}^*_t(s, a)$ equals the results of Algorithm \ref{algo:dp-valiter-nvmdp} with horizon $H$.
\end{theorem}

As noted by Lan et al. \cite{Maxmin-q-learn-2020}, generalized Q-learning provides a unified framework in classic MDPs that encompasses classic Q-learning, Averaged Q-learning \cite{avg-q-learn-2017}, Ensemble Q-learning \cite{avg-q-learn-2017}, and Historical Best Q-learning \cite{hist-q-learn-2018} as special cases.
In NVMDPs, when $f_1 = f_2 = \dots = f_n = f$, a variant generalizing Averaged and Ensemble Q-learning is:
\begin{equation*}\phantomsection\label{avg-q-learn-f}
f\big(\bm{\hat Q}_t(s, \cdot)\big) = \max_{a \in A} {1 \over nl} \sum_{i = 1}^n \sum_{j = 1}^l {\hat Q}^{(i, j)}_t(s, a) \ ,
\end{equation*}
which we continue to refer to as \textbf{Averaged Q-learning}.
Alternatively, \textbf{Maxmin Q-learning} is defined by
\begin{equation*}\phantomsection\label{mmx-q-learn-f}
f\big(\bm{\hat Q}_t(s, \cdot)\big) = \max_{a \in A} \min_{1 \le i \le n, 1 \le j \le l} {\hat Q}^{(i, j)}_t(s, a) \ ,
\end{equation*}
originally proposed by Lan et al. \cite{Maxmin-q-learn-2020} for $l = 1$.
We extend this formulation to any $l \in \mathbb{Z}_{\ge 1}$, supported by the following lemma:
\begin{lemma}\label{thm:maxmin-q-learn-inequal}
For any $\bm{Q}(s, \cdot), \bm{Q'}(s, \cdot) \in {\mathbb R}^{|A| \times n \times l}$ and any non-empty set $F \subseteq \{1, 2, ..., n\} \times \{1, 2, ..., l\}$,
\begin{equation*}
\big|\max_{a \in A} \min_{(i, j) \in F} Q^{(i, j)}(s, a) - \max_{a \in A} \min_{(i, j) \in F} {Q'}^{(i, j)}(s, a)\big| \le \max_{a \in A, (i, j) \in F} \big| Q^{(i, j)}(s, a) - {Q'}^{(i, j)}(s, a) \big| \ .
\end{equation*}
\end{lemma}

When $f_i \ (1 \le i \le n)$ differ in NVMDPs, Averaged Q-learning can be extended to
\begin{equation*}\phantomsection\label{wta-q-learn-f}
{f_i}\big(\bm{\hat Q}_t(s, \cdot)\big) = \max_{a \in A} \Big(\eta \sum_{j = 1}^l \lambda^{j - 1} {{1 - \lambda} \over {1 - \lambda^l}} {\hat Q}^{(i, j)}_t(s, a) + {{1 - \eta} \over {n - 1}} \sum_{\bar{i} \neq i} \sum_{j = 1}^l \lambda^{j - 1} {{1 - \lambda} \over {1 - \lambda^l}} {\hat Q}^{(\bar{i}, j)}_t(s, a)\Big) \ ,
\end{equation*}
which we term \textbf{Weighted-Average Q-learning (WtAvg-Q)}. Here, $\lambda \in (0,1)$ controls the decay rate of older estimates, and $\eta \in (0,1)$ specifies the weight of the current estimate track.
For $n = 1$, the $\eta$ and ${{1 - \eta} \over {n - 1}}$ terms on the R.H.S are set to 1 and 0, respectively.
Similarly, Maxmin Q-learning can be adapted as \textbf{Present-Track-Maxmin Q-learning (PTMxm-Q)}:
\begin{equation*}\phantomsection\label{ptm-q-learn-f}
{f_i}\big(\bm{\hat Q}_t(s, \cdot)\big) = \max_{a \in A} \min_{1 \le j \le l} {\hat Q}^{(i, j)}_t(s, a) \ ,
\end{equation*}
where the minimum is taken over the present estimate track only.

The settings of $f_i \ (1 \le i \le n)$ in NVMDP-Q-learning (Section \ref{sec:q-learn}), Averaged Q-learning, Maxmin Q-learning, WtAvg-Q, and PTMxm-Q all satisfy Assumption \ref{asm:geq-f-maxq-cond} and Assumption \ref{asm:geq-f-absbnd-cond}.
Verification of Assumption \ref{asm:geq-f-maxq-cond} is straightforward for all variants; Assumption \ref{asm:geq-f-absbnd-cond} holds directly for NVMDP-Q-learning, Averaged Q-learning, and WtAvg-Q, while Maxmin Q-learning and PTMxm-Q follow from Lemma \ref{thm:maxmin-q-learn-inequal}.
Therefore, these variants converge almost surely by Theorem \ref{thm:geq-learn-nvmdp-convg}.
Experimental results in Section \ref{sec:expt} further confirm convergence and illustrate performance differences under multiple reward and discount rate settings.


\section{NVMDPs with Function Approximation}\label{sec:func-approx}

As discussed in Section \ref{sec:tab-solu}, tabular solution methods are intended for FMDP tasks in which both the state and action spaces are relatively small.
When either space becomes large, tabular methods typically struggle and quickly lose practicality.
For instance, modern RL tasks often fall outside the scope of FMDPs, featuring state or action spaces that are continuous.
Although fine discretization can manage low-dimensional cases, the curse of dimensionality renders tabular methods computationally and memory intensive in higher dimensions.

A common remedy is to replace value and policy tables with more efficient forms of ``storage''---function approximators.
Function approximation techniques mitigate the curse of dimensionality by enabling generalization to previously unseen states or actions.
Rather than storing a distinct value for each entry, value estimates and policies are represented as parameterized or non-parametric functions.
Such approximations allow the algorithms to generalize across similar states or neighboring actions, facilitating efficient learning in high-dimensional or continuous spaces.

However, this flexibility comes at a cost: function approximators introduce additional sources of error into state- and action-value estimates.
Consequently, RL with function approximation typically offers weaker convergence guarantees, and stability in practice depends heavily on the choice of approximator and hyperparameters.
Common categories of function approximators include linear models (Sutton et al. \cite{Sutton-RL-2018}), neural networks (Arulkumaran et al. \cite{DRL-survey-2017}), tree-based models (Ernst et al. \cite{Ernst-treeRL-2005}), and non-parametric models (Ormoneit et al. \cite{Ormoneit-kerRL-2002}).

Like tabular solution methods, RL with function approximation remains a large and active field.
To demonstrate the potential for our NVMDP framework to also accommodate function approximation, we extend two fundamental theoretical results in classic MDPs---the Policy Gradient Theorem and the policy improvement bound used in Trust Region Policy Optimization (TRPO)---to NVMDPs in the remainder of this section.
As this work primarily focuses on the theoretical structure of NVMDPs, we defer the design and empirical evaluation of algorithms incorporating function approximation to future work.

\subsection{Policy Gradient Theorem}

Sutton et al. \cite{Sutton-PolGrad-1999} introduced the policy gradient framework in classic MDPs, providing a foundational approach for directly optimizing parameterized policies in RL.
Unlike value-based methods that derive policies indirectly from value functions, policy gradient methods optimize the policy parameters directly via gradient on the expected cumulative reward.
For the objective of maximizing $V^\pi(s_0)$ in a classic MDP, Sutton et al. \cite{Sutton-PolGrad-1999} derived the policy gradient as
\begin{equation*}
\nabla V^\pi(s_0) = \sum_s \sum_{t = 0}^\infty \gamma^t P(s_t = s \mid s_0, \pi) \E^{a \sim \pi} [Q^\pi(s, a) {\nabla \pi(a | s) \over \pi(a | s)}] \ ,
\end{equation*}
where ``$\nabla$'' denotes the gradient with respect to the parameters of the policy approximator, as in the expressions that follow.

For NVMDPs, the policy gradient theorem extends as follows:
\begin{theorem}\label{thm:pol-grad-nvmdp}
For a NVMDP and its parameterized policy $\pi$,
\begin{equation}\phantomsection\label{pol-grad-nvmdp}
\nabla V^{\pi}_t(s_t) = \E^\pi [\sum_{i = t}^\infty \Gamma^\tau_{t, i} A^\pi_i(s_i, a_i) {\nabla \pi_i(a_i | s_i) \over \pi_i(a_i | s_i)} \mid s_t] \ .
\end{equation}
\end{theorem}

Notably, the advantage function $A^\pi_i(s_i, a_i)$ on the R.H.S. of Equation \eqref{pol-grad-nvmdp} can be replaced by the action-value function $Q^\pi_i(s_i, a_i)$, as shown in the proofs of Theorem \ref{thm:pol-grad-nvmdp} in Appendix \ref{sec:apndx-proofs}.
Furthermore, with this replacement and $\Gamma^\tau_{t, i}$ replaced by $\gamma^{i - t}$ in Equation \eqref{pol-grad-nvmdp}, Theorem \ref{thm:pol-grad-nvmdp} reduces to its counterpart in classic MDPs under the assumption of stationarity.

The policy gradient theorem offers several advantages that make it a powerful tool in RL.
By directly optimizing over the expected return, it provides a clear and theoretically grounded routine for policy improvement.
Moreover, it naturally supports stochastic policies and can handle continuous or high-dimensional action spaces, making it highly flexible in complex environments.
These properties allow algorithms based on the policy gradient theorem to generalize effectively and adapt to a wide range of RL problems.
For many RL algorithms integrated with function approximation, the principle of following the policy gradient lies at their core.

Despite these strengths, policy gradient methods also face notable challenges.
The gradient estimates often exhibit high variance, which slows learning and requires careful variance reduction techniques.
They can become sample-inefficient, achieving stable performance only after extensive interaction with the environment.
Furthermore, these methods are sensitive to step sizes and hyperparameter settings, and without additional mechanisms, updates sometimes reduce policy performance rather than improve it.


\subsection{Policy Improvement Framework of TRPO}

Trust Region Policy Optimization (TRPO) proposed by Schulman et al. \cite{TRPO-2015} is a RL algorithm introduced to address stability and reliability in policy gradient methods.
Unlike standard policy gradient algorithms that may take large, destabilizing updates to the policy, TRPO constrains each update to stay within a trust region.

The cornerstone of TRPO is the performance difference lemma, which quantifies the difference in state-values between two policies. 
Originally established by Kakade et al. \cite{App-OptimRL-2002} for classic MDPs, we adapt the performance difference lemma to NVMDPs as follows:
\begin{lemma}\label{thm:prfm-diff-lemma-nvmdp}
For a NVMDP and any two policies $\pi, \pi'$:
\begin{equation}\phantomsection\label{prfm-diff-lemma-nvmdp}
V^{\pi'}_t(s) - V^\pi_t(s) = \E^{\pi'} [\sum_{i = t}^\infty \Gamma^\tau_{t, i} A^\pi_i(s_i, a_i) \mid s_t = s] \ .
\end{equation}
\end{lemma}

Equation \eqref{prfm-diff-lemma-nvmdp} can be viewed as:
\begin{equation}\phantomsection\label{prfm-diff-exp}
V^{\pi'}_t(s) - V^\pi_t(s) = \sum_{i = t}^\infty \E^{a_t, s_{t + 1}, ..., s_i \sim \pi', a_i \sim \pi'} [\Gamma^\tau_{t, i} A^\pi_i(s_i, a_i) \mid s_t = s] \ .
\end{equation}
Consider $\pi$ and $\pi'$ as the old and new policies, respectively.
In the R.H.S of Equation \eqref{prfm-diff-exp}, both action $a_i$ and the trajectory preceding $a_i$ follow the new policy $\pi'$, which makes the expression difficult to handle analytically.

The \textbf{policy advantage function} $D^{\pi, \pi'}_t(s)$ is introduced to address this issue, which preserves the essential structure of the R.H.S in Equation \eqref{prfm-diff-exp}.
The key distinction is that the trajectory preceding $a_i$ is generated under the old policy $\pi$ while action $a_i$ is still selected by the new policy $\pi'$:
\begin{equation}\phantomsection\label{D-oper-def}
D^{\pi, \pi'}_t(s) = \sum_{i = t}^\infty \E^{a_t, s_{t + 1}, ..., s_i \sim \pi, a_i \sim \pi'} [\Gamma^\tau_{t, i} A^\pi_i(s_i, a_i) \mid s_t = s] \ .
\end{equation}
Moreover, the following lemma gives another form of $D^{\pi, \pi'}_t(s)$.

\begin{lemma}\label{thm:D-oper-trans}
\begin{equation*}
D^{\pi, \pi'}_t(s) = \E^\pi [\sum_{i = t}^\infty \Gamma^\tau_{t, i} A^\pi_i(s_i, a_i) {\pi'(a_i | s_i) \over \pi(a_i | s_i)} \mid s_t = s] \ .
\end{equation*}
\end{lemma}

The expression of $D^{\pi, \pi'}_t(s)$ in Lemma \ref{thm:D-oper-trans} closely resembles the policy gradient in Equation~\eqref{pol-grad-nvmdp}, differing only in the terms $\nabla \pi_i(a_i | s_i)$ and $\pi'_i(a_i | s_i)$.
Their relationship is analogous to that between a difference and a derivative in calculus.
By Equation \eqref{adv-nvmdp} and Equation \eqref{D-oper-def}, $D^{\pi, \pi}_t(s) = 0$.
Therefore,
\begin{equation*}
\begin{aligned}
D^{\pi, \pi'}_t(s) &= D^{\pi, \pi'}_t(s) - D^{\pi, \pi}_t(s) \\
&= \E^\pi [\sum_{i = t}^\infty \Gamma^\tau_{t, i} A^\pi_i(s_i, a_i) {{\pi'(a_i | s_i) - \pi(a_i | s_i)} \over \pi(a_i | s_i)} \mid s_t = s] \ .
\end{aligned}
\end{equation*}
If the old policy $\pi$ is fixed and the new policy $\pi'$ differs from the old policy $\pi$ by one parameter $\theta$ only, then
\begin{equation*}\phantomsection\label{D-oper-pol-grad}
\begin{aligned}[b]
{1 \over \Delta\theta} \bigl(D^{\pi, \pi'}_t(s) - D^{\pi, \pi}_t(s)\bigr) &= \E^\pi [\sum_{i = t}^\infty \Gamma^\tau_{t, i} A^\pi_i(s_i, a_i) {{\pi'(a_i | s_i) - \pi(a_i | s_i)} \over \pi(a_i | s_i) \Delta\theta} \mid s_t = s] \quad(\Delta\theta \text{ does not affect } \pi) \\
\Rightarrow {\partial \over \partial \theta} D^{\pi, \pi'}_t(s) &= \E^\pi [\sum_{i = t}^\infty \Gamma^\tau_{t, i} A^\pi_i(s_i, a_i) {{\partial \pi(a_i | s_i) / \partial \theta}  \over \pi(a_i | s_i)} \mid s_t = s] \quad(\text{Let } \Delta\theta \rightarrow 0) \\
\Rightarrow \nabla_{\pi'} D^{\pi, \pi'}_t(s) &= \E^\pi [\sum_{i = t}^\infty \Gamma^\tau_{t, i} A^\pi_i(s_i, a_i) {\nabla \pi(a_i | s_i) \over \pi(a_i | s_i)} \mid s_t = s] = \nabla V^\pi_t(s) \ .
\end{aligned}
\end{equation*}
Thus, when $\pi'$ is obtained by updating the parameters of $\pi$, the derivative of $D^{\pi, \pi'}_t(s)$ with respect to the parameters of $\pi'$ equals the policy gradient in Equation \eqref{pol-grad-nvmdp}.

The following theorem bounds the difference between $V^{\pi'}_t(s) - V^\pi_t(s)$ and $D^{\pi, \pi'}_t(s)$, with an additional Assumption \ref{asm:trpo-Gamma-bd}.

\begin{assumption}\label{asm:trpo-Gamma-bd}
There exists $M > 0$, such that
\begin{equation*}
\sum_{i = t}^\infty (i - t) \prod_{j = t + 1}^i \max_{s, a, s'} \gamma_j(s, a, s') < M \quad(\forall t \ge 0) \ .
\end{equation*}
\end{assumption}

\begin{theorem}\label{thm:prfm-diff-absbnd}
If a NVMDP satisfies Assumption \ref{asm:trpo-Gamma-bd}, then for any two policies $\pi, \pi'$ and state $s$:
\begin{equation}\phantomsection\label{prfm-diff-absbnd}
\bigl| V^{\pi'}_t(s) - V^\pi_t(s) - D^{\pi, \pi'}_t(s) \bigr| \le C \alpha_t^2(\pi, \pi') \ ,
\end{equation}
where $D^{\pi, \pi'}_t(s)$ is defined by Equation \eqref{D-oper-def},
C is some constant, and
\begin{equation}\phantomsection\label{tvdist-upb}
\alpha_t(\pi, \pi') = \max_{j \ge t, x \in S} D_{TV} \bigl( \pi_j(\cdot | x)\|\pi'_j(\cdot | x) \bigr) \ .
\end{equation}
($D_{TV}$ is the total variation distance.)
\end{theorem}

The derivation of the TRPO objective in NVMDPs using the above results follows nearly the same logic as in Schulman et al. \cite{TRPO-2015}.
By Theorem \ref{thm:prfm-diff-absbnd}, for a given state distribution $s_t \sim \rho$,
\begin{equation}\phantomsection\label{prfm-diff-lwb-startdistr}
\E^{s_t \sim \rho} [V^{\pi'}_t(s_t) - V^\pi_t(s_t)] \ge \E^{s_t \sim \rho} [D^{\pi, \pi'}_t(s_t)] - C \alpha_t^2(\pi, \pi') \ .
\end{equation}
Since $D^{\pi, \pi}_t(s) = 0$ and $\alpha_t(\pi, \pi) = 0$, the R.H.S of Inequality \eqref{prfm-diff-lwb-startdistr} is 0 when $\pi' = \pi$.
For an old policy $\pi$, if the new policy $\pi'$ maxmizes the R.H.S of Inequality \eqref{prfm-diff-lwb-startdistr}, then
\begin{equation*}
\E^{s_t \sim \rho} [V^{\pi'}_t(s_t) - V^\pi_t(s_t)] \ge \max_{\bar{\pi}} \bigl\{\E^{s_t \sim \rho} [D^{\pi, \bar{\pi}}_t(s_t)] - C \alpha_t^2(\pi, \bar{\pi}) \bigr\} \ge 0 \ ,
\end{equation*}
which guarantees $\E^{s_t \sim \rho} [V^{\pi'}_t(s_t)] \ge \E^{s_t \sim \rho} [V^\pi_t(s_t)]$.

If we start from an initial policy and iteratively update the policy using
$\pi' \in \text{argmax}_{\bar{\pi}} \bigl\{\E^{s_t \sim \rho} [D^{\pi, \bar{\pi}}_t(s_t)] - C \alpha_t^2(\pi, \bar{\pi}) \bigr\}$,
we obtain a sequence of policies that is non-decreasing with respect to the objective $\E^{s_t \sim \rho} [V^{\pi}_t(s_t)]$.
This optmizing task can be approximated in a trust region style by constraining $\alpha_t(\pi, \pi')$ with a small threshold:
\begin{equation*}
 \max_{\pi'} \E^{s_t \sim \rho} [D^{\pi, \pi'}_t(s_t)] \quad\text{s.t. } \alpha_t(\pi, \pi') \le \delta \ ,
\end{equation*}
where $\delta > 0$ is a pre-determined threshold controlling the allowed deviation in $\alpha_t(\pi, \pi')$ (Conn et al. \cite{Trust-Region-2000}).
By Lemma \ref{thm:D-oper-trans}, the task becomes
\begin{equation*}
\max_{\pi'} \E^{s_t \sim \rho, \ a_t, s_{t + 1}, ... \sim \pi} [\sum_{i = t}^\infty \Gamma^\tau_{t, i} A^\pi_i(s_i, a_i) {\pi'(a_i | s_i) \over \pi(a_i | s_i)}] \quad\text{s.t. } \alpha_t(\pi, \pi') \le \delta \ .
\end{equation*}

A noteworthy point for applying TRPO in NVMDPs is that the policy improvement bound in Theorem \ref{thm:prfm-diff-absbnd} requires Assumption \ref{asm:trpo-Gamma-bd} in addition to Assumption \ref{asm:nonneg-gamma-nvmdp}, \ref{asm:rwd-absbnd-nvmdp}, \ref{asm:Gamma-sum-absbnd-nvmdp}, and \ref{asm:gamma-bellman-cond-nvmdp} presented in Section \ref{sec:nvmdp-fda}.
For the TRPO algorithm, rollouts of episodes are first collected before a policy update is performed.
As discussed in Section \ref{sec:spcl-cases-nvmdp}, in episodic RL tasks we can view $\gamma_{t+1}(\cdot) = 0$ for all $t \ge H-1$, where $H$ is the (implicit) horizon.
Thus, Assumption \ref{asm:trpo-Gamma-bd} is naturally satisfied, effectively reducing the requirement on the discount rates to non-negativity when employing TRPO in NVMDPs.


\section{Experiments}\label{sec:expt}

In this section, we first introduce \textbf{Tricky Gridworld}---a non-stationary gridworld environment.
We then present experimental results obtained using the control algorithms (i.e., algorithms seeking optimal policies) described in Section~\ref{sec:tab-solu}.
These results illustrate how these algorithms perform under non-stationarity and how the resulting trajectories are influenced by different reward and discount rate settings.

\subsection{The Environment}\label{sec:tricky-gwd-intro}

Gridworlds are classic MDPs on 2-D grids where an agent moves \textit{up}, \textit{down}, \textit{left}, or \textit{right} each time step, shifting one cell or staying put if the move exceeds grid boundaries.
The goal is typically to reach a designated cell in the fewest steps.

\begin{figure}[htbp]
\centerline{\includegraphics[width=.83\linewidth]{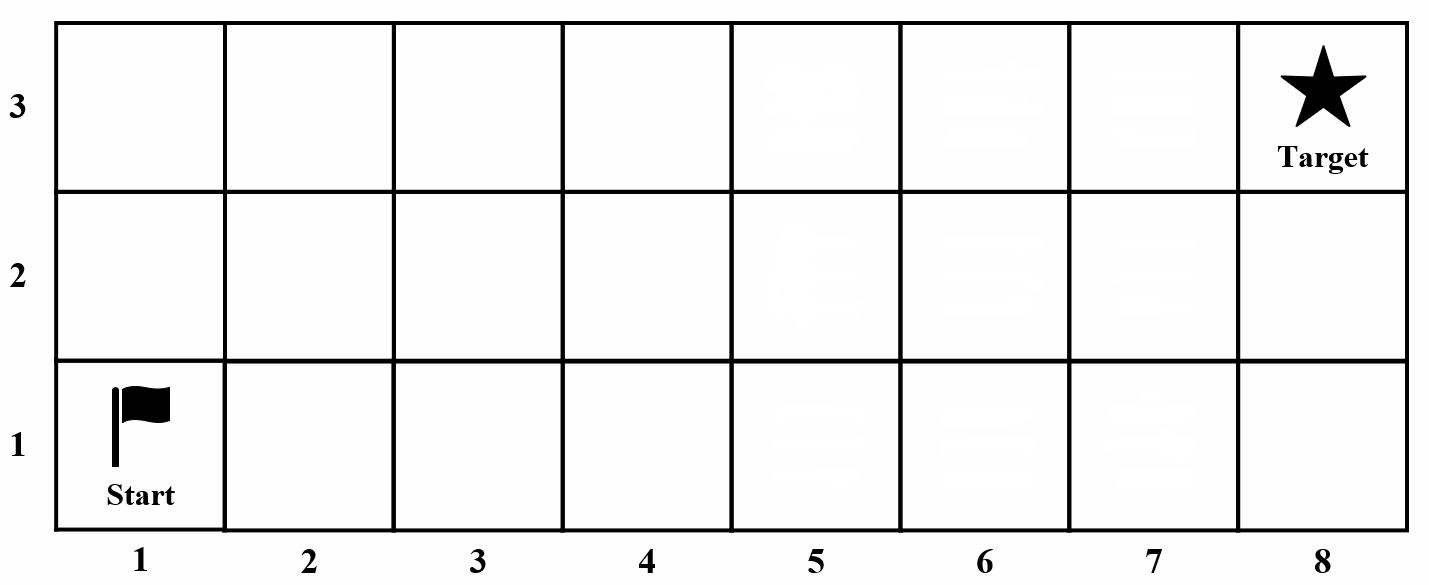}}
\caption{A Vanilla Gridworld}
\phantomsection\label{fig:van-gwd}
\end{figure}

Figure \ref{fig:van-gwd} shows a vanilla gridworld of size $8 \times 3$, where each cell is labeled by its horizontal (1–8) and vertical (1–3) coordinates.
Here, each state corresponds to the robot's current position, and the agent aims to reach the target state (8, 3) from the start state (1, 1) as soon as possible.
Each move incurs a reward of -10, and a constant discount factor $\gamma = 0.999$ encourages shorter paths.
A policy that maximizes state-values under this reward and discount rate structure is therefore optimal.

\begin{figure}[htbp]
\centerline{\includegraphics[width=.83\linewidth]{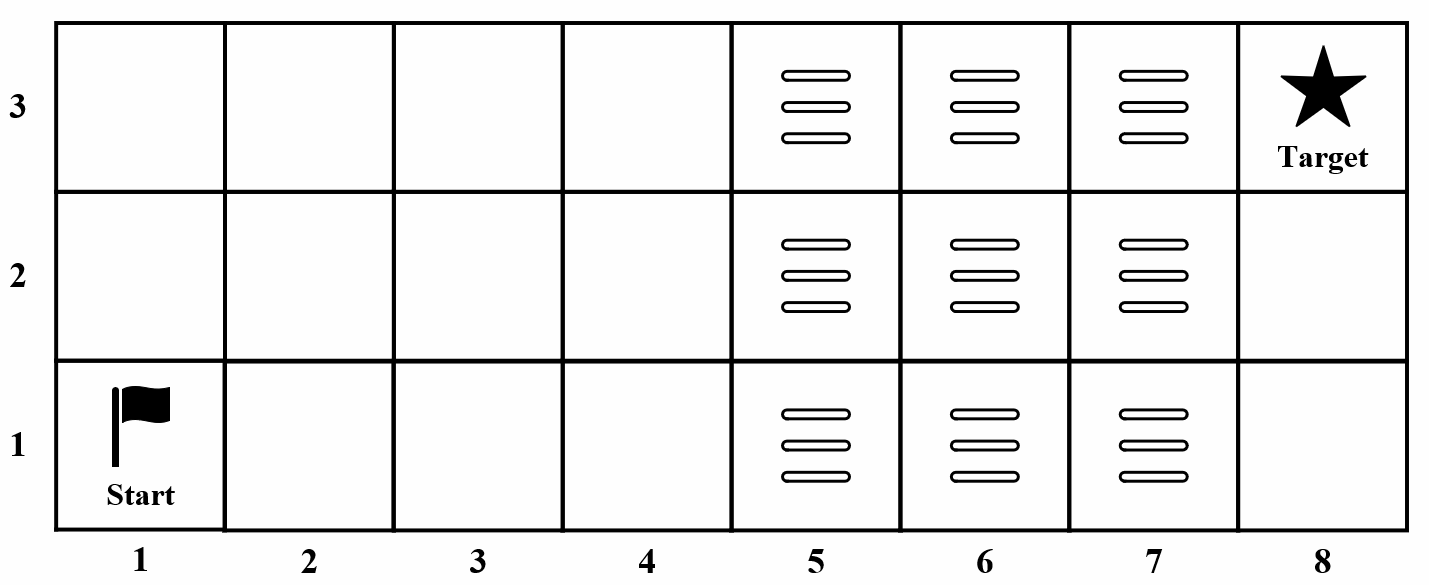}}
\caption{Tricky Gridworld}
\phantomsection\label{fig:tricky-gwd}
\end{figure}

Tricky Gridworld in Figure \ref{fig:tricky-gwd} shares the same state and action spaces as the vanilla version but introduces non-stationary dynamics.
In Tricky Gridworld, cells with horizontal coordinate $x \in \{5, 6, 7\}$ contain vent-like strips that periodically emit strong winds depending on time $t$: \\
(i) For $x=5$, wind pushes the robot 2 cells left, except when $t \bmod 6 = 0$; \\
(ii) For $x=6$, wind pushes 3 cells left, except when $t \bmod 6 \in \{1, 2, 3, 4\}$; \\
(iii) For $x=7$, wind pushes 4 cells left, except when $t \bmod 6 = 5$. \\
When the robot is on a cell with active wind, it is first blown leftward before executing its selected action.

Tricky Gridworld has time horizon 200 for each episode.
The reward is noisy:
\begin{equation}\phantomsection\label{tricky-gwd-r-lvn}
r = -10 + 3.0398 z \quad(z \sim \mathcal{N}(0,1) \text{ i.i.d.}) \ ,
\end{equation}
which has a 90\% confidence interval of [-15, -5].
Moreover, we examine three discount rate configurations, denoted \textbf{dr-0} through \textbf{dr-2}, and discuss their effects subsequently: \\
(0) $\gamma_{t + 1}(s, a, s') = 0.999$ for all $t, s, a, s'$; \\
(1) $\gamma_{t + 1}\bigl(s, a, s') = 1.02$ if $s' \in \{(3, 1), (4, 2)\}$, otherwise $\gamma_{t + 1}(\cdot) = 0.999$; \\
(2) $\gamma_{t + 1}\bigl(s, a, s') = 1.02$ for $t < 50, s' \in \{(3, 1), (4, 2)\}$, otherwise $\gamma_{t + 1}(\cdot) = 0.999$.


\subsection{Results of Model-based Method}

With the dynamics known, we apply a model-based method---DP Value Iteration (Algorithm \ref{algo:dp-valiter-nvmdp})---to find optimal trajectories.
Given that $\E[r] = -10$, we set $r = -10$ in the algorithm.

\begin{figure}[htbp]
\centerline{\includegraphics[width=.83\linewidth]{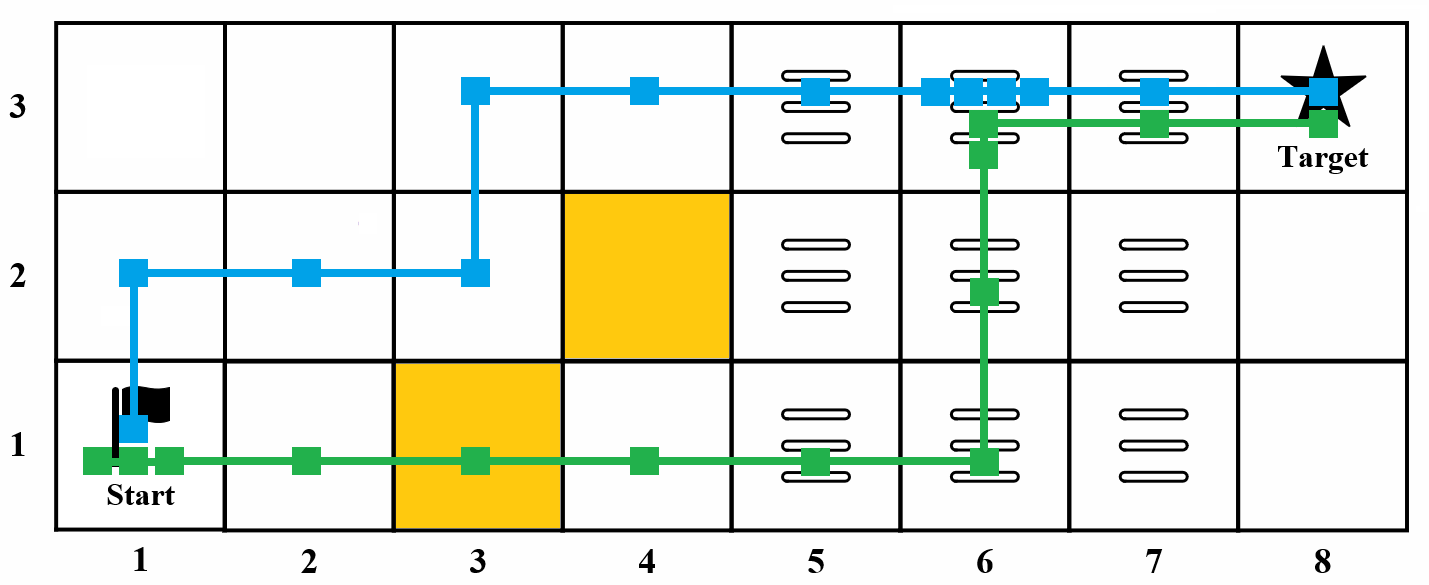}}
\caption{Trajectories on Tricky Gridworld}
\phantomsection\label{fig:tricky-gwd-traj}
\end{figure}

Figure \ref{fig:tricky-gwd-traj} displays the optimal trajectories obtained.
Here, the green line corresponds to dr-0, and the blue line corresponds to dr-1 and dr-2.
Each square represents the robot's position at a timestep; multiple squares on the same cell indicate intentional staying.
Both trajectories reach the target in 12 steps.
Since $\E[r] < 0$ at each step, dr-1 and dr-2 encourage the robot to avoid cells (3, 1) and (4, 2) (marked orange in Figure \ref{fig:tricky-gwd-traj}), producing the blue trajectory.


\subsection{Results of Model-free Methods}\label{sec:tricky-gwd-mfres}

When the dynamics are unknown, optimal policies are obtained via model-free methods using the Tricky Gridworld simulator.
Throughout this subsection, all algorithms employ $\varepsilon$-greedy exploration with $\varepsilon = 0.05$ and a constant step size of 0.1.
Action-value estimates are evaluated every 500 episodes using a policy greedy with respect to the current estimates.
Convergence to optimality is regarded as successfully reaching the target state from the start state in 12 steps (return $\approx$ -119) and maintaining the trajectory in 12 steps in subsequent updates.

\begin{figure}[htbp]
\centerline{\includegraphics[width=.85\linewidth]{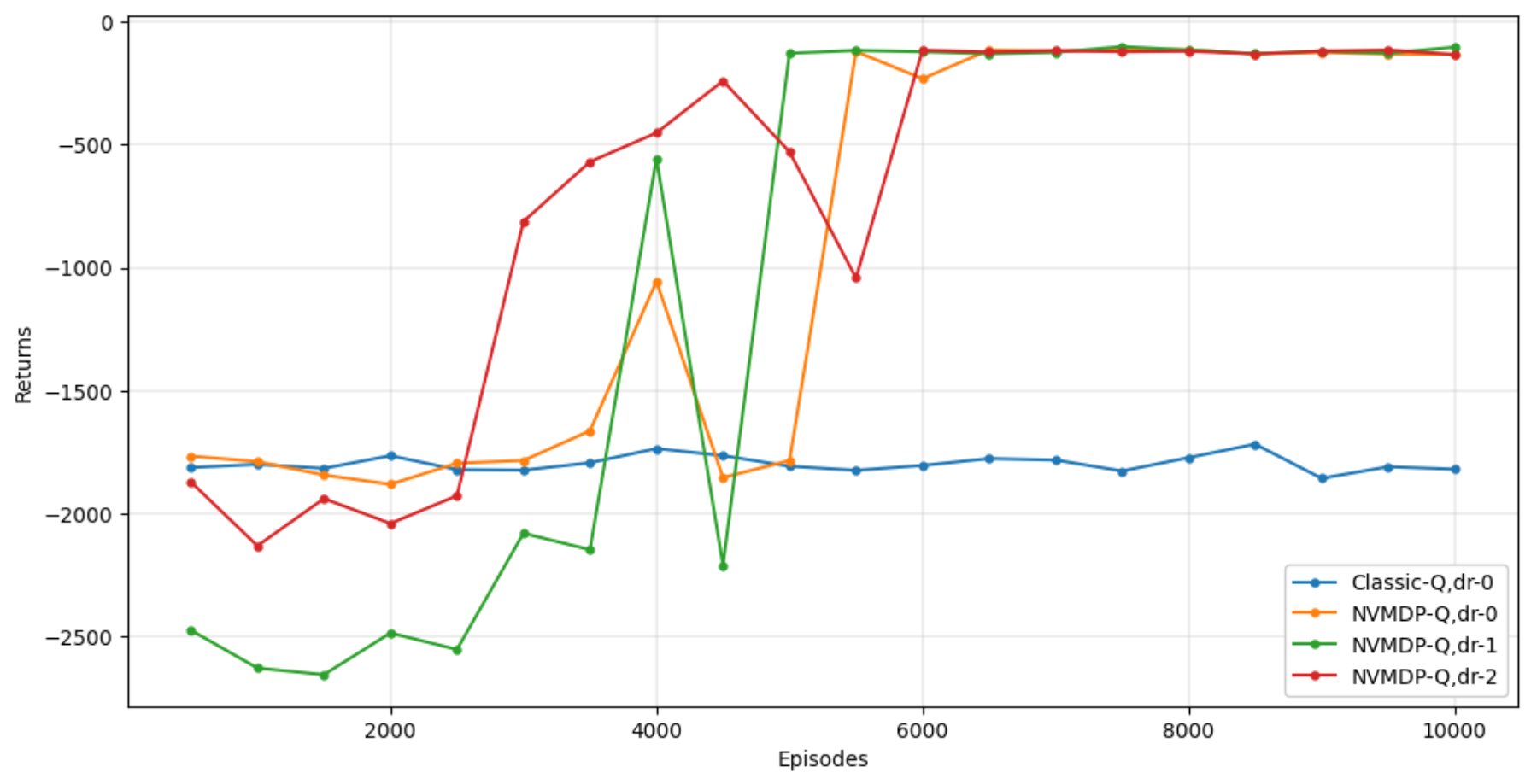}}
\caption{Performance of Classic Q-learning and NVMDP-Q-learning}
\phantomsection\label{fig:trickgwd-cmp-clascvsnvdmp}
\end{figure}

We begin by comparing classic Q-learning and NVMDP-Q-learning (Section \ref{sec:q-learn}) to evaluate their ability to handle non-stationarity.
Classic Q-learning is applied under dr-0, whereas NVMDP-Q-learning is tested under dr-0 through dr-2.
Figure \ref{fig:trickgwd-cmp-clascvsnvdmp} presents the results of a single trial for each algorithm, where the horizontal and vertical axes display the number of episodes and returns, respectively.

As shown in Figure \ref{fig:trickgwd-cmp-clascvsnvdmp}, classic Q-learning (blue line) fails to improve since the stationarity assumption of classic MDPs is violated.
Aside from random $\varepsilon$-greedy exploration, classic Q-learning seeks a single greedy action per state.
However, success in Tricky Gridworld requires time-dependent behavior---e.g., staying at horizontal coordinate 6 for four steps and moving right only when $t \bmod 6 = 5$.
Without time-awareness, learning such a strategy is practically impossible for classic Q-learning.
The robot either avoids moving right or moves at the wrong time and is blown away.

In contrast, NVMDP-Q-learning achieves convergence to optimality across dr-0 through dr-2.
While uniform discounting in dr-0 (orange line in Figure \ref{fig:trickgwd-cmp-clascvsnvdmp}) yields relatively stable learning, dr-1 (green line) encourages avoidance of cells (3, 1) and (4, 2), at the cost of higher return variance.
dr-2 (red line) confines this effect to the time window [0, 50), thereby reducing variance.

\begin{figure}[htbp]
\centerline{\includegraphics[width=.85\linewidth]{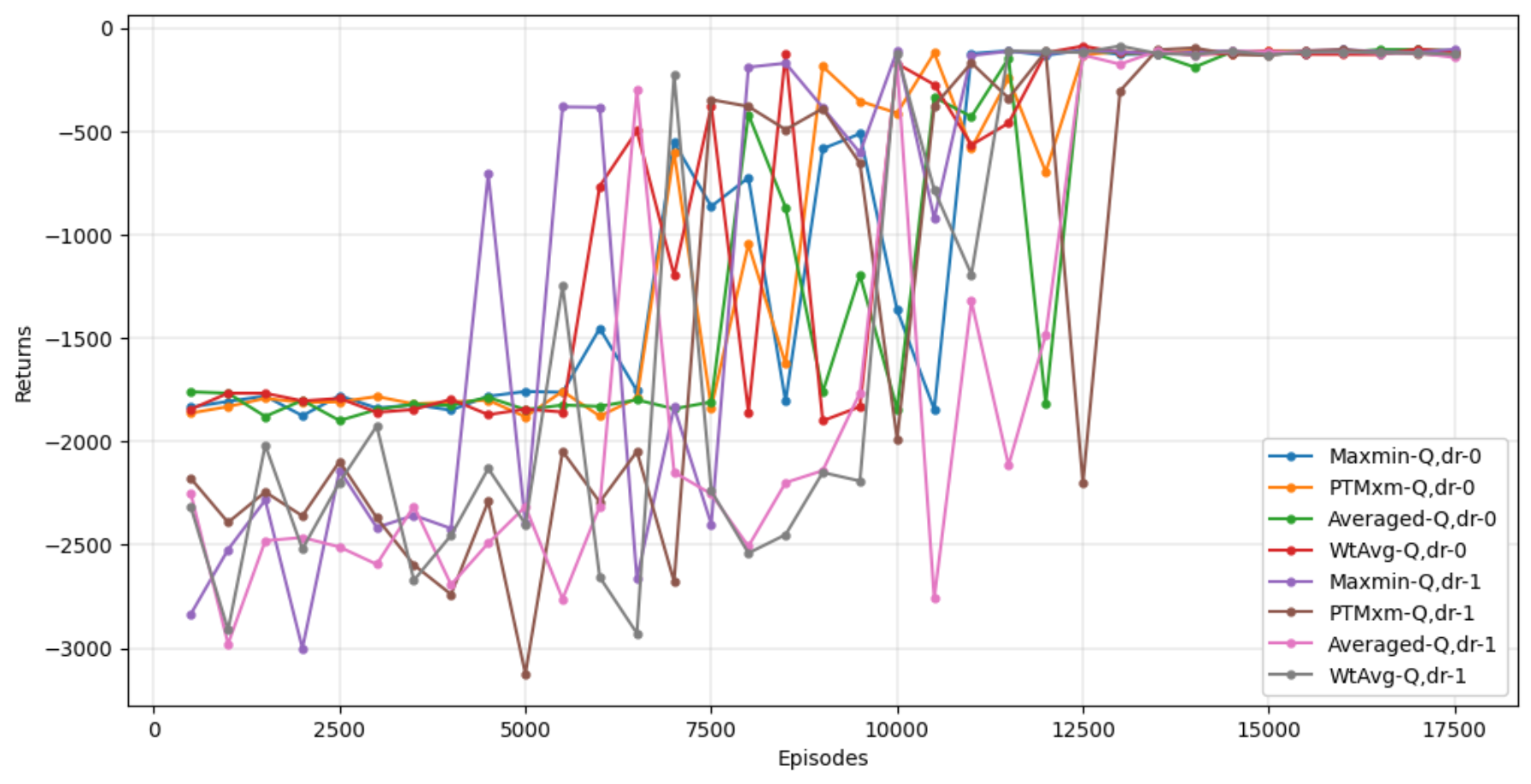}}
\caption{Performance of Generalized Q-learning Algorithms ($n = 2, l = 3$)}
\phantomsection\label{fig:trickgwd-cmp-geqalgr}
\end{figure}

For the generalized Q-learning variants (Section \ref{sec:geq-learn}), Figure \ref{fig:trickgwd-cmp-geqalgr} presents single-trial results for each method.
All variants use estimate track sizes $n = 2, l = 3$, with $\lambda = 0.5$ and $\eta = 0.7$ for WtAvg-Q.
Both dr-0 and dr-1 are tested.
According to Figure \ref{fig:trickgwd-cmp-geqalgr}, all algorithms converge to optimality and exhibit higher learning variance under dr-1 due to the avoidance effect.

It is noteworthy that higher discount rates do not necessarily guarantee the avoidance effect when $E[r] < 0$.
This effect becomes more pronounced with smaller variance in reward noise and weakens as the noise variance increases.
To investigate this, we define the following reward:
\begin{equation}\phantomsection\label{tricky-gwd-r-svn}
r = -10 + 0.6080 z, \quad z \sim \mathcal{N}(0,1) \text{ i.i.d.} \ ,
\end{equation}
which yields a 90\% confidence interval of $[-11, -9]$.
We refer to the original reward in Equation \eqref{tricky-gwd-r-lvn} and the above reward in Equation \eqref{tricky-gwd-r-svn} as \textbf{r-lvn} and \textbf{r-svn}, respectively.
In addition, we introduce another discount rate configuration, denoted as \textbf{dr-3}: \\
(3) $\gamma_{t + 1}\bigl(s, a, s') = 1.05$ for $t < 50, s' \in \{(3, 1), (4, 2)\}$, otherwise $\gamma_{t + 1}(\cdot) = 0.999$. \\
Compared to dr-2, dr-3 assigns higher discount rates to transitions toward cell $(3, 1)$ or $(4, 2)$ for $0 \le t < 50$, thereby strengthening the avoidance effect under larger variance in reward noise.

Table \ref{tbl:trickgwd-q-learn-res} in Appendix \ref{sec:apndx-long-table} presents the comprehensive results of the model-free algorithms analyzed in this study, where r-lvn is evaluated under discount rates dr-0 to dr-3, while r-svn is tested with dr-1 and dr-2.
The results indicate that NVMDP-Q-learning and the generalized Q-learning variants with $n = 1$ exhibit comparatively rapid convergence across all combinations of reward settings and discount rate configurations.
Nevertheless, for rewards with larger noise variance (i.e., under r-lvn), the avoidance effects observed under dr-1 and dr-2 are not consistently preserved.
Generally, increasing the estimate track row number $n$ (while maintaining the overall size $n \times l$) tends to reinforce the avoidance effect (except for PTMxm-Q), albeit at the cost of slower convergence to optimality.
Alternatively, the enhancement can be achieved by switching the discount rate configuration from dr-2 to dr-3, though at the expense of higher learning variance.


\section{Conclusions and Future Work}\label{sec:concl}

MDPs are the theoretical foundation of RL.
In this work, we concisely outline, explain, and compare several categories of MDPs, including Finite MDPs (FMDPs), finite- and infinite-horizon MDPs, and stationary versus non-stationary MDPs.

To address the limitations of algorithms developed under stationary MDP assumptions when applied to non-stationary environments, and to generalize the commonly uniform discounting scheme in RL to a non-uniform form analogous to those used in finance, we propose the NVMDP framework.
The NVMDP framework inherently accommodates non-stationarity, with discount rates varying with both time and transitions which enables explicit shaping of optimal policy trajectories.

We establish the theoretical foundations of NVMDPs, including value function formulation and recursion, matrix representation, optimality conditions, and policy improvement under finite state and action spaces with mild assumptions.
For practical tasks with finite horizons, we argue that these assumptions are naturally satisfied.
Furthermore, we show that classic MDPs are special cases of NVMDPs for the objective of identifying an optimal policy, and that the NVMDP framework unifies both finite- and infinite-horizon formulations.

For tabular methods, we extend two fundamental DP approaches---policy evaluation and value iteration---to the NVMDP setting.
We adapt the Q-learning algorithm by Watkins \cite{Watkins-PhD-1989} and extend the generalized Q-learning framework of Lan et al. \cite{Maxmin-q-learn-2020} to NVMDPs, introducing several variants: NVMDP-Q-learning, Maxmin Q-learning, Averaged Q-learning, PTMxm-Q, and WtAvg-Q.
Convergence of these variants is established under the adapted generalized Q-learning framework for NVMDPs.

For RL with function approximation, we extend both the Policy Gradient Theorem by Sutton et al. \cite{Sutton-PolGrad-1999} and the theoretical foundation of TRPO by Schulman et al. \cite{TRPO-2015} to the NVMDP setting.
Two proofs are provided for each result: one in scalar form and another in matrix representation.

To empirically evaluate the NVMDP framework, we design a non-stationary Tricky Gridworld environment incorporating multiple reward and discount rate settings.
Our experimental results show that the proposed tabular algorithms effectively handle non-stationarity and successfully recover optimal trajectories, whereas the original Q-learning algorithm fails in the absence of time-awareness.
We further illustrate how varying discount rates influence trajectory shaping, i.e., encouraging the avoidance of specific states in our experiments.

In summary, we demonstrate that the NVMDP framework is both theoretically sound and empirically effective.
While our theoretical results extend those of classic MDPs, the necessary modifications to algorithmic design are relatively minor, yet they yield substantial benefits in handling non-stationarity and enabling optimal trajectory shaping.
We believe that the NVMDP framework provides a promising foundation for generalizing further results from existing MDP theories and inspiring new research directions.

Due to time constraints, we were unable to extend additional tabular solution methods to the NVMDP framework in the current version of this work.
Furthermore, we did not address the theoretical foundations of convergence guarantees under function approximation, nor did we present algorithms employing function approximators. Additionally, we have not yet explored NVMDPs in continuous state or action spaces, scalable algorithms for large-scale non-stationary environments, or formulations incorporating risk-aware RL.
These aspects remain open directions for future exploration.
We are currently investigating these issues and plan to present more comprehensive theoretical analyses, extended algorithmic developments, and empirical evaluations in future work.


\section*{Acknowledgments}

An earlier version of this work was submitted as a course project for the Stochastic Processes course, instructed by Prof. Cathy H. Xia in Autumn 2024, whose suggestions were invaluable. We also thank Prof. Abhishek Gupta for reviewing a recent version of the manuscript and providing insightful feedback.



\pagebreak
\appendix
\renewcommand{\thetheorem}{A.\arabic{theorem}}
\setcounter{theorem}{0}
\renewcommand{\theassumption}{A.\arabic{assumption}}
\setcounter{assumption}{0}
\section{Appendix}

\subsection{Proofs}\label{sec:apndx-proofs}

\begin{manuallemma}{\ref{thm:sv-av-nvmdp}}
State-values and action-values in an NVMDP satisfy:
\begin{equation}
V^{\pi}_t(s_t) = \sum_{a_t} \pi_t(a_t | s_t) Q^{\pi}_t(s_t, a_t) \ , \tag{\ref{sv-av-nvmdp}}
\end{equation}
\begin{equation}
Q^{\pi}_t(s_t, a_t) = \sum_{s_{t + 1}, r_t} p_t(s_{t + 1}, r_t | s_t, a_t) (r_t(s_t, a_t, s_{t + 1}) + \gamma_{t + 1}(s_t, a_t, s_{t + 1}) V^{\pi}_{t + 1}(s_{t + 1})) \ . \tag{\ref{av-iter-nvmdp}}
\end{equation}
\end{manuallemma}

\noindent\textbf{Proof:} 
\begin{equation*}
\begin{aligned}
V^{\pi}_t(s) &= \sum_g gP(G_t(\tau_t) = g \mid s_t = s) \quad(\text{by Eq. \eqref{sv-nvmdp}}) \\
&= \sum_g {gP(G_t(\tau_t) = g, s_t = s) \over P(s_t = s)} \\
&= \sum_{g, a} {gP(G_t(\tau_t) = g, s_t = s, a_t = a) \over P(s_t = s)} \\
&= \sum_{g, a} gP(G_t(\tau_t) = g \mid s_t = s, a_t = a){P(s_t = s, a_t = a) \over P(s_t = s)} \\
&= \sum_{g, a} gP(G_t(\tau_t) = g \mid s_t = s, a_t = a) \pi_t(a|s) \\
&= \sum_a \pi_t(a|s) \sum_g gP(G_t(\tau_t) = g \mid s_t = s, a_t = a) \\
&= \sum_a \pi_t(a|s) Q^{\pi}_t(s, a) \quad (\text{by Eq. \eqref{av-nvmdp}}).
\end{aligned}
\end{equation*}

Equation \eqref{sv-av-nvmdp} is thus proven.
For Equation \eqref{av-iter-nvmdp},
\begin{equation*}
\begin{aligned}
Q^\pi_t(s, a) &= \E^\pi[\sum_{i = t}^{\infty}\Gamma^\tau_{t, i}r_i(s_i, a_i, s_{i + 1}) \mid s_t = s, a_t = a] \quad (\text{by Eq. \eqref{return-nvmdp}\eqref{av-nvmdp}}) \\
&= \E^\pi[r_t(s_t, a_t, s_{t + 1}) + \sum_{i = t + 1}^{\infty}\Gamma^\tau_{t, i}r_i(s_i, a_i, s_{i + 1}) \mid s_t = s, a_t = a] \quad (\Gamma^\tau_{t, t} = 1).
\end{aligned}
\end{equation*}

The first item in R.H.S is:
\begin{equation*}
\begin{aligned}
&\E^\pi[r_t(s_t, a_t, s_{t + 1}) \mid s_t = s, a_t = a] = \E^{s_{t + 1}, r_t \sim p_t(\cdot \mid s, a)}[r_t(s, a, s_{t + 1})] \\
&= \E^{s_{t + 1} \sim s, a}[\E^{r_t \sim s_{t + 1}, s, a}[r_t(s, a, s_{t + 1}) \mid s_{t + 1}]] \\
&= \sum_{s_{t + 1}} p(s_{t + 1} | s, a) \sum_{r_t} r_t p(r_t | s_{t + 1}, s, a) \\
&= \sum_{s_{t + 1}, r_t} r_t p(s_{t + 1}, r_t | s, a) \ .
\end{aligned}
\end{equation*}

The second item in R.H.S is:
\begin{equation*}
\begin{aligned}
&\E^\pi[\sum_{i = t + 1}^{\infty}\Gamma^\tau_{t, i}r_i(s_i, a_i, s_{i + 1}) \mid s_t = s, a_t = a] \\
&= \E^{s_{t + 1} \sim s, a, \ a_{t + 1}, r_{t + 1}, s_{t + 2} ... \sim \pi}[\gamma_{t + 1}(s, a, s_{t + 1}) \sum_{i = t + 1}^{\infty}\Gamma^\tau_{t + 1, i}r_i(s_i, a_i, s_{i + 1})] \\
&= \E^{s_{t + 1} \sim s, a} \Bigl[ \E^{a_{t + 1}, r_{t + 1}, s_{t + 2} ... \sim \pi}[\gamma_{t + 1}(s, a, s_{t + 1}) \sum_{i = t + 1}^{\infty}\Gamma^\tau_{t + 1, i}r_i(s_i, a_i, s_{i + 1}) \mid s_{t + 1}] \Bigr] \\
&= \E^{s_{t + 1} \sim s, a} \Bigl[ \gamma_{t + 1}(s, a, s_{t + 1}) \E^{a_{t + 1}, r_{t + 1}, s_{t + 2} ... \sim \pi} [\sum_{i = t + 1}^{\infty}\Gamma^\tau_{t + 1, i}r_i(s_i, a_i, s_{i + 1}) \mid s_{t + 1}] \Bigr] \\
&= \E^{s_{t + 1} \sim s, a}[\gamma_{t + 1}(s, a, s_{t + 1}) V^{\pi}_{t + 1}(s_{t + 1})] \quad (\text{by Eqs. \eqref{return-nvmdp} and \eqref{sv-nvmdp}}) \\
&= \sum_{s_{t + 1}, r_t} p(s_{t + 1}, r_t | s, a) \gamma_{t + 1}(s, a, s_{t + 1}) V^{\pi}_{t + 1}(s_{t + 1}) \ .
\end{aligned}
\end{equation*}

Combine the results above established Equation \eqref{av-iter-nvmdp}.
$\blacksquare$

\begin{manuallemma}{\ref{thm:sv-av-avgrwd-nvmdp}}
With $\bar{r}_t(s, a)$ defined by Equation \eqref{avg-rwd-nvmdp},
\begin{equation*}
\begin{aligned}
V^\pi_t(s) &= \E^\pi [\sum_{i = t}^{\infty}\Gamma^\tau_{t, i} \bar{r}_i(s_i, a_i) \mid s_t = s] \\
Q^\pi_t(s, a) &= \E^\pi [\sum_{i = t}^{\infty}\Gamma^\tau_{t, i} \bar{r}_i(s_i, a_i) \mid s_t = s, a_t = a] \\
\end{aligned}
\end{equation*}
\end{manuallemma}

\noindent\textbf{Proof:}
By Equation \eqref{return-nvmdp} and Equation \eqref{sv-nvmdp}:
\begin{equation*}
\begin{aligned}
& V^\pi_t(s) = \E^\pi [\sum_{i = t}^{\infty}\Gamma^\tau_{t, i} r_i(s_i, a_i, s_{i + 1}) \mid s_t = s] = \sum_{i = t}^{\infty} \E^\pi [\Gamma^\tau_{t, i} r_i(s_i, a_i, s_{i + 1}) \mid s_t = s] \\
&= \sum_{i = t}^{\infty} \E^{a_t, s_{t + 1}, ..., s_i, a_i \sim \pi} \Bigl[\E^{s_{i + 1}}[\Gamma^\tau_{t, i} r_i(s_i, a_i, s_{i + 1}) \mid s_i, ..., s_t] \mid s_t = s \Bigr] \\
&= \sum_{i = t}^{\infty} \E^{a_t, s_{t + 1}, ..., s_i, a_i \sim \pi} [\Gamma^\tau_{t, i} \bar{r}_i(s_i, a_i) \mid s_t = s] \quad(\text{by Eq. \eqref{avg-rwd-nvmdp}}) \\
&= \E^\pi [\sum_{i = t}^{\infty}\Gamma^\tau_{t, i} \bar{r}_i(s_i, a_i) \mid s_t = s] \ .
\end{aligned}
\end{equation*}

By the above result and Equation \eqref{av-iter-nvmdp-e},
\begin{equation*}
\begin{aligned}
&Q^{\pi}_t(s, a) = \bar{r}_t(s, a) + \E^{s' \sim p_t(\cdot | s, a)}\big[\gamma_{t + 1}(s, a, s') \E^\pi[\sum_{i = t + 1}^{\infty}\Gamma^\tau_{t + 1, i} \bar{r}_i(s_i, a_i) \mid s_{t + 1} = s']\mid s_t = s, a_t = a\big] \\
&= \bar{r}_t(s, a) + \E^{s' \sim p_t(\cdot | s, a)} \big[\E^\pi[\sum_{i = t + 1}^{\infty} \gamma_{t + 1}(s, a, s')\Gamma^\tau_{t + 1, i} \bar{r}_i(s_i, a_i) \mid s_{t + 1} = s', s_t = s, a_t = a]\mid s_t = s, a_t = a\big] \\
&= \bar{r}_t(s, a) + \E^\pi [\sum_{i = t + 1}^{\infty}\Gamma^\tau_{t, i} \bar{r}_i(s_i, a_i) \mid s_t = s, a_t = a] = \E^\pi [\sum_{i = t}^{\infty}\Gamma^\tau_{t, i} \bar{r}_i(s_i, a_i) \mid s_t = s, a_t = a] \ .\ \blacksquare
\end{aligned}
\end{equation*}

\begin{manuallemma}{\ref{thm:val-cns-nvmdp}}
Policy $\pi$ and $\pi'$ are identical for all time $t \ge n$, then
\begin{equation*}
\begin{aligned}
V^{\pi'}_t(s) &= V^\pi_t(s) \quad(\forall t \ge n, s \in S) \\
Q^{\pi'}_t(s, a) &= Q^\pi_t(s, a) \quad(\forall t \ge n - 1, s \in S, a \in A)
\end{aligned}
\end{equation*}
\end{manuallemma}

\noindent\textbf{Proof:}
By definition of matrix $\bm{K}^\pi_{t + 1}$ in Equation \eqref{k-def-nvmdp-mat}:
\begin{equation}
(\bm{K}^\pi_{t + 1})_{(s, a), (s', a')} = \gamma_{t + 1}(s, a, s') p_t(s' | s, a)\pi_{t + 1}(a' | s') \ .
\end{equation}
Thus $\forall t \ge n - 1, \ \bm{K}^{\pi'}_{t + 1} = \bm{K}^\pi_{t + 1}$ since $\pi'_t = \pi_t$ for all $t \ge n$.

Therefore by Equation \eqref{av-selfiter-nvmdp-mat}, for any time $ t \ge n - 1$,
\begin{equation}\phantomsection\label{qt-inequ-pol-imprv}
\begin{aligned}[b]
\bm{Q}^{\pi'}_t - \bm{Q}^\pi_t &= \bm{K}^\pi_{t + 1} (\bm{Q}^{\pi'}_{t + 1} - \bm{Q}^{\pi}_{t + 1}) \\
\|\bm{Q}^{\pi'}_t - \bm{Q}^\pi_t\|_\infty &= \|\bm{K}^\pi_{t + 1} (\bm{Q}^{\pi'}_{t + 1} - \bm{Q}^\pi_{t + 1})\|_\infty \\
&\le \|\bm{K}^\pi_{t + 1}\|_\infty\|\bm{Q}^{\pi'}_{t + 1} - \bm{Q}^\pi_{t + 1}\|_\infty \ ,
\end{aligned}
\end{equation}

where
\begin{equation}\phantomsection\label{kt-upb}
\begin{aligned}[b]
\|\bm{K}^\pi_{t + 1}\|_\infty &= \max_{s, a} \sum_{s', a'} \gamma_{t + 1}(s, a, s') p_t(s' | s, a)\pi_{t + 1}(a' | s') \\
&= \max_{s, a} \sum_{s'} \gamma_{t + 1}(s, a, s') p_t(s' | s, a) \\
&\le \max_{s, a} \sum_{s'} p_t(s' | s, a) \max_{\bar{s}} \gamma_{t + 1}(s, a, \bar{s}) \\
&\le \max_{s, a} \max_{\bar{s}} \gamma_{t + 1}(s, a, \bar{s}) \\
&= \max_{s, a, s'} \gamma_{t + 1}(s, a, s') \ .
\end{aligned}
\end{equation}

Combine Inequality \eqref{qt-inequ-pol-imprv} with Inequality \eqref{kt-upb} and use the result iteratively,
\begin{equation*}
\|\bm{Q}^{\pi'}_t - \bm{Q}^\pi_t\|_\infty \le \Bigl( \prod^m_{j = t + 1} \max_{s, a, s'} \gamma_j(s, a, s') \Bigr)\|\bm{Q}^{\pi'}_m - \bm{Q}^\pi_m\|_\infty \quad(\forall t \ge n - 1, m \ge t + 1) \ .
\end{equation*}

By Assumption \ref{asm:gamma-bellman-cond-nvmdp} and the boundness of $Q^\pi_t(\cdot)$, $\bm{Q}^{\pi'}_t = \bm{Q}^\pi_t$ for all $t \ge n - 1$.
Furthermore by Equation \eqref{sv-av-nvmdp-mat}, $\bm{V}^{\pi'}_t = \bm{V}^\pi_t$ for all $t \ge n$. $\blacksquare$

\begin{manualtheorem}{\ref{thm:optim-pol-nvmdp}}
Consider $\Pi = \{(\pi_0, \pi_1, \pi_2, ...) \mid \pi_i: S \rightarrow \mathrm{Dist}(A) \ , \forall i \ge 0\}$.
Denote the optimal state-values and action-values at time $t$ as:
\begin{equation*}
\begin{aligned}
V^*_t(s) &= \sup_{\pi \in \Pi} V^{\pi}_t(s) \ , \\
Q^*_t(s, a) &= \sup_{\pi \in \Pi} Q^{\pi}_t(s, a) \ .
\end{aligned}
\end{equation*}
Then there exists a deterministic policy $\pi^* \in \Pi$ which is optimal:
\begin{equation*}
\begin{aligned}
V^{\pi^*}_t(s) &= V^*_t(s) \quad(\forall t \ge 0), \\
Q^{\pi^*}_t(s, a) &= Q^*_t(s, a) \quad(\forall t \ge 0).
\end{aligned}
\end{equation*}
(A deterministic policy selects one action w.p. 1 in every state at every time.)
\end{manualtheorem}

\noindent\textbf{Proof:}
Since $V^\pi_t(\cdot)$ is bounded in absolute value by $V_B$ (R.H.S of Inequality \eqref{sv-av-bnd}) and its codomain is $\mathbb{R}$, $V^*_t(\cdot)$ exists for all time steps and states.

Let $\gamma_{t + 1}(s_t, a_t, s_{t + 1})$ be abbreviated as $\gamma_{t + 1}$ for convenience.
By Lemma \ref{thm:sv-av-avgrwd-nvmdp},
\begin{equation*}
\begin{aligned}
&V^*_t(s) = \sup_{\pi} \E^\pi [\sum_{i = t}^{\infty}\Gamma^\tau_{t, i} \bar{r}_i(s_i, a_i) \mid s_t = s] \\
&= \sup_{\pi} \E^\pi [\bar{r}_t(s_t, a_t) + \gamma_{t + 1} \sum_{i = t + 1}^{\infty}\Gamma^\tau_{t + 1, i} \bar{r}_i(s_i, a_i) \mid s_t = s] \\
&= \sup_{\pi} \E^{a_t, s_{t + 1} \sim \pi} \Bigl[ \E^{a_{t + 1}, s_{t + 2}, ... \sim \pi}[\bar{r}_t(s_t, a_t) + \gamma_{t + 1} \sum_{i = t + 1}^{\infty}\Gamma^\tau_{t + 1, i} \bar{r}_i(s_i, a_i) \mid s_t, a_t, s_{t + 1}] \mid s_t = s \Bigr] \\
&= \sup_{\pi} \E^{a_t, s_{t + 1} \sim \pi} \Bigl[ \bar{r}_t(s_t, a_t) + \gamma_{t + 1} \E^{a_{t + 1}, s_{t + 2}, ... \sim \pi}[\sum_{i = t + 1}^{\infty}\Gamma^\tau_{t + 1, i} \bar{r}_i(s_i, a_i) \mid s_t, a_t, s_{t + 1}] \mid s_t = s \Bigr] \\
&\le \sup_{\pi} \E^{a_t, s_{t + 1} \sim \pi}[\bar{r}_t(s_t, a_t) + \gamma_{t + 1} V^*_{t + 1}(s_{t + 1}) \mid s_t = s] \\
&= \sup_{a_t} \E^{s_{t + 1} \sim p_t(\cdot \mid s_t, a_t)}[\bar{r}_t(s_t, a_t) + \gamma_{t + 1} V^*_{t + 1}(s_{t + 1}) \mid s_t = s] \ .
\end{aligned}
\end{equation*}

Let
\begin{equation}\phantomsection\label{theo-optim-act-nvmdp}
\widetilde{a}_t(s) \in \text{argmax}_{a_t} \E^{s_{t + 1} \sim p_t(\cdot \mid s_t, a_t)}[\bar{r}_t(s_t, a_t) + \gamma_{t + 1} V^*_{t + 1}(s_{t + 1}) \mid s_t = s] \ ,
\end{equation}

then
\begin{equation}\phantomsection\label{optim-sv-inequ-t}
V^*_t(s) \le \E^{s_{t + 1} \sim s_t, \widetilde{a}_t}[\widetilde{r}_t + \widetilde{\gamma}_{t + 1} V^*_{t + 1}(s_{t + 1}) \mid s_t = s] \ .
\end{equation}

($\widetilde{r}_t$ denotes $\bar{r}_t(s_t, \widetilde{a}_t(s_t))$, $\widetilde{\gamma}_{t + 1}$ denotes $\gamma_{t + 1}(s_t, \widetilde{a}_t(s_t), s_{t + 1})$.)
Replace $t$ by $t + 1$:
\begin{equation}\phantomsection\label{optim-sv-inequ-tp1}
V^*_{t + 1}(s') \le \E^{s_{t + 2} \sim s_{t + 1}, \widetilde{a}_{t + 1}}[\widetilde{r}_{t + 1} + \widetilde{\gamma}_{t + 2} V^*_{t + 2}(s_{t + 2}) \mid s_{t + 1} = s'] \ .
\end{equation}

Combine Inequality \eqref{optim-sv-inequ-t} and Inequality \eqref{optim-sv-inequ-tp1}:
\begin{equation*}
\begin{aligned}
V^*_t(s) &\le \E^{s_{t + 1} \sim s_t, \widetilde{a}_t}[\widetilde{r}_t + \widetilde{\gamma}_{t + 1}  \E^{s_{t + 2} \sim s_{t + 1}, \widetilde{a}_{t + 1}}[\widetilde{r}_{t + 1} + \widetilde{\gamma}_{t + 2} V^*_{t + 2}(s_{t + 2}) \mid s_{t + 1}] \mid s_t = s] \\
&= \E^{s_{t + 1} \sim s_t, \widetilde{a}_t}[\widetilde{r}_t + \widetilde{\gamma}_{t + 1}  \E^{s_{t + 2} \sim s_{t + 1}, \widetilde{a}_{t + 1}}[\widetilde{r}_{t + 1} + \widetilde{\gamma}_{t + 2} V^*_{t + 2}(s_{t + 2}) \mid s_{t + 1}, \widetilde{a}_t] \mid s_t = s] \\
&= \E^{s_{t + 1} \sim s_t, \widetilde{a}_t}[\E^{s_{t + 2} \sim s_{t + 1}, \widetilde{a}_{t + 1}}[\widetilde{r}_t + \widetilde{\gamma}_{t + 1}\widetilde{r}_{t + 1} + \widetilde{\gamma}_{t + 1}\widetilde{\gamma}_{t + 2} V^*_{t + 2}(s_{t + 2}) \mid s_{t + 1}, \widetilde{a}_t] \mid s_t = s] \\
&= \E^{s_{t + 1} \sim s_t, \widetilde{a}_t, \ s_{t + 2} \sim s_{t + 1}, \widetilde{a}_{t + 1}}[\widetilde{r}_t + \widetilde{\gamma}_{t + 1}\widetilde{r}_{t + 1} + \widetilde{\gamma}_{t + 1}\widetilde{\gamma}_{t + 2} V^*_{t + 2}(s_{t + 2}) \mid s_t = s]
\end{aligned}
\end{equation*}

Continue this recursion:
\begin{equation}\phantomsection\label{optim-sv-inequ-iter}
\begin{aligned}[b]
V^*_t(s) &\le \E^{s_{t + 1} \sim s_t, \widetilde{a}_t, \ s_{t + 2} \sim s_{t + 1}, \widetilde{a}_{t + 1}, \ ...}[\widetilde{r}_t + \widetilde{\gamma}_{t + 1}\widetilde{r}_{t + 1} + \widetilde{\gamma}_{t + 1}\widetilde{\gamma}_{t + 2}\widetilde{r}_{t + 1} + ... \mid s_t = s] \\
&\le V^*_t(s)
\end{aligned}
\end{equation}

The last inequality above is due to the optimality of $V^*_t(s)$.
Since $V^*_t(s)$ must equal to itself, every ``$\le$'' in Inequality \eqref{optim-sv-inequ-iter} must be ``=''.
This means always selecting $\widetilde{a}_t(s)$ (defined by Equation \eqref{theo-optim-act-nvmdp}) under state $s$ at time $t$ is an optimal policy that reaches $V^*_t(\cdot)$ for each time-state pair $(t, s)$.

By Equation \eqref{av-iter-nvmdp-e}, this deterministic policy is also optimal for the action-values.
$\blacksquare$

\begin{manualcorollary}{\ref{thm:optim-sv-av-relp}}
For all $t \ge 0, s \in S$,
\begin{equation}
V^*_t(s) = \max_a Q^*_t(s, a) \ .\tag{\ref{optim-sv-av-relp}}
\end{equation}
\end{manualcorollary}

\noindent\textbf{Proof:}
By Theorem \ref{thm:optim-pol-nvmdp}, there exists a deterministic policy $\pi^*$ which is optimal.
Further by Lemma \ref{thm:val-cns-nvmdp}, for a policy that deterministically selects $a$ in state $s$ at time $t$ and follows $\pi^*$ thereafter,
\begin{equation*}
V^{\pi^*}_t(s) = V^*_t(s) = \sup_{\pi \in \Pi} V^{\pi}_t(s) \ge Q^{\pi^*}_t(s, a) = Q^*_t(s, a) \ .
\end{equation*}
The above inequality holds for any action $a \in A$, therefore
\begin{equation*}
V^*_t(s) \ge \max_a Q^*_t(s, a) \ .
\end{equation*}

Moreover, by Equation \eqref{sv-av-nvmdp} and $\pi^*$ is deterministic,
\begin{equation*}
V^*_t(s) = V^{\pi^*}_t(s) \le \max_{a} Q^{\pi^*}_t(s, a) = \max_a Q^*_t(s, a) \ .
\end{equation*}

Combining the results above proves the corollary. $\blacksquare$

\begin{manualtheorem}{\ref{thm:bellman-nvmdp}}
For the following Bellman optimality equation in an NVMDP:
\begin{equation}
Q^\pi_t(s, a) = \bar{r}_t(s, a) + \E^{s' \sim p_t(\cdot \mid s, a)}[\gamma_{t + 1}(s, a, s') \max_{a'} Q^{\pi}_{t + 1}(s', a')] \ , \tag{\ref{bellman-equ-nvmdp}}
\end{equation}
where $\bar{r}_t(s, a)$ is defined by Equation \eqref{avg-rwd-nvmdp}: \\
(1) Equation \eqref{bellman-equ-nvmdp} holds for any policy $\pi$ that is optimal. \\
(2) Action-values satisfy Equation \eqref{bellman-equ-nvmdp} are optimal.
\end{manualtheorem}

\noindent\textbf{Proof:}
We prove part (1) first.
By Theorem \ref{thm:optim-pol-nvmdp}, there exists a deterministic policy $\pi^*$ which is optimal.
Furthermore, by Corollary \ref{thm:optim-sv-av-relp},
\begin{equation*}
V^{\pi^*}_{t + 1}(s') = \max_{a'} Q^{\pi^*}_{t + 1}(s', a')
\end{equation*}
Combining it with Equation \eqref{av-iter-nvmdp-e} leads to
\begin{equation*}
\begin{aligned}
Q^{\pi^*}_t(s, a) &= \bar{r}_t(s, a) + \E^{s' \sim p_t(\cdot \mid s, a)}[\gamma_{t + 1}(s, a, s') \max_{a'} Q^{\pi^*}_{t + 1}(s', a')] \\
\Rightarrow Q^*_t(s, a) &= \bar{r}_t(s, a) + \E^{s' \sim p_t(\cdot \mid s, a)}[\gamma_{t + 1}(s, a, s') \max_{a'} Q^*_{t + 1}(s', a')] \ .
\end{aligned}
\end{equation*}
Part (1) is thus proven.

For part (2), assume policy $\pi$ satisfies Equation \eqref{bellman-equ-nvmdp}.
By Equation \eqref{av-iter-nvmdp-mat}:
\begin{equation*}
\bm{Q}^{\pi}_t = \bm{r}_t + \bm{J}_{t + 1} \bm{V}^{Q^\pi}_{t + 1} \ .
\end{equation*}
Where $\bm{V}^{Q^\pi}_{t + 1}$ is a $|S|$-length vector that: $\bm{V}^{Q^\pi}_{t + 1}(s) = \max_a \bm{Q}^{\pi}_{t + 1}(s, a)$.

By the result of part (1),
\begin{equation*}
\bm{Q}^*_t = \bm{r}_t + \bm{J}_{t + 1} \bm{V}^{Q^*}_{t + 1} \ .
\end{equation*}

Thus,
\begin{equation}
\begin{aligned}[b]
\bm{Q}^{\pi}_t - \bm{Q}^*_t &= \bm{J}_{t + 1} (\bm{V}^{Q^\pi}_{t + 1} - \bm{V}^{Q^*}_{t + 1}) \\
\|\bm{Q}^{\pi}_t - \bm{Q}^*_t\|_\infty &= \|\bm{J}_{t + 1} (\bm{V}^{Q^\pi}_{t + 1} - \bm{V}^{Q^*}_{t + 1})\|_\infty \\
&\le \|\bm{J}_{t + 1}\|_\infty\|\bm{V}^{Q^\pi}_{t + 1} - \bm{V}^{Q^*}_{t + 1}\|_\infty \ .
\end{aligned}
\end{equation}

Since $(\bm{J}_{t + 1})_{(s, a), s'} = \gamma_{t + 1}(s, a, s') p_t(s' | s, a)$,
\begin{equation}\phantomsection\label{jt-upb}
\begin{aligned}[b]
\|\bm{J}_{t + 1}\|_\infty &= \max_{s, a} \sum_{s'} \gamma_{t + 1}(s, a, s') p_t(s' | s, a) \\
&\le \max_{s, a} \sum_{s'} p_t(s' | s, a) \max_{\bar{s}} \gamma_{t + 1}(s, a, \bar{s}) \\
&\le \max_{s, a} \max_{\bar{s}} \gamma_{t + 1}(s, a, \bar{s}) \\
&= \max_{s, a, s'} \gamma_{t + 1}(s, a, s') \ .
\end{aligned}
\end{equation}

For any two policies $\bar{\pi}$ and $\widetilde{\pi}$ and time $t$, denote
$\bar a \in \text{argmax}_a Q^{\bar{\pi}}_t(s, a)$ and $\widetilde a \in \text{argmax}_a Q^{\widetilde{\pi}}_t(s, a)$, then
\begin{equation*}
\begin{aligned}
Q^{\bar{\pi}}_t(s, \bar a) - Q^{\widetilde{\pi}}_t(s, \widetilde a)
&\le Q^{\bar{\pi}}_t(s, \bar a) - Q^{\widetilde{\pi}}_t(s, \bar a)
\le |Q^{\bar{\pi}}_t(s, \bar a) - Q^{\widetilde{\pi}}_t(s, \bar a)|
\le \max_a |Q^{\bar{\pi}}_t(s, a) - Q^{\widetilde{\pi}}_t(s, a)| \\
Q^{\bar{\pi}}_t(s, \bar a) - Q^{\widetilde{\pi}}_t(s, \widetilde a)
&\ge Q^{\bar{\pi}}_t(s, \widetilde a) - Q^{\widetilde{\pi}}_t(s, \widetilde a)
\ge -|Q^{\bar{\pi}}_t(s, \widetilde a) - Q^{\widetilde{\pi}}_t(s, \widetilde a)|
\ge -\max_a |Q^{\bar{\pi}}_t(s, a) - Q^{\widetilde{\pi}}_t(s, a)|
\end{aligned}
\end{equation*}

Thus,
\begin{equation}\phantomsection\label{absmaxqdiff-le-maxabsqdiff}
|\max_a Q^{\bar{\pi}}_t(s, a) - \max_a Q^{\widetilde{\pi}}_t(s, a)| \le \max_a |Q^{\bar{\pi}}_t(s, a) - Q^{\widetilde{\pi}}_t(s, a)| \ .
\end{equation}

Consequently, 
\begin{equation}\phantomsection\label{vdiff-le-qdiff}
\|\bm{V}^{Q^\pi}_{t + 1} - \bm{V}^{Q^*}_{t + 1}\|_\infty \le \|\bm{Q}^{\pi}_{t + 1} - \bm{Q}^*_{t + 1}\|_\infty \ .
\end{equation}

Combine Inequality \eqref{jt-upb} and Inequality \eqref{vdiff-le-qdiff},
\begin{equation*}
\begin{aligned}
\|\bm{Q}^{\pi}_t - \bm{Q}^*_t\|_\infty &\le \Bigl( \max_{s, a, s'} \gamma_{t + 1}(s, a, s') \Bigr)\|\bm{Q}^{\pi}_{t + 1} - \bm{Q}^*_{t + 1}\|_\infty \\
\Rightarrow \|\bm{Q}^{\pi}_t - \bm{Q}^*_t\|_\infty &\le \Bigl( \prod^m_{j = t + 1} \max_{s, a, s'} \gamma_j(s, a, s') \Bigr)\|\bm{Q}^{\pi}_m - \bm{Q}^*_m\|_\infty \quad(\forall m \ge t + 1) \ .
\end{aligned}
\end{equation*}

By Assumption \ref{asm:gamma-bellman-cond-nvmdp} and the boundness of $Q^\pi_t(\cdot)$, $\bm{Q}^{\pi}_t = \bm{Q}^*_t$ for all $t$.
This proves that the action-values of policy $\pi$ are optimal.
$\blacksquare$

\begin{manualcorollary}{\ref{thm:reward-shaping-nvmdp}}
For an NVMDP and functions $\Phi_t: S \rightarrow \mathbb{R} \ (\forall t \ge 0)$, if
\begin{equation*}
\widetilde{r}_t(s, a, s') = r_t(s, a, s') + \gamma_{t + 1}(s, a, s') \Phi_{t + 1}(s') - \Phi_t(s) \quad(\forall t \ge 0, s, s' \in S, a \in A)
\end{equation*}
is bounded in absolute-value, then an optimal policy under the reward $r_t(s, a, s')$ is also optimal under the reward $\widetilde{r}_t(s, a, s')$, and vice versa.
\end{manualcorollary}

\noindent\textbf{Proof:}
Consider an optimal policy $\pi^*$ under $r_t(s, a, s')$.
By Equation \eqref{avg-rwd-nvmdp} and part (1) of Theorem \ref{thm:bellman-nvmdp}, 
\begin{equation*}
Q^{\pi^*}_t(s, a) = \E^{s' \sim p_t(\cdot | s, a)}[r_t(s, a, s') + \gamma_{t + 1}(s, a, s') \max_{a'} Q^{\pi^*}_{t + 1}(s', a')] \ .
\end{equation*}

Therefore,
\begin{equation*}
\begin{aligned}
& Q^{\pi^*}_t(s, a) - \Phi_t(s) = \E^{s' \sim p_t(\cdot | s, a)}[r_t(s, a, s') - \Phi_t(s) + \gamma_{t + 1}(s, a, s') \max_{a'} Q^{\pi^*}_{t + 1}(s', a')] \\
&= \E^{s' \sim p_t(\cdot | s, a)}\bigg[r_t(s, a, s') + \gamma_{t + 1}(s, a, s') \Phi_{t + 1}(s') - \Phi_t(s) + \gamma_{t + 1}(s, a, s') \bigg(\Big(\max_{a'} Q^{\pi^*}_{t + 1}(s', a')\Big) - \Phi_{t + 1}(s')\bigg)\bigg] \\
&= \E^{s' \sim p_t(\cdot | s, a)}[\widetilde{r}_t(s, a, s')] + \E^{s' \sim p_t(\cdot | s, a)}\bigg[\gamma_{t + 1}(s, a, s') \bigg(\max_{a'} \Big(Q^{\pi^*}_{t + 1}(s', a') - \Phi_{t + 1}(s')\Big)\bigg)\bigg] \ .
\end{aligned}
\end{equation*}
By part (2) of Theorem \ref{thm:bellman-nvmdp}, $Q^{\pi^*}_t(s, a) - \Phi_t(s)$ is the optimal action-value when $\widetilde{r}_t(s, a, s')$ becomes the reward of the NVMDP, proving that $\pi^*$ is also optimal under $\widetilde{r}_t(s, a, s')$.
Moreover, since $r_t(s, a, s')$ is bounded in absolute value and
\begin{equation*}
r_t(s, a, s') = \widetilde{r}_t(s, a, s') + \gamma_{t + 1}(s, a, s') \Big(-\Phi_{t + 1}(s')\Big) - \Big(-\Phi_t(s)\Big) \ ,
\end{equation*}
an optimal policy under $\widetilde{r}_t(s, a, s')$ is also optimal under $r_t(s, a, s')$.
$\blacksquare$

\begin{manualtheorem}{\ref{thm:pol-imprv-nvmdp}}
For a given time $n$, policy $\pi$ and $\pi'$ are identical for all time $t \ge n$.
Furthermore,
\begin{equation*}
\sum_{\bar{a} \in A} \pi'_t(\bar{a} | s) Q^\pi_t(s, \bar{a}) \ge V^\pi_t(s) \quad\forall s \in S, t < n \ .
\end{equation*}
Then: \\
(1) for any $s \in S$, $V^{\pi'}_t(s) = V^\pi_t(s) \ (\forall t \ge n)$ and $V^{\pi'}_t(s) \ge V^\pi_t(s) \ (\forall t \le n - 1)$. \\
(2) for any $(s, a) \in S \times A$, $Q^{\pi'}_t(s, a) = Q^\pi_t(s, a) \ (\forall t \ge n - 1)$ and $Q^{\pi'}_t(s, a) \ge Q^\pi_t(s, a) \ (\forall t < n - 1)$.
\end{manualtheorem}

\noindent\textbf{Proof:}
By Lemma \ref{thm:val-cns-nvmdp}, $\bm{Q}^{\pi'}_t = \bm{Q}^\pi_t$ for all $t \ge n - 1$ and $\bm{V}^{\pi'}_t = \bm{V}^\pi_t$ for all $t \ge n$.

By Equation \eqref{sv-av-nvmdp-mat},
\begin{equation}\phantomsection\label{sv-av-nvmdp-polimpr}
\begin{aligned}[b]
\bm{V}^{\pi'}_t - \bm{V}^\pi_t &= \bm{\Pi}^{\pi'}_t \bm{Q}^{\pi'}_t - \bm{V}^\pi_t \\
&= \bm{\Pi}^{\pi'}_t (\bm{Q}^{\pi'}_t - \bm{Q}^\pi_t) + (\bm{\Pi}^{\pi'}_t \bm{Q}^\pi_t - \bm{V}^\pi_t) \ .
\end{aligned}
\end{equation}

Thus $\bm{V}^{\pi'}_{n - 1} \ge \bm{V}^\pi_{n - 1}$ since $\bm{Q}^{\pi'}_{n - 1} = \bm{Q}^\pi_{n - 1}$ and $\bm{\Pi}^{\pi'}_{n - 1} \bm{Q}^\pi_{n - 1} \ge \bm{V}^\pi_{n - 1}$.

By Equation \eqref{av-iter-nvmdp-mat},
\begin{equation*}
\bm{Q}^{\pi'}_{n - 2} - \bm{Q}^\pi_{n - 2} = \bm{J}_{n - 1} (\bm{V}^{\pi'}_{n - 1} - \bm{V}^\pi_{n - 1}) \ge \bm{0} \ .
\end{equation*}

By Equation \eqref{sv-av-nvmdp-polimpr},
\begin{equation*}
\bm{V}^{\pi'}_{n - 2} - \bm{V}^\pi_{n - 2} = \bm{\Pi}^{\pi'}_{n - 2} (\bm{Q}^{\pi'}_{n - 2} - \bm{Q}^\pi_{n - 2}) + (\bm{\Pi}^{\pi'}_{n - 2} \bm{Q}^\pi_{n - 2} - \bm{V}^\pi_{n - 2}) \ge \bm{0} \ .
\end{equation*}

Then,
\begin{equation*}
\begin{aligned}
\bm{Q}^{\pi'}_{n - 3} - \bm{Q}^\pi_{n - 3} &= \bm{J}_{n - 2} (\bm{V}^{\pi'}_{n - 2} - \bm{V}^\pi_{n - 2}) \ge \bm{0} \\
\bm{V}^{\pi'}_{n - 3} - \bm{V}^\pi_{n - 3} &= \bm{\Pi}^{\pi'}_{n - 3} (\bm{Q}^{\pi'}_{n - 3} - \bm{Q}^\pi_{n - 3}) + (\bm{\Pi}^{\pi'}_{n - 3} \bm{Q}^\pi_{n - 3} - \bm{V}^\pi_{n - 3}) \ge \bm{0} \ .
\end{aligned}
\end{equation*}

Continuing the deduction iteratively until time 0 completes the proof.
$\blacksquare$

\begin{manualtheorem}{\ref{thm:station-nvmdp-is-classic}}
If a NVMDP satisfies $r_t(s, a, s') = r(s, a, s')$ and $p_t(s', r | s, a) = p(s', r | s, a)$ for all $t \ge 0, s \in S, a \in A$ and all discount rates $\gamma_{t + 1}(s, a, s') = \gamma(s, a, s') \in [0, 1)$, then: \\
(1) $V^*_t(s) = V^*_k(s)$ and $Q^*_t(s, a) = Q^*_k(s, a)$ for all $t, k \ge 0, s \in S, a \in A$. \\
(2) for optimal policy $\pi^* = (\pi^*_0, \pi^*_1, \pi^*_2, ...)$ and any integer ${\bar k} \ge 0$, policy $(\pi^*_{\bar k}, \pi^*_{\bar k}, \pi^*_{\bar k}, ...)$ is also optimal.
\end{manualtheorem}

\noindent\textbf{Proof:}
Both $p(s' | s, a) = \sum_{r} p(s', r | s, a)$ and $\bar{r}(s, a) = \E^{s' \sim p(\cdot | s, a)}[r(s, a, s')]$ are stationary now.
Moreover, $\gamma_{t + 1}(s, a, s') = \gamma(s, a, s')$, thus Equation \eqref{j-def-nvmdp-mat} becomes
\begin{equation}\phantomsection\label{j-mat-const}
\bm{J}_{t + 1} = \bm{J} = \bm{W} \odot \bm{P} \ ,
\end{equation}
where $\bm{W}_{(s, a), s'} = \gamma(s, a, s')$ and $\bm{P}_{(s, a), s'} = p(s' | s, a)$.
Futhermore, since both state and action space are finite, there exists
\begin{equation*}\phantomsection\label{max-gamma-station}
\begin{aligned}[b]
\gamma_M &= \max_{s, a, s'} \gamma(s, a, s') \in [0, 1) \\
\Rightarrow \|\bm{J}\|_\infty &= \|\bm{W} \odot \bm{P}\|_\infty \\
&\le \|\gamma_M \bm{P}\|_\infty = \gamma_M \|\bm{P}\|_\infty = \gamma_M \ .
\end{aligned}
\end{equation*}

By Equation \eqref{av-iter-nvmdp-mat} and Equation \eqref{j-mat-const}, 
\begin{equation}\phantomsection\label{av-iter-clascmdp-mat}
\bm{Q}^*_t = \bm{r} + \bm{J} \bm{V}^*_{t + 1} \ .
\end{equation}
Replace $t$ by $t + 1$ in Equation \eqref{av-iter-clascmdp-mat}:
\begin{equation*}
\begin{aligned}
\bm{Q}^*_{t + 1} &= \bm{r} + \bm{J} \bm{V}^*_{t + 2} \\
\Rightarrow \bm{Q}^*_{t + 1} - \bm{Q}^*_t &= \bm{J} (\bm{V}^*_{t + 2} - \bm{V}^*_{t + 1}) \\
\Rightarrow\|\bm{Q}^*_{t + 1} - \bm{Q}^*_t\|_\infty &\le \|\bm{J}\|_\infty\|\bm{V}^*_{t + 2} - \bm{V}^*_{t + 1}\|_\infty \\
&\le \gamma_M \|\bm{V}^*_{t + 2} - \bm{V}^*_{t + 1}\|_\infty \quad(\|\bm{J}\|_\infty \le \gamma_M)
\end{aligned}
\end{equation*}

Similar to Inequality \eqref{absmaxqdiff-le-maxabsqdiff},
\begin{equation}\phantomsection\label{vdiff-le-qdiff-clascmdp}
\begin{aligned}[b]
|V^*_{t + 1}(s) - V^*_t(s)| &= |\max_a Q^*_{t + 1}(s, a) - \max_a Q^*_t(s, a)| \\
&\le \max_a |Q^*_{t + 1}(s, a) - Q^*_t(s, a)| \\
\Rightarrow\|\bm{V}^*_{t + 1} - \bm{V}^*_t\|_\infty &\le \|\bm{Q}^*_{t + 1} - \bm{Q}^*_t\|_\infty \quad(\forall t \ge 0)
\end{aligned}
\end{equation}

Thus,
\begin{equation*}
\begin{aligned}
\|\bm{Q}^*_{t + 1} - \bm{Q}^*_t\|_\infty &\le \gamma_M \|\bm{V}^*_{t + 2} - \bm{V}^*_{t + 1}\|_\infty \le \gamma_M \|\bm{Q}^*_{t + 2} - \bm{Q}^*_{t + 1}\|_\infty \\
\Rightarrow\|\bm{Q}^*_{t + 1} - \bm{Q}^*_t\|_\infty &\le \gamma_M^m\|\bm{Q}^*_{t + m + 1} - \bm{Q}^*_{t + m}\|_\infty \quad(\forall m \ge 1) \ .
\end{aligned}
\end{equation*}
Since $0 \le \gamma_M < 1$ and action-values are bounded in absolute values, $\bm{Q}^*_{t + 1} = \bm{Q}^*_t$.
Also by Inequality \eqref{vdiff-le-qdiff-clascmdp}, $\bm{V}^*_{t + 1} = \bm{V}^*_t$.
Part (1) is proven.

Denote $\bar{\pi} = \{\pi^*_{\bar k}, \pi^*_{\bar k}, ..., \pi^*_{\bar k}, ...\}$ for a given $\bar{k}$.
By Equation \eqref{sv-av-nvmdp-mat} and part (1),
\begin{equation}\phantomsection\label{sv-av-clascmdp-mat}
\begin{aligned}[b]
\bm{V}^*_{\bar{k}} &= \bm{\Pi}^{\bar{\pi}}_{\bar{k}} \bm{Q}^*_{\bar{k}} \\
\Rightarrow \bm{V}^*_t &= \bm{\Pi}^{\bar{\pi}}_{\bar{k}} \bm{Q}^*_t \quad(\forall t \ge 0) \ .
\end{aligned}
\end{equation}
Equation \eqref{av-iter-clascmdp-mat} thus becomes
\begin{equation*}
\bm{Q}^*_t = \bm{r} + \bm{J} \bm{\Pi}^{\bar{\pi}}_{\bar{k}} \bm{Q}^*_{t + 1} \quad(\forall t \ge 0) \ .
\end{equation*}
Moreover, by Equation \eqref{av-selfiter-nvmdp-mat} and Equation \eqref{jk-relp-nvmdp-mat},
\begin{equation*}
\begin{aligned}
\bm{Q}^{\bar{\pi}}_t &= \bm{r} + \bm{J} \bm{\Pi}^{\bar{\pi}}_{\bar{k}} \bm{Q}^{\bar{\pi}}_{t + 1} \quad(\forall t \ge 0) \\
\Rightarrow \bm{Q}^{\bar{\pi}}_t - \bm{Q}^*_t &= \bm{J} \bm{\Pi}^{\bar{\pi}}_{\bar{k}} (\bm{Q}^{\bar{\pi}}_{t + 1} - \bm{Q}^*_{t + 1}) \\
\Rightarrow \|\bm{Q}^{\bar{\pi}}_t - \bm{Q}^*_t\|_\infty &\le \gamma_M \| \bm{Q}^{\bar{\pi}}_{t + 1} - \bm{Q}^*_{t + 1}\|_\infty \quad(\|\bm{J}\|_\infty \le \gamma_M, \ \|\bm{\Pi}^{\bar{\pi}}_{\bar{k}}\|_\infty = 1) \\
\Rightarrow\|\bm{Q}^{\bar{\pi}}_t - \bm{Q}^*_t\|_\infty &\le \gamma_M^m \|\bm{Q}^{\bar{\pi}}_{t + m + 1} - \bm{Q}^*_{t + m + 1}\|_\infty \quad(\forall m \ge 1)
\end{aligned}
\end{equation*}

By $0 \le \gamma_M < 1$ and action-values being bounded in absolute values, $\bm{Q}^{\bar{\pi}}_t = \bm{Q}^*_t$ for all time $t$.
Futhermore, by Equation \eqref{sv-av-nvmdp-mat} and Equation \eqref{sv-av-clascmdp-mat},
\begin{equation*}
\bm{V}^{\bar{\pi}}_t = \bm{\Pi}^{\bar{\pi}}_{\bar{k}} \bm{Q}^{\bar{\pi}}_t = \bm{\Pi}^{\bar{\pi}}_{\bar{k}} \bm{Q}^*_t = \bm{V}^*_t \ ,
\end{equation*}
part (2) is thus proven.
$\blacksquare$

\begin{manualtheorem}{\ref{thm:dp-poleval-nvmdp-limit0}}
Consider a NVMDP and a given policy $\pi$.
For any time $t$, the estimated state-values $\hat{V}^\pi_t(\cdot)$ and action-values $\hat{Q}^\pi_t(\cdot)$ by Algorithm \ref{algo:dp-poleval-nvmdp} with horizon $H$ satisfy:
\begin{equation*}
\lim_{H \rightarrow \infty} \hat{V}^\pi_t(s) = V^\pi_t(s) \ , \quad \lim_{H \rightarrow \infty} \hat{Q}^\pi_t(s, a) = Q^\pi_t(s, a) \quad(\forall s \in S, a \in A) \ .
\end{equation*}
\end{manualtheorem}

\noindent\textbf{Proof:}
For $t \ge H$, $\hat{Q}^\pi_t(\cdot) = 0$.

For $t < H$, the $\hat{Q}^\pi_t(\cdot)$ by Algorithm \ref{algo:dp-poleval-nvmdp} satisfies Equation \eqref{av-selfiter-nvmdp-mat}:
\begin{equation*}
\bm{\hat{Q}}^{\pi}_t = \bm{r}_t + \bm{K}^\pi_{t + 1} \bm{\hat{Q}}^{\pi}_{t + 1} \ .
\end{equation*}

Thus,
\begin{equation*}
\bm{\hat{Q}}^{\pi}_t - \bm{Q}^\pi_t = \bm{K}^\pi_{t + 1} (\bm{\hat{Q}}^{\pi}_{t + 1} - \bm{Q}^{\pi}_{t + 1}) \quad(\forall t < H) \ .
\end{equation*}

Similar to the proof of part (1) of Theorem \ref{thm:pol-imprv-nvmdp}, $\forall t < H$,
\begin{equation*}
\begin{aligned}
\|\bm{\hat{Q}}^{\pi}_t - \bm{Q}^\pi_t\|_\infty &\le \Bigl( \prod^H_{j = t + 1} \max_{s, a, s'} \gamma_j(s, a, s') \Bigr)\|\bm{\hat{Q}}^{\pi}_H - \bm{Q}^\pi_H\|_\infty \\
&= \Bigl( \prod^H_{j = t + 1} \max_{s, a, s'} \gamma_j(s, a, s') \Bigr)\|\bm{Q}^\pi_H\|_\infty \quad(\bm{\hat{Q}}^{\pi}_H = \bm{0}) \ .
\end{aligned}
\end{equation*}

By Assumption \ref{asm:gamma-bellman-cond-nvmdp} and the boundness of $Q^\pi_t(\cdot)$, $\lim_{H \rightarrow \infty}\|\bm{\hat{Q}}^\pi_t - \bm{Q}^\pi_t\|_\infty = 0$.

Also $\forall t < H$,
\begin{equation*}
\begin{aligned}
\bm{\hat{V}}^{\pi}_t &= \bm{\Pi}^\pi_t \bm{\hat{Q}}^{\pi}_t \\
\Rightarrow\|\bm{\hat{V}}^{\pi}_t - \bm{V}^\pi_t\|_\infty &\le \|\bm{\Pi}^\pi_t\|_\infty\|\bm{\hat{Q}}^{\pi}_t - \bm{Q}^\pi_t\|_\infty \ .
\end{aligned}
\end{equation*}

By Equation \eqref{pi-def-nvmdp-mat}, $\|\bm{\Pi}^\pi_t\|_\infty = 1$.
Thus $\lim_{H \rightarrow \infty}\|\bm{\hat{V}}^{\pi}_t - \bm{V}^\pi_t\|_\infty = 0$.
$\blacksquare$

\begin{manualtheorem}{\ref{thm:dp-valiter-nvmdp-limit0}}
For a NVMDP and any time $t$, the estimated optimal state-values $\hat{V}^*_t(\cdot)$ and action-values $\hat{Q}^*_t(\cdot)$ by Algorithm \ref{algo:dp-valiter-nvmdp} with horizon $H$ satisfy:
\begin{equation*}
\lim_{H \rightarrow \infty} \hat{V}^*_t(s) = V^*_t(s) \ , \quad \lim_{H \rightarrow \infty} \hat{Q}^*_t(s, a) = Q^*_t(s, a) \quad(\forall s \in S, a \in A) \ .
\end{equation*}
\end{manualtheorem}

\noindent\textbf{Proof:}
For $t \ge H$, $\hat{Q}^*_t(\cdot) = 0$.

For $t < H$, the $\hat{Q}^*_t(\cdot)$ by Algorithm \ref{algo:dp-valiter-nvmdp} satisfies Equation \eqref{av-iter-nvmdp-mat}:
\begin{equation*}
\bm{\hat{Q}}^*_t = \bm{r}_t + \bm{J}_{t + 1} \bm{\hat{V}}^*_{t + 1} \ ,
\end{equation*}
where $\bm{\hat{V}}^*_t(s) = \max_a \bm{\hat{Q}}^*_t(s, a) \ (\forall t < H)$ and $\bm{\hat{V}}^*_H = \bm{0}$.

Thus,
\begin{equation*}
\bm{\hat{Q}}^*_t - \bm{Q}^*_t = \bm{J}_{t + 1} (\bm{\hat{V}}^*_{t + 1} - \bm{V}^*_{t + 1}) \quad(\forall t < H) \ .
\end{equation*}

Similar to the proof of part (2) of Theorem \ref{thm:bellman-nvmdp}, $\forall t < H$,
\begin{equation*}
\begin{aligned}
\|\bm{\hat{V}}^*_t - \bm{V}^*_t\|_\infty &\le \|\bm{\hat{Q}}^{\pi}_t - \bm{Q}^*_t\|_\infty \quad(\bm{V}^*_t(s) = \max_a \bm{Q}^*_t(s, a)) \\
\|\bm{\hat{Q}}^*_t - \bm{Q}^*_t\|_\infty &\le \Bigl( \prod^H_{j = t + 1} \max_{s, a, s'} \gamma_j(s, a, s') \Bigr)\|\bm{\hat{Q}}^*_H - \bm{Q}^*_H\|_\infty \\
&= \Bigl( \prod^H_{j = t + 1} \max_{s, a, s'} \gamma_j(s, a, s') \Bigr)\|\bm{Q}^*_H\|_\infty \quad(\bm{\hat{Q}}^*_H = \bm{0}) \ .
\end{aligned}
\end{equation*}

By Assumption \ref{asm:gamma-bellman-cond-nvmdp} and the boundness of $Q^*_t(\cdot)$,
\begin{equation*}
\lim_{H \rightarrow \infty}\|\bm{\hat{Q}}^*_t - \bm{Q}^*_t\|_\infty = 0 \ \Rightarrow\  \lim_{H \rightarrow \infty}\|\bm{\hat{V}}^*_t - \bm{V}^*_t\|_\infty = 0 \ .\ \blacksquare
\end{equation*}

Generalized Q-learning, introduced by Lan et al. \cite{Maxmin-q-learn-2020}, extends the classic Q-learning algorithm proposed by Watkins et al. \cite{Watkins-PhD-1989} to a broader framework applicable to classic MDPs.
To establish the convergence of generalized Q-learning in NVMDPs, we first adapt the theoretical framework and proofs from Tsitsiklis \cite{stoch-approx-ql-1994}.

Consider updating $\bm{x} = (x_1, x_2, ..., x_n)$ asynchronously with the following rule:
\begin{equation}\phantomsection\label{stoapp-x-upd}
x_i(k + 1) = x_i(k) + \alpha_i(k) \big(F_i(\bm{X}_k^i) - x_i(k) + w_i(k)\big) \ ,
\end{equation}
where
\begin{equation}\phantomsection\label{stoapp-Xik-def}
\bm{X}_k^i = \begin{bmatrix}
x_1\big(\tau_{11}^i(k)\big) & x_1\big(\tau_{12}^i(k)\big) & \cdots & x_1\big(\tau_{1l}^i(k)\big) \\
x_2\big(\tau_{21}^i(k)\big) & x_2\big(\tau_{22}^i(k)\big) & \cdots & x_2\big(\tau_{2l}^i(k)\big) \\
\vdots & \vdots & \ddots & \vdots \\
x_n\big(\tau_{n1}^i(k)\big) & x_n\big(\tau_{n2}^i(k)\big) & \cdots & x_n\big(\tau_{nl}^i(k)\big) \\
\end{bmatrix}
\end{equation}
is a $n \times l$ matrix and function $F_i: {\mathbb R}^{n \times l} \rightarrow {\mathbb R}$.
Here, $k$ is a non-negative integer that denotes the update step.
Row $u \ (1 \le u \le n)$ of $\bm{X}_k^i$ contains historical estimates of $x_u$, where $0 \le \tau_{uv}^i(k) \le k \ (\forall 1 \le v \le l)$.
The superscript $i$ in $\bm{X}_k^i$ and $\tau_{uv}^i(k)$ refer to $i$-th element of $\bm{x}$ instead of exponentials.
Stepsize $\alpha_i(k) \in [0, 1]$, and $\alpha_i(k) = 0$ if $x_i$ is not updated at step $k$.

Moreover, we view all $x(k), \tau_{uv}^i(k), \alpha_i(k), w_i(k)$ as random variables defined on a probability space $(\Omega$, $\mathcal{F}, P)$ and have the following assumptions:
\begin{assumption}\phantomsection\label{asm:tau-go-infty}
$\forall 1 \le i, u \le n, 1 \le v \le l, \lim_{k \rightarrow \infty} \tau_{uv}^i(k) = \infty$ w.p. 1.
\end{assumption}

\begin{assumption}\phantomsection\label{asm:F-measurable}
(a) $\forall i$, $x_i(0)$ is $\mathcal{F}(0)$-measurable. \\
(b) $\forall i, k$, $w_i(k)$ is $\mathcal{F}(k + 1)$-measurable. \\
(c) $\forall i, k$, $\alpha_i(k)$ and $\tau_{uv}^i(k)$ are $\mathcal{F}(k)$-measurable. \\
(d) $\forall i, k$, $\E[w_i(k) \mid \mathcal{F}(k)] = 0$. \\
(e) Constants $A, B$ exist s.t. $\E[w_i^2(k) \mid \mathcal{F}(k)] \le A + B \max_{\tau \le k}\|\bm{x}(\tau)\|_\infty$.
\end{assumption}

For event space $\mathcal{F}$, $\mathcal{F}(k)$ represents the algorithm history up to (and including) the point that stepsize $\alpha_i(k)$ are obtained, but before the generation of noise $w_i(k)$.

\begin{assumption}\phantomsection\label{asm:stoapp-stepsize}
(a) $\forall i, \sum^\infty_{k = 0} \alpha_i(k) = \infty$ w.p. 1. \\
(b) Constant $C$ exists s.t. $\forall i, \sum^\infty_{k = 0} \alpha^2_i(k) \le C$ w.p. 1.
\end{assumption}

\begin{assumption}\phantomsection\label{asm:stoapp-contract}
There exists constant $\beta \in [0, 1)$ and $(x^*_1, x^*_2, ..., x^*_n) \in \mathbb{R}^n$ s.t.
\begin{equation*}
\|F_X(\bm{X}) - \bm{X}^*\|_{\max} \le \beta\|\bm{X} - \bm{X}^*\|_{\max} \quad(\forall 1 \le i \le n, \bm{X} \in \mathbb{R}^{n \times l})\ ,
\end{equation*}
\end{assumption}
where
\begin{equation*}
F_X(\bm{X}) = \begin{bmatrix}
F_1(\bm{X})\bm{1} \\
F_2(\bm{X})\bm{1} \\
\vdots \\
F_n(\bm{X})\bm{1}
\end{bmatrix} \ , \ 
\bm{X}^* = \begin{bmatrix}
x^*_1\bm{1} \\
x^*_2\bm{1} \\
\vdots \\
x^*_n\bm{1}
\end{bmatrix} \ , \
\bm{1} = (1, 1, ..., 1) \in \mathbb{R}^l \ .
\end{equation*}

According to probablity theory, if $P(A) = 1, A \Rightarrow B$, then $P(B) \ge P(A) = 1$.
Moreover, if $P(A_i) = 1, \forall i \in \{1, 2, ..., p\}$, then $P(\bigcap_{i = 1}^p A_i) \ge \big(\sum_{i = 1}^p P(A_i)\big) - p + 1 = 1$.
Therefore, if $B$ is deducted from all events $A_i$ that hold always or almost surely, then $P(B) = 1$.
In the following Lemma \ref{thm:stoapp-xk-absbnd} and \ref{thm:stoapp-xk-convg}, we prove that $\bm{x} = (x_1, x_2, ..., x_n)$ is bounded in absolute values and converges with probability 1, respectively.
Throughout, we treat all conditions that hold almost surely as if they hold always.
Consequently, the resulting conclusions are almost surely true.

\begin{lemma}\label{thm:stoapp-xk-absbnd}
With Assumption \ref{asm:tau-go-infty}, \ref{asm:F-measurable}, \ref{asm:stoapp-stepsize}, and \ref{asm:stoapp-contract}, $\bm{x}(k)$ in Equation \eqref{stoapp-x-upd} is bounded in absolute values w.p. 1.
\end{lemma}

\noindent\textbf{Proof:}
By Assumption \ref{asm:stoapp-contract}, 
\begin{equation}\phantomsection\label{FX-Xik-upb}
\begin{aligned}[b]
\|F_X(\bm{X}_k^i)\|_{\max} &\le \|F_X(\bm{X}_k^i) - \bm{X}^*\|_{\max} +\|\bm{X}^*\|_{\max} \\
&\le \beta\|\bm{X}_k^i - \bm{X}^*\|_{\max} +\|\bm{X}^*\|_{\max} \\
&\le \beta\|\bm{X}_k^i\|_{\max} + D \quad(D = (1 + \beta)\|\bm{X}^*\|_{\max}) \\
&\le \beta \max_{\tau \le k}\|\bm{x}(\tau)\|_\infty + D
\end{aligned}
\end{equation}

Consider some constants $\gamma$ and $G_0$ s.t. $\beta < \gamma < 1, G_0 \ge {{D}\over {\gamma - \beta}}$, then
\begin{equation*}
\beta G_0 + D \le \gamma G_0 \ .
\end{equation*}
If $\max_{\tau \le k}\|\bm{x}(\tau)\|_\infty \le G_0$, then by Inequality \eqref{FX-Xik-upb},
\begin{equation*}
\|F_X(\bm{X}_k^i)\|_{\max} \le \beta G_0 + D \le \gamma G_0 \ .
\end{equation*}
Otherwise $\max_{\tau \le k}\|\bm{x}(\tau)\|_\infty > G_0$, then by Inequality \eqref{FX-Xik-upb} and $G_0 \ge {{D}\over {\gamma - \beta}}$,
\begin{equation*}
\begin{aligned}
\|F_X(\bm{X}_k^i)\|_{\max} &\le \gamma \max_{\tau \le k}\|\bm{x}(\tau)\|_\infty + (\beta - \gamma) \max_{\tau \le k}\|\bm{x}(\tau)\|_\infty + D \\
&< \gamma \max_{\tau \le k}\|\bm{x}(\tau)\|_\infty + (\beta - \gamma) {{D}\over {\gamma - \beta}} + D \quad(\beta - \gamma < 0)\\
&= \gamma \max_{\tau \le k}\|\bm{x}(\tau)\|_\infty \ .
\end{aligned}
\end{equation*}
Therefore,
\begin{equation}\phantomsection\label{FX-Xik-max-upb}
\|F_X(\bm{X}_k^i)\|_{\max} \le \gamma \max\{\max_{\tau \le k}\|\bm{x}(\tau)\|_\infty, G_0\} \ .
\end{equation}

The remainder of the proof follows the same argument as Section 4 ``Proof of Theorem 1'' in Tsitsiklis \cite{stoch-approx-ql-1994}, with the following modifications: Inequality (11) and $F_i(\bm{x}^i(t))$ in the udpate formula are replaced by Inequality \eqref{FX-Xik-max-upb} and $F_i(\bm{X}_k^i)$, respectively.
Additionally, while \cite{stoch-approx-ql-1994} uses $t$ as the time index for updates, we use $k$ here.
$\blacksquare$

\begin{lemma}\label{thm:stoapp-xk-convg}
With Assumption \ref{asm:tau-go-infty}, \ref{asm:F-measurable}, \ref{asm:stoapp-stepsize}, and \ref{asm:stoapp-contract}, $\bm{x}(k)$ in Equation \eqref{stoapp-x-upd} satisfies
\begin{equation*}
\lim_{k \rightarrow \infty} \bm{x}(k) = (x^*_1, x^*_2, ..., x^*_n) \quad\textit{w.p. 1.}
\end{equation*}
\end{lemma}

\noindent\textbf{Proof:}
Denote $\bm{x}^* = (x^*_1, x^*_2, ..., x^*_n)$. 
Since $0 \le \beta < 1$, $\exists \text{ constant } \delta > 0 \text{ s.t. } \beta + \delta < 1$. 

By Equation \eqref{stoapp-x-upd},
\begin{equation}\phantomsection\label{stoapp-x-diff-upd}
\begin{aligned}[b]
x_i(k + 1) &= \big(1 - \alpha_i(k)\big) x_i(k) + \alpha_i(k) F_i(\bm{X}_k^i) + \alpha_i(k) w_i(k) \\
x_i(k + 1) - x^*_i &= \big(1 - \alpha_i(k)\big) \big(x_i(k) - x^*_i\big) + \alpha_i(k) \big(F_i(\bm{X}_k^i) - x^*_i\big) + \alpha_i(k) w_i(k)
\end{aligned}
\end{equation}

By Lemma \ref{thm:stoapp-xk-absbnd}, $\bm{x}(k)$ is bounded in absolute value.
Thus there exists $D_0 > 0$ s.t. for all $k \ge 0$, $\|\bm{x}(k) - \bm{x}^*\|_\infty \le D_0$ w.p. 1.
Assume there exists $k_s$ and $D_s > 0$ s.t. $\forall k \ge k_s, \|\bm{x}(k) - \bm{x}^*\|_\infty \le D_s$.
Then by Assumption \ref{asm:tau-go-infty}, there exists $\tau_s \ge k_s$ s.t.
\begin{equation*}
\|\bm{X}_k^i - \bm{X}^*\|_{\max} \le D_s \quad(\forall k \ge \tau_s) \ .
\end{equation*}
By Assumption \ref{asm:stoapp-contract},
\begin{equation}\phantomsection\label{Fx-xik-diff-absupd}
\begin{aligned}[b]
\|F_X(\bm{X}_k^i) - \bm{X}^*\|_{\max} &\le \beta \|\bm{X}_k^i - \bm{X}^*\|_{\max} \le \beta D_s \quad(\forall k \ge \tau_s)\\
\Rightarrow |F_i(\bm{X}_k^i) - x_i(k)| &\le \beta D_s \quad(\forall k \ge \tau_s)\ .
\end{aligned}
\end{equation}

Consider sequence $\{Y_i(k)\}$ for $k \ge \tau_s$ that
\begin{equation}\phantomsection\label{Yi-def}
Y_i(k + 1) = \big(1 - \alpha_i(k)\big) Y_i(k) + \alpha_i(k) \beta D_s \ ,
\end{equation}
and $Y_i(\tau_s) = D_s$. By Equation \eqref{Yi-def},
\begin{equation*}
Y_i(k + 1) - \beta D_s = \big(1 - \alpha_i(k)\big) \big(Y_i(k) - \beta D_s\big) \ .
\end{equation*}
Bey Lemma 1 of \cite{stoch-approx-ql-1994}, $Y_i(k) \rightarrow \beta D_s$ w.p. 1.

Additionally, consider sequence $\{W_i(k)\}$ for $k \ge \tau_s$ where
\begin{equation}\phantomsection\label{Wi-def}
W_i(k + 1) = \big(1 - \alpha_i(k)\big) W_i(k) + \alpha_i(k) w_i(k) \ ,
\end{equation}
and $W_i(\tau_s) = 0$.
By Assumption \ref{asm:F-measurable} (e) and Lemma \ref{thm:stoapp-xk-absbnd}, $\E[w_i^2(k) \mid \mathcal{F}(k)]$ is bounded w.p. 1.
Thus by Lemma 1 of \cite{stoch-approx-ql-1994}, $W_i(k) \rightarrow 0$ w.p. 1.

Now we prove that $\forall k \ge \tau_s$,
\begin{equation}\phantomsection\label{YiWi-xik-inequ}
-Y_i(k) + W_i(k) \le x_i(k) - x^*_i \le Y_i(k) + W_i(k) \ .
\end{equation}

Since $|x_i(\tau_s) - x^*_i| \le \|\bm{x}(\tau_s) - \bm{x}^*\|_\infty \le D_s = Y_i(\tau_s)$ and $W_i(\tau_s) = 0$, Inequality \eqref{YiWi-xik-inequ} holds for $k = \tau_s$. Assume Inequality \eqref{YiWi-xik-inequ} holds for some $k$, then
\begin{equation*}
\begin{aligned}
x_i(k + 1) - x^*_i &\le \big(1 - \alpha_i(k)\big)\big(Y_i(k) + W_i(k)\big) + \alpha_i(k) \beta D_s + \alpha_i(k) w_i(k) \quad(\text{by Eq. } \eqref{stoapp-x-diff-upd} \text{ and Ineq. }\eqref{Fx-xik-diff-absupd}) \\
&= \big(1 - \alpha_i(k)\big)Y_i(k) + \alpha_i(k) \beta D_s + \big(1 - \alpha_i(k)\big) W_i(k) + \alpha_i(k) w_i(k) \\
&= Y_i(k + 1) + W_i(k + 1) \quad(\text{by Eqs. } \eqref{Yi-def} \text{ and } \eqref{Wi-def}) \ , \\
x_i(k + 1) - x^*_i &\ge \big(1 - \alpha_i(k)\big)\big(-Y_i(k) + W_i(k)\big) - \alpha_i(k) \beta D_s + \alpha_i(k) w_i(k) \quad(\text{by Eq. } \eqref{stoapp-x-diff-upd} \text{ and Ineq. } \eqref{Fx-xik-diff-absupd}) \\
&= -\big(1 - \alpha_i(k)\big)Y_i(k) - \alpha_i(k) \beta D_s + \big(1 - \alpha_i(k)\big) W_i(k) + \alpha_i(k) w_i(k) \\
&= -Y_i(k + 1) + W_i(k + 1) \quad(\text{by Eqs. } \eqref{Yi-def} \text{ and } \eqref{Wi-def}) \ .
\end{aligned}
\end{equation*}
Therefore, Inequality \eqref{YiWi-xik-inequ} is true for any $k \ge \tau_s$.
Consequently,
\begin{equation}
\begin{aligned}
|x_i(k) - x^*_i - W_i(k)| &\le |Y_i(k)| \\
\Rightarrow |x_i(k) - x^*_i| &\le |Y_i(k)| + |W_i(k)| \ .
\end{aligned}
\end{equation}

Since $Y_i(k) \rightarrow \beta D_s \ (D_s > 0)$ and $W_i(k) \rightarrow 0$,
\begin{equation*}
\begin{aligned}
\exists k_{s + 1} \ge \tau_s \text{ s.t. } &\forall k \ge k_{s + 1}, |Y_i(k)| \le (\beta + {\delta \over 2}) D_s, |W_i(k)| \le {\delta \over 2} D_s \\
\Rightarrow & |x_i(k) - x^*_i| \le (\beta + \delta) D_s \quad(\forall k \ge k_{s + 1}) \ .
\end{aligned}
\end{equation*}

By $\|x_i(k) - x^*_i\|_\infty \le D_0 \ (\forall k \ge 0)$ and $0 < \beta + \delta < 1$, we have $x_i(k) \rightarrow x^*_i$ w.p. 1.
$\blacksquare$

Before proceeding to the proof of Theorem \ref{thm:geq-learn-nvmdp-convg}, we present the following lemma, which is also used in the proof.

\begin{lemma}\label{thm:sub-seq-inf-sum}
$X_i \ (i = 0, 1, 2, ...)$ are non-negative real numbers and upper bounded by $R_X$.
$\{I_i\} \ (i = 0, 1, 2, ...)$ are independent random numbers that:
\begin{equation*}
I_i \in \{0, 1\} ,\quad \sum_{i=0}^\infty {P(I_i = 1) X_i} = \infty \ .
\end{equation*}
Then $\sum_{i=0}^\infty I_i X_i = \infty$ w.p. 1.
\end{lemma}

\noindent\textbf{Proof:}
Without loss of generality, assume $R_X > 1$. Denote $p_i = P(I_i = 1)$ and
\begin{equation*}
A_0 = \{i \mid 1 \le X_i < R_X\} ,\quad A_k = \{i \mid 2^{-k} \le X_i < 2^{-k + 1}\} \quad(\forall k \ge 1) \ .
\end{equation*}
Then
\begin{equation*}
\sum_{i=0}^\infty {p_i X_i} = \sum_{k = 0}^\infty \sum_{i \in A_k} {p_i X_i} = \infty \ .
\end{equation*}

Therefore, there exists at least one set $A_u$ s.t. $\sum_{i \in A_u} {p_i X_i} = \infty$.
Moreover, $\sum_{i \in A_u} {p_i} = \infty$ since every $X_i$ is upper bounded.
Furthermore, $|A_u| = \infty$.

Denote event $E_i$ as $I_i X_i \ge 2^{-u}$.
For every $i \in A_u$, $X_i \ge 2^{-u}$, thus $E_i$ occurs if and only if $I_i = 1$ when $i \in A_u$.
As a result,
\begin{equation*}
\sum_{i \in A_u} P(E_i) = \sum_{i \in A_u} {p_i} = \infty \ .
\end{equation*}
Moreover, $E_i$ events are independent, thus infinitely many events $E_i$ will occur w.p. 1 by the second Borel-Cantelli lemma.
Consequently, $\sum_{i=0}^\infty I_i X_i = \infty$ w.p. 1.
$\blacksquare$

With the above theoretic results, we now prove the convergence of generalized Q-learning in NVMDPs.

\begin{manualtheorem}{\ref{thm:geq-learn-nvmdp-convg}}
Let $n_m(t, s, a)$ and $k$ denote, respectively, the index of the $m$-th visit to the state–action pair $(s, a)$ at time step $t$ and the total number of completed iterations of the inner for-loop in Algorithm \ref{algo:geq-learn-nvmdp} across all episodes.
If there exists constant $\bar C$ s.t.
\begin{equation*}
\sum^\infty_{m = 1} \alpha_{n_m(t, s, a)} = \infty, \ \sum^\infty_{m = 1} \alpha^2_{n_m(t, s, a)} \le \bar C \quad\textit{w.p. 1} \ (\forall 0 \le t < H, s \in S, a \in A)\ ,
\end{equation*}
and
\begin{equation*}
\lim_{k \rightarrow \infty} n_m(t, s, a) = \infty \quad\textit{w.p. 1} \ .
\end{equation*}
Then for all $1 \le i \le n, 1 \le j \le l, 0 \le t < H, s \in S, a \in A$: 
\begin{equation*}
\lim_{k \rightarrow \infty} {\hat Q}_t^{(i, j)}(s, a) = \hat{Q}^*_t(s, a) \quad\textit{w.p. 1},
\end{equation*}
where $\hat{Q}^*_t(s, a)$ equals the results of Algorithm \ref{algo:dp-valiter-nvmdp} with horizon $H$.
\end{manualtheorem}

\noindent\textbf{Proof:}
When running Algorithm \ref{algo:geq-learn-nvmdp}, all rewards of $t \ge H$ are treated as zeros.
This is equivalent to run the algorithm on another identical NVMDP with exceptions that
\begin{equation}\phantomsection\label{mdp-0-zero-rewards}
r_t(s, a, s') = 0 \quad(\forall s, s' \in S, a \in A, t \ge H) \ .
\end{equation}
We call this equivalent NVMDP \textit{MDP-1} in the following.

Moreover, consider some constant $\epsilon_0 > 0$, and
\begin{equation*}
\eta = \gamma_M + \epsilon_0 \text{ , where } \gamma_M = \max_{0 \le t \le H - 1} \max_{s, a, s'} \gamma_{t + 1}(s, a, s') \ .
\end{equation*}
The update rule in Algorithm \ref{algo:geq-learn-nvmdp} running on \textit{MDP-1} is:
\begin{equation*}
\begin{aligned}
{\hat{Q'}}^{(i)}_t(s, a) &= {\hat{Q}}^{(i, 1)}_t(s, a) + \alpha_i(k) \Big(r_t(s, a, s') + \gamma_{t + 1}(s, a, s') {f_i}\big(\bm{\hat{Q}}_{t + 1}(s', \cdot)\big) - {\hat{Q}}^{(i, 1)}_t(s, a)\Big) \\
\Leftrightarrow \eta^t {\hat{Q'}}^{(i)}_t(s, a) &= \eta^t {\hat{Q}}^{(i, 1)}_t(s, a) + \alpha_i(k) \Big(\eta^t r_t(s, a, s') + {\gamma_{t + 1}(s, a, s') \over \eta} \eta^{t + 1} {f_i}\big(\bm{\hat{Q}}_{t + 1}(s', \cdot)\big) - \eta^t {\hat{Q}}^{(i, 1)}_t(s, a)\Big) \ ,
\end{aligned}
\end{equation*}
where $k$ is the index of the current update, $\alpha_i(k)$ is the corresponding stepsize, ${\hat{Q'}}^{(i)}_t(s, a)$ in the L.H.S represents the next action-value estimate of $(s, a)$ in track $i$ at time $t$.

Consider another stationary MDP---which is called \textit{MDP-2}---with state space $\widetilde{S} = \{(t, s) \mid t \in \{0, 1, ..., H\}, s \in S\}$, action space $\widetilde{A} = A$, and for $0 \le t \le H - 1$,
\begin{equation}\phantomsection\label{mdp1-to-mdp2}
\begin{aligned}
\widetilde{p}\big((t + 1, s') | (t, s), a\big) &= p_t(s' | s, a) \\
\widetilde{r}\big((t, s), a, (t + 1, s')\big) &= \eta^t r_t(s, a, s') \\
\widetilde{\gamma}\big((t, s), a, (t + 1, s')\big) &= {1 \over \eta} \gamma_{t + 1}(s, a, s') \ .
\end{aligned}
\end{equation}
Denote $\beta = \gamma_M / \eta$, then $0 \le \beta < 1$ and
\begin{equation}\phantomsection\label{mdp2-gamma-upb}
\widetilde{\gamma}\big((t, s), a, (t + 1, s')\big) \le \beta \quad(\forall (t, s), (t + 1, s') \in \widetilde{S}, a \in \widetilde{A})
\end{equation}

Moreover, denote
\begin{equation}\phantomsection\label{mdp1-to-mdp2-Qaf}
\begin{aligned}
{\widetilde{Q}}^{(i, j)}\big((t, s), a\big) &= \eta^t {\hat{Q}}^{(i, j)}_t(s, a) \\
{{\widetilde{f}}_i}\big(\bm{\widetilde{Q}}\big((t, s), \cdot\big)\big) &= \eta^t {f_i}\big({1 \over \eta^t} \bm{\widetilde{Q}}\big((t, s), \cdot\big)\big) = \eta^t {f_i}\big(\bm{\hat Q}_t(s, \cdot)\big) \ .
\end{aligned}
\end{equation}
for all $s, s' \in S, a \in A, 0 \le t \le H$.
Therefore, the update rule becomes:
\begin{equation}\phantomsection\label{geq-learn-mdp2-upd}
\begin{aligned}
{\widetilde{Q'}}^{(i)}\big((t, s), a\big) &= {\widetilde{Q}}^{(i, 1)}\big((t, s), a\big) +
\alpha_i(k) \bigg(\widetilde{r}\big((t, s), a, (t + 1, s')\big) \\
&+ \widetilde{\gamma}\big((t, s), a, (t + 1, s')\big) {{\widetilde{f}}_i}\Big(\bm{\widetilde{Q}}\big((t + 1, s'), \cdot\big)\Big) - {\widetilde{Q}}^{(i, 1)}\big((t, s), a\big)\bigg) \ ,
\end{aligned}
\end{equation}
for $0 \le t \le H - 1$ in \textit{MDP-2} (all ${\widetilde{Q}}^{(i, j)}\big((t, s), a\big) = 0$ for $t = H$).
Thus, running Algorithm \ref{algo:geq-learn-nvmdp} on \textit{MDP-1} is equivalent to running the algorithm on \textit{MDP-2} with update rule \eqref{geq-learn-mdp2-upd}.
Futhermore, it is straightforward to verify that $f_i$ satisfies Assumptions \ref{asm:geq-f-maxq-cond} and \ref{asm:geq-f-absbnd-cond} if and only if ${\widetilde{f}}_i$ satisfies these two assumptions.

By Equation \eqref{mdp-0-zero-rewards}, we only need to consider sub-policies for $0 \le t \le H - 1$ (i.e., $\pi = \{\pi_0, \pi_1, ... , \pi_{H - 1}\}$) in \textit{MDP-1} since all rewards for $t \ge H$ are zeroes.
Futhermore, for any policy $\pi = \{\pi_0, \pi_1, ... , \pi_{H - 1}\}$ in \textit{MDP-1}, we can infer a corresponding stationary policy in \textit{MDP-2}, and vice versa.

By Equation \eqref{return-nvmdp}, \eqref{Gamma-def}, and \eqref{av-nvmdp},
\begin{equation*}
\begin{aligned}
Q^\pi_t(s, a) &= \E^\pi[\sum_{i = t}^{\infty} \big(\prod_{j = t + 1}^{i}\gamma_j(s_{j - 1}, a_{j - 1}, s_j)\big) r_i(s_i, a_i, s_{i + 1}) \mid s_t = s, a_t = a] \\
\Rightarrow \eta^t Q^\pi_t(s, a) &= \E^\pi[\sum_{i = t}^{\infty} \big(\prod_{j = t + 1}^{i}{\gamma_j(s_{j - 1}, a_{j - 1}, s_j) \over \eta}\big) \eta^i r_i(s_i, a_i, s_{i + 1}) \mid s_t = s, a_t = a] \\
&= \E^\pi\Big[\sum_{i = t}^{\infty} \Big(\prod_{j = t + 1}^{i}{\widetilde{\gamma}\big((j - 1, s_{j - 1}), a_{j - 1}, (j, s_j)\big)}\Big) \widetilde{r}\big((i, s_i), a_i, (i + 1, s_{i + 1})\big) \mid s_t = s, a_t = a\Big] \\
&= {\widetilde{Q}}^\pi\big((t, s), a\big)
\end{aligned}
\end{equation*}

By the above equation, for any $t \in [0, H), s \in S, a \in A$ and policy $\pi$ in \textit{MDP-1},
\begin{equation*}
\eta^t Q^\pi_t(s, a) \le {\widetilde{Q}}^*\big((t, s), a\big) \Rightarrow \eta^t Q^*_t(s, a) \le {\widetilde{Q}}^*\big((t, s), a\big) \ .
\end{equation*}
Moreover, for any $t \in [0, H), s \in S, a \in A$ and stationary policy $\pi$ in \textit{MDP-2},
\begin{equation*}
\begin{aligned}[b]
{\widetilde{Q}}^\pi\big((t, s), a\big) \le \eta^t Q^*_t(s, a) \Rightarrow {\widetilde{Q}}^*\big((t, s), a\big) \le \eta^t Q^*_t(s, a) \ ,
\end{aligned}
\end{equation*}
where the second inequality is by Theorem \ref{thm:optim-pol-nvmdp} and Theorem \ref{thm:station-nvmdp-is-classic} that \textit{MDP-2} has one optimal policy which is also stationary.
Combining the results above gives
\begin{equation}\phantomsection\label{mdp1-to-mdp2-optimav}
{\widetilde{Q}}^*\big((t, s), a\big) = \eta^t Q^*_t(s, a) \quad(\forall t \in [0, H), s \in S, a \in A) \ .
\end{equation}
Therefore, if action-value estimates ${\widetilde{Q}}^{(i, j)}\big((t, s), a\big)$ converge to ${\widetilde{Q}}^*\big((t, s), a\big)$ in \textit{MDP-2} w.p. 1, then by Equation \eqref{mdp1-to-mdp2-Qaf} and Equation \eqref{mdp1-to-mdp2-optimav} the action-value estimates ${\hat{Q}}^{(i, j)}_t(s, a)$ converge to $Q^*_t(s, a)$ in \textit{MDP-1} w.p. 1.

Let $x, x'$ deonte $(t, s), (t + 1, s')$, respectively.
The update rule \eqref{geq-learn-mdp2-upd} in \textit{MDP-2} thus becomes:
\begin{equation}\phantomsection\label{mdp2-Q-upd}
{\widetilde{Q'}}^{(i)}(x, a) = {\widetilde{Q}}^{(i, 1)}(x, a) +
\alpha_i(k) \big(F^{(x, a, i)}(\bm{\widetilde{Q}}_k) - {\widetilde{Q}}^{(i, 1)}(x, a) + \widetilde{w}_i(k) \big) \ ,
\end{equation}
where
\begin{equation*}
\begin{aligned}
F^{(x, a, i)}(\bm{\widetilde{Q}}_k) &= \E[\widetilde{r}(x, a, x') + \widetilde{\gamma}(x, a, x') {{\widetilde{f}}_i}\big(\bm{\widetilde{Q}}(x', \cdot)\big)] \\
\widetilde{w}_i(k) &= \widetilde{r}(x, a, x') + \widetilde{\gamma}(x, a, x') {{\widetilde{f}}_i}\big(\bm{\widetilde{Q}}(x', \cdot)\big) - F^{(x, a, i)}(\bm{\widetilde{Q}}_k) \\
&= w_I + w_{II} \\
w_I &= \widetilde{r}(x, a, x') - \E[\widetilde{r}(x, a, x')] \\
w_{II} &= \widetilde{\gamma}(x, a, x') {{\widetilde{f}}_i}\big(\bm{\widetilde{Q}}(x', \cdot)\big) - \E[\widetilde{\gamma}(x, a, x') {{\widetilde{f}}_i}\big(\bm{\widetilde{Q}}(x', \cdot)\big)] \ .
\end{aligned}
\end{equation*}
Here we use $\E$ to denote $\E^{x' \sim \widetilde{p}(\cdot | x, a)}$, with the conditioning on $\mathcal{F}(k)$ omitted for brevity, and continue this notation until the end of the proof.

In order to apply Lemma \ref{thm:stoapp-xk-convg}, we must verify that Assumptions \ref{asm:tau-go-infty}, \ref{asm:F-measurable}, \ref{asm:stoapp-stepsize}, and \ref{asm:stoapp-contract} are satisfied.
In Equation \eqref{mdp2-Q-upd}, the terms ${\widetilde{Q}}^{(i, j)}(x, a)$, $F^{(x, a, i)}(\bm{\widetilde{Q}}_k)$, and $\widetilde{w}_i(k)$ correspond, respectively, to $x_i(k)$, $F_i(\bm{X}_k^i)$, and $w_i(k)$ in Equation \eqref{stoapp-x-upd}.
More concretely, at update step $k$, all values ${\widetilde{Q}}^{(i, j)}(x, a)$ form a $(|\widetilde{S}| \cdot |A| \cdot n) \times l$ matrix, where the row is indexed by $(x, a, i)$ and the column by $j$. In particular, ${\widetilde{Q}}^{(i, 1)}(x, a)$ corresponds to $x_i(k)$, while other ${\widetilde{Q}}^{(i, j)}(x, a)$ values (for $j > 1$) correspond to the elements of the matrix $\bm{X}_k^i$ defined in Equation \eqref{stoapp-Xik-def}.

By $\lim_{k \rightarrow \infty} n_m(t, s, a) = \infty$ w.p. 1 and each estimation track is selected uniformly at random, each ${\widetilde{Q}}^{(i, j)}(x, a)$ will be udpated infinitely often w.p. 1, therefore Assumption \ref{asm:tau-go-infty} is satisfied.

For Assumption \ref{asm:F-measurable}, parts (a), (b), and (c) are straightforward to verify.
Since
\begin{equation*}
\begin{aligned}
\E[w_I] &= \E[w_{II}] = 0 \\
\E[w_i(k)] &= \E[w_I] + \E[w_{II}] = 0 \\
\E[w_i^2(k)] &\le 2\E[w_I^2] + 2\E[w_{II}^2] \\
&= 2\Var[\widetilde{r}(x, a, x')] + 2\Var[\widetilde{\gamma}(x, a, x') {{\widetilde{f}}_i}\big(\bm{\widetilde{Q}}(x', \cdot)\big)] \\
&\le 2\E[{\widetilde{r}}^2(x, a, x')] + 2\E[{\widetilde{\gamma}}^2(x, a, x') {{\widetilde{f}}_i^2}\big(\bm{\widetilde{Q}}(x', \cdot)\big)] \\
&\le 2(\max\{1, \eta^{H - 1}\} R_B)^2 + 2 \beta^2 \E[{{\widetilde{f}}_i^2}\big(\bm{\widetilde{Q}}(x', \cdot)\big)] \ ,
\end{aligned}
\end{equation*}
where $|r_t(\cdot)| \le R_B$ by Assumption \ref{asm:rwd-absbnd-nvmdp} and $\big|{{\widetilde{f}}_i}\big(\bm{\widetilde{Q}}(x', \cdot)\big)\big| \le \max_{a, i, j} |{\widetilde{Q}}^{(i, j)}(x, a)|$ by Assumption \ref{asm:geq-f-absbnd-cond},
part (d) and (e) are also satisfied.

Denote the stepsize as $\alpha^{(x, a, i)}(k)$ in Equation \eqref{mdp2-Q-upd} for update step $k$ with station-action pair $(x, a)$ in the $i$-th track ($x = (t, s)$).
As $\sum^\infty_{m = 1} \alpha^2_{n_m(t, s, a)} \le \bar C$ in \textit{MDP-1} almost surely, $\sum^\infty_{k = 0} \big(\alpha^{(x, a, i)}(k)\big)^2 \le \bar C$ in in \textit{MDP-2} w.p. 1.
Moreover, since $\sum^\infty_{m = 1} \alpha_{n_m(t, s, a)} = \infty$ in \textit{MDP-1} almost surely,
$\sum^\infty_{k = 0} \alpha^{(x, a, i)}(k) = \infty$ in \textit{MDP-2} w.p. 1 by Lemma \ref{thm:sub-seq-inf-sum}.
Assumption \ref{asm:stoapp-stepsize} is therefore satisfied.

Consider ${\bm{\widetilde{Q}}}^* \in \mathbb{R}^{(|\widetilde{S}| \cdot |A| \cdot n) \times l}$,
where $\forall x \in \widetilde{S}, a \in A, i \in \{1, 2, ..., n\}, j \in \{1, 2, ..., l\}$,
\begin{equation*}
\big({\bm{\widetilde{Q}}}^*\big)_{(x, a, i), j} = {\widetilde{Q}}^*(x, a) \ .
\end{equation*}
Then by Assumption \ref{asm:geq-f-maxq-cond},
\begin{equation*}
\begin{aligned}
{{\widetilde{f}}_i}\big(\bm{\widetilde{Q}}^*(x', \cdot)\big) &= \max_{a'} {\widetilde{Q}}^*(x', a') \\
\Rightarrow F^{(x, a, i)}(\bm{\widetilde{Q}}^*) &= \E[\widetilde{r}(x, a, x') + \widetilde{\gamma}(x, a, x') \max_{a'} {\widetilde{Q}}^*(x', a')] = {\widetilde{Q}}^*(x, a)
\end{aligned}
\end{equation*}
Denote $F_A(\bm{\widetilde{Q}}) = \big[F^{(x, a, i)}(\bm{\widetilde{Q}}) \bm{1}\big]$ for $\bm{\widetilde{Q}} \in \mathbb{R}^{(|\widetilde{S}| \cdot |A| \cdot n) \times l}$ and $\bm{1} = (1, 1, ..., 1) \in \mathbb{R}^l$.
Then
\begin{equation*}
F_A(\bm{\widetilde{Q}}^*) = \bm{\widetilde{Q}}^* \ .
\end{equation*}
Consider any $\bm{\widetilde{Q}}, \bm{\widetilde{Q'}} \in \mathbb{R}^{(|\widetilde{S}| \cdot |A| \cdot n) \times l}$ and their corresponding $\bm{\widetilde{Q}}(\cdot), \bm{\widetilde{Q'}}(\cdot) \in \mathbb{R}^{|\widetilde{S}| \times |A| \times n \times l}$:
\begin{equation*}
\bm{\widetilde{Q}}^{(i, j)}(x, a) = \big({\bm{\widetilde{Q}}}\big)_{(x, a, i), j} \ , \bm{\widetilde{Q'}}^{(i, j)}(x, a) = \big({\bm{\widetilde{Q'}}}\big)_{(x, a, i), j} \quad(\forall x, a, i, j)
\end{equation*}
\begin{equation*}
\begin{aligned}
F^{(x, a, i)}(\bm{\widetilde{Q}}) - F^{(x, a, i)}(\bm{\widetilde{Q'}}) &= \E\big[\widetilde{\gamma}(x, a, x') \big({{\widetilde{f}}_i}\big(\bm{\widetilde{Q}}(x', \cdot)\big) - {{\widetilde{f}}_i}\big(\bm{\widetilde{Q'}}(x', \cdot)\big)\big] \\
\Rightarrow \Big|F^{(x, a, i)}(\bm{\widetilde{Q}}) - F^{(x, a, i)}(\bm{\widetilde{Q'}})\big| &\le \E\Big[\big|\widetilde{\gamma}(x, a, x') \big({{\widetilde{f}}_i}\big(\bm{\widetilde{Q}}(x', \cdot)\big) - {{\widetilde{f}}_i}\big(\bm{\widetilde{Q'}}(x', \cdot)\big)\big|\Big] \\
&= \E\Big[\big|\widetilde{\gamma}(x, a, x')\big| \big|{{\widetilde{f}}_i}\big(\bm{\widetilde{Q}}(x', \cdot)\big) - {{\widetilde{f}}_i}\big(\bm{\widetilde{Q'}}(x', \cdot)\big|\Big] \\
&\le \beta \E\Big[\big|{{\widetilde{f}}_i}\big(\bm{\widetilde{Q}}(x', \cdot)\big) - {{\widetilde{f}}_i}\big(\bm{\widetilde{Q'}}(x', \cdot)\big|\Big] \quad(\text{by Inequ. }\eqref{mdp2-gamma-upb})\ .
\end{aligned}
\end{equation*}
By Assumption \ref{asm:geq-f-absbnd-cond},
\begin{equation*}
\begin{aligned}
\Big|F^{(x, a, i)}(\bm{\widetilde{Q}}) - F^{(x, a, i)}(\bm{\widetilde{Q'}})\big| &\le \beta \E\Big[\max_{a', i', j'} \big|\bm{\widetilde{Q}}^{(i', j')}(x', a') - \bm{\widetilde{Q'}}^{(i', j')}(x', a')\big|\Big] \\
&\le \beta \|\bm{\widetilde{Q}} - \bm{\widetilde{Q'}}\|_{\max} \quad(\forall x, a, i) \\
\Rightarrow \|F_A(\bm{\widetilde{Q}}) - F_A(\bm{\widetilde{Q'}})\|_{\max} &\le \beta \|\bm{\widetilde{Q}} - \bm{\widetilde{Q'}}\|_{\max} \\
\Rightarrow \|F_A(\bm{\widetilde{Q}}) - {\bm{\widetilde{Q}}}^*\|_{\max} &\le \beta \|\bm{\widetilde{Q}} - {\bm{\widetilde{Q}}}^*\|_{\max} \quad\big(F_A(\bm{\widetilde{Q}}^*) = \bm{\widetilde{Q}}^*\big) \ .
\end{aligned}
\end{equation*}
Thus Assumption \ref{asm:stoapp-contract} is satisfied.
By Lemma \ref{thm:stoapp-xk-convg}, ${\widetilde{Q}}^{(i, j)}\big((t, s), a\big)$ converges to ${\widetilde{Q}}^*\big((t, s), a\big)$ in \textit{MDP-2} for any $s \in S, a \in A, 0 \le t \le H - 1$ w.p. 1, which completes the proof.
$\blacksquare$

\begin{manuallemma}{\ref{thm:maxmin-q-learn-inequal}}
For any $\bm{Q}(s, \cdot), \bm{Q'}(s, \cdot) \in {\mathbb R}^{|A| \times n \times l}$ and any non-empty set $F \subseteq \{1, 2, ..., n\} \times \{1, 2, ..., l\}$,
\begin{equation*}
\big|\max_{a \in A} \min_{(i, j) \in F} Q^{(i, j)}(s, a) - \max_{a \in A} \min_{(i, j) \in F} {Q'}^{(i, j)}(s, a)\big| \le \max_{a \in A, (i, j) \in F} \big| Q^{(i, j)}(s, a) - {Q'}^{(i, j)}(s, a) \big| \ .
\end{equation*}
\end{manuallemma}

\noindent\textbf{Proof:}
Assume $\max_{a \in A} \min_{(i, j) \in F} Q^{(i, j)}(s, a)$ is reached by $\bar{a}, \bar{i}, \bar{j}$, $\max_{a \in A} \min_{(i, j) \in F} {Q'}^{(i, j)}(s, a)$ is reached by $\widetilde{a}, \widetilde{i}, \widetilde{j}$.
Consider $(i_0, j_0) \in \text{argmin}_{(i, j) \in F}{Q'}^{(i, j)}(s, \bar{a})$, then
\begin{equation*}
\begin{aligned}
&Q^{(\bar{i}, \bar{j})}(s, \bar{a}) - {Q'}^{(\widetilde{i}, \widetilde{j})}(s, \widetilde{a}) \\
&= \big(Q^{(\bar{i}, \bar{j})}(s, \bar{a}) - Q^{(i_0, j_0)}(s, \bar{a})\big) + \big(Q^{(i_0, j_0)}(s, \bar{a}) - {Q'}^{(i_0, j_0)}(s, \bar{a})\big) + \big({Q'}^{(i_0, j_0)}(s, \bar{a}) - {Q'}^{(\widetilde{i}, \widetilde{j})}(s, \widetilde{a})\big) \\
&\le Q^{(i_0, j_0)}(s, \bar{a}) - {Q'}^{(i_0, j_0)}(s, \bar{a}) \le \max_{a \in A, (i, j) \in F} \big| Q^{(i, j)}(s, a) - {Q'}^{(i, j)}(s, a) \big| \ .
\end{aligned}
\end{equation*}
Similarly, consider $(i_1, j_1) \in \text{argmin}_{(i, j) \in F}Q^{(i, j)}(s, \widetilde{a})$,
\begin{equation*}
\begin{aligned}
&Q^{(\bar{i}, \bar{j})}(s, \bar{a}) - {Q'}^{(\widetilde{i}, \widetilde{j})}(s, \widetilde{a}) \\
&= \big(Q^{(\bar{i}, \bar{j})}(s, \bar{a}) - Q^{(i_1, j_1)}(s, \widetilde{a})\big) + \big(Q^{(i_1, j_1)}(s, \widetilde{a}) - {Q'}^{(i_1, j_1)}(s, \widetilde{a})\big) + \big({Q'}^{(i_1, j_1)}(s, \widetilde{a}) - {Q'}^{(\widetilde{i}, \widetilde{j})}(s, \widetilde{a})\big) \\
&\ge Q^{(i_1, j_1)}(s, \widetilde{a}) - {Q'}^{(i_1, j_1)}(s, \widetilde{a}) \ge -\max_{a \in A, (i, j) \in F} \big| Q^{(i, j)}(s, a) - {Q'}^{(i, j)}(s, a) \big| \ .
\end{aligned}
\end{equation*}

Combining the inequalities above establishes the lemma.
$\blacksquare$

For each theorem and lemma in Section \ref{sec:func-approx}, we present two proofs---one using a scalar representation and the other a matrix representation---except for Lemma \ref{thm:D-oper-trans}.

\begin{manualtheorem}{\ref{thm:pol-grad-nvmdp}}
For a NVMDP and its parameterized policy $\pi$,
\begin{equation*}
\nabla V^{\pi}_t(s_t) = \E^\pi [\sum_{i = t}^\infty \Gamma^\tau_{t, i} A^\pi_i(s_i, a_i) {\nabla \pi_i(a_i | s_i) \over \pi_i(a_i | s_i)} \mid s_t] \ . \tag{\refeq{pol-grad-nvmdp}}
\end{equation*}
\end{manualtheorem}

\noindent\textbf{Proof 1:}
By Equation \eqref{sv-av-nvmdp},
\begin{equation}\phantomsection\label{vs-grd}
\begin{aligned}[b]
\nabla V^{\pi}_t(s_t) &= \nabla \sum_{a_t} \pi_t(a_t | s_t) Q^{\pi}_t(s_t, a_t) \\
&= \sum_{a_t} \bigl((\nabla \pi_t(a_t | s_t)) Q^{\pi}_t(s_t, a_t) + \pi_t(a_t | s_t) \nabla Q^{\pi}_t(s_t, a_t) \bigr) \ .
\end{aligned}
\end{equation}

By Equation \eqref{av-iter-nvmdp},
\begin{equation}\phantomsection\label{qs-grd}
\begin{aligned}[b]
\nabla Q^{\pi}_t(s_t, a_t) &= \nabla \sum_{s_{t + 1}, r_t} p_t(s_{t + 1}, r_t | s_t, a_t) (r_t(s_t, a_t, s_{t + 1}) + \gamma_{t + 1}V^{\pi}_{t + 1}(s_{t + 1}))  \\
&= \nabla \sum_{s_{t + 1}, r_t} p_t(s_{t + 1}, r_t | s_t, a_t) \gamma_{t + 1}V^{\pi}_{t + 1}(s_{t + 1}) \\
&= \nabla \sum_{s_{t + 1}} p_t(s_{t + 1} | s_t, a_t) \gamma_{t + 1}V^{\pi}_{t + 1}(s_{t + 1}) \\
&= \sum_{s_{t + 1}} p_t(s_{t + 1} | s_t, a_t) \gamma_{t + 1}\nabla V^{\pi}_{t + 1}(s_{t + 1}) \ .
\end{aligned}
\end{equation}
($\gamma_{t + 1}(s_t, a_t, s_{t + 1})$ is abbreviated as $\gamma_{t + 1}$ for convenience, same as below.)

Combine Equation \eqref{vs-grd} and Equation \eqref{qs-grd}:
\begin{equation}\phantomsection\label{vs-grd-iter}
\nabla V^{\pi}_t(s_t) = \sum_{a_t} (\nabla \pi_t(a_t | s_t)) Q^{\pi}_t(s_t, a_t) + \sum_{a_t, s_{t + 1}} \pi_t(a_t | s_t) p_t(s_{t + 1} | s_t, a_t) \gamma_{t + 1} \nabla V^{\pi}_{t + 1}(s_{t + 1}) \ .
\end{equation}

Since
\begin{equation*}
\sum_{a_t} (\nabla \pi_t(a_t | s_t)) V^{\pi}_t(s_t) = V^{\pi}_t(s_t) \sum_{a_t} \nabla \pi_t(a_t | s_t) = V^{\pi}_t(s_t) \nabla \sum_{a_t} \pi_t(a_t | s_t) = 0 \ ,
\end{equation*}

The first item in R.H.S of Equation \eqref{vs-grd-iter} becomes:
\begin{equation*}
\begin{aligned}
\sum_{a_t} (\nabla \pi_t(a_t | s_t)) Q^{\pi}_t(s_t, a_t) &= \sum_{a_t} \nabla \pi_t(a_t | s_t) \bigl(Q^{\pi}_t(s_t, a_t) - V^{\pi}_t(s_t)\bigr) \\
&= \sum_{a_t} (\nabla \pi_t(a_t | s_t)) A^\pi_t(s_t, a_t) \\
&= \sum_{a_t} \pi_t(a_t | s_t) {\nabla \pi_t(a_t | s_t) \over \pi_t(a_t | s_t)} A^\pi_t(s_t, a_t) \\
&= \E^\pi [A^\pi_t(s_t, a_t) {\nabla \pi_t(a_t | s_t) \over \pi_t(a_t | s_t)} \mid s_t] \ .
\end{aligned}
\end{equation*}

Thus Equation \eqref{vs-grd-iter} is transformed into:
\begin{equation}\phantomsection\label{vs-grd-iter-ex}
\begin{aligned}[b]
\nabla V^{\pi}_t(s_t) = \E^\pi [A^\pi_t(s_t, a_t) {\nabla \pi_t(a_t | s_t) \over \pi_t(a_t | s_t)} \mid s_t] + \sum_{a_t, s_{t + 1}} \pi_t(a_t | s_t) p_t(s_{t + 1} | s_t, a_t) \gamma_{t + 1} \nabla V^{\pi}_{t + 1}(s_{t + 1}) \ .
\end{aligned}
\end{equation}

By Equation \eqref{vs-grd-iter-ex}:
\begin{equation}\phantomsection\label{rhs-2nd-item}
\begin{aligned}[b]
&\sum_{a_t, s_{t + 1}} \pi_t(a_t | s_t) p_t(s_{t + 1} | s_t, a_t) \gamma_{t + 1} \nabla V^{\pi}_{t + 1}(s_{t + 1}) \\
&=\sum_{a_t, s_{t + 1}} \pi_t(a_t | s_t) p_t(s_{t + 1} | s_t, a_t) \gamma_{t + 1}  \E^\pi [A^\pi_{t + 1}(s_{t + 1}, a_{t + 1}) {\nabla \pi_{t + 1}(a_{t + 1} | s_{t + 1}) \over \pi_{t + 1}(a_{t + 1} | s_{t + 1})} \mid s_{t + 1}] \\
&+ \sum_{a_t, s_{t + 1}} \pi_t(a_t | s_t) p_t(s_{t + 1} | s_t, a_t) \gamma_{t + 1} \sum_{a_{t + 1}, s_{t + 2}} \pi_{t + 1}(a_{t + 1} | s_{t + 1}) p_t(s_{t + 2} | s_{t + 1}, a_{t + 1}) \gamma_{t + 2} \nabla V^{\pi}_{t + 2}(s_{t + 2})
\end{aligned}
\end{equation}

The first item in R.H.S of Equation \eqref{rhs-2nd-item} is:
\begin{equation*}
\begin{aligned}
&\sum_{a_t, s_{t + 1}} \pi_t(a_t | s_t) p_t(s_{t + 1} | s_t, a_t) \Gamma^\tau_{t, t + 1} \E^\pi [A^\pi_{t + 1}(s_{t + 1}, a_{t + 1}) {\nabla \pi_{t + 1}(a_{t + 1} | s_{t + 1}) \over \pi_{t + 1}(a_{t + 1} | s_{t + 1})} \mid s_{t + 1}, a_t, s_t] \\
&= \sum_{a_t, s_{t + 1}} \pi_t(a_t | s_t) p_t(s_{t + 1} | s_t, a_t) \E^\pi [\Gamma^\tau_{t, t + 1} A^\pi_{t + 1}(s_{t + 1}, a_{t + 1}) {\nabla \pi_{t + 1}(a_{t + 1} | s_{t + 1}) \over \pi_{t + 1}(a_{t + 1} | s_{t + 1})} \mid s_{t + 1}, a_t, s_t] \\
&= \E^{a_t, s_{t + 1} \sim \pi} \Bigl[ \E^{a_{t + 1}, s_{t + 2}, ... \sim \pi} [\Gamma^\tau_{t, t + 1} A^\pi_{t + 1}(s_{t + 1}, a_{t + 1}) {\nabla \pi_{t + 1}(a_{t + 1} | s_{t + 1}) \over \pi_{t + 1}(a_{t + 1} | s_{t + 1})} \mid s_{t + 1}, a_t, s_t] \mid s_t \Bigr] \\
&= \E^{a_t, s_{t + 1}, a_{t + 1}, s_{t + 2}, ... \sim \pi} [\Gamma^\tau_{t, t + 1} A^\pi_{t + 1}(s_{t + 1}, a_{t + 1}) {\nabla \pi_{t + 1}(a_{t + 1} | s_{t + 1}) \over \pi_{t + 1}(a_{t + 1} | s_{t + 1})} \mid s_t] \\
&= \E^\pi [\Gamma^\tau_{t, t + 1} A^\pi_{t + 1}(s_{t + 1}, a_{t + 1}) {\nabla \pi_{t + 1}(a_{t + 1} | s_{t + 1}) \over \pi_{t + 1}(a_{t + 1} | s_{t + 1})} \mid s_t] \ .
\end{aligned}
\end{equation*}

The second item in R.H.S of Equation \eqref{rhs-2nd-item} is:
\begin{equation*}
\begin{aligned}
\sum_{a_t, s_{t + 1}, a_{t + 1}, s_{t + 2}} \pi_t(a_t | s_t) p_t(s_{t + 1} | s_t, a_t) \pi_{t + 1}(a_{t + 1} | s_{t + 1}) p_t(s_{t + 2} | s_{t + 1}, a_{t + 1}) \Gamma^\tau_{t, t + 2} \nabla V^{\pi}_{t + 2}(s_{t + 2}) \ .
\end{aligned}
\end{equation*}

Substitute the above results into R.H.S of Equation \eqref{vs-grd-iter-ex}:
\begin{equation*}
\begin{aligned}[b]
&\nabla V^{\pi}_t(s_t) = \sum_{i = t}^{t + 1} \E^\pi [\Gamma^\tau_{t, i} A^\pi_i(s_i, a_i) {\nabla \pi_i(a_i | s_i) \over \pi_i(a_i | s_i)} \mid s_t] \\
&+ \sum_{a_t, s_{t + 1}, a_{t + 1}, s_{t + 2}} \pi_t(a_t | s_t) p_t(s_{t + 1} | s_t, a_t) \pi_{t + 1}(a_{t + 1} | s_{t + 1}) p_t(s_{t + 2} | s_{t + 1}, a_{t + 1}) \Gamma^\tau_{t, t + 2} \nabla V^{\pi}_{t + 2}(s_{t + 2}) \ .
\end{aligned}
\end{equation*}

Continue the recursion:
\begin{equation*}
\nabla V^{\pi}_t(s_t) = \sum_{i = t}^\infty \E^\pi [\Gamma^\tau_{t, i} A^\pi_i(s_i, a_i) {\nabla \pi_i(a_i | s_i) \over \pi_i(a_i | s_i)} \mid s_t] = \E^\pi [\sum_{i = t}^\infty \Gamma^\tau_{t, i} A^\pi_i(s_i, a_i) {\nabla \pi_i(a_i | s_i) \over \pi_i(a_i | s_i)} \mid s_t] \ .\ \blacksquare
\end{equation*}

\noindent\textbf{Proof 2:}
Let $\partial$ denotes partial derivative to one parameter.
By Equation \eqref{sv-selfiter-nvmdp-mat},
\begin{equation}\phantomsection\label{svderiv-selfiter-nvmdp-mat}
\begin{aligned}[b]
\partial \bm{V}^{\pi}_t &= (\partial \bm{\Pi}^\pi_t) \bm{r}_t + (\partial \bm{L}_{t + 1}^\pi) \bm{V}^{\pi}_{t + 1} + \bm{L}_{t + 1}^\pi (\partial \bm{V}^{\pi}_{t + 1}) \\
&= (\partial \bm{\Pi}^\pi_t)(\bm{r}_t + \bm{J}_{t + 1} \bm{V}^{\pi}_{t + 1}) + \bm{L}_{t + 1}^\pi (\partial \bm{V}^{\pi}_{t + 1}) \quad(\partial \bm{L}_{t + 1}^\pi = (\partial \bm{\Pi}^\pi_t) \bm{J}_{t + 1} \text{ by Eq. } \eqref{l-def-nvmdp-mat}) \\
&= (\partial \bm{\Pi}^\pi_t) \bm{Q}^{\pi}_t + \bm{L}_{t + 1}^\pi (\partial \bm{V}^{\pi}_{t + 1}) \quad(\text{by Eq. } \eqref{av-iter-nvmdp-mat}) \ .
\end{aligned}
\end{equation}

Applying Equation \eqref{svderiv-selfiter-nvmdp-mat} recursively,
\begin{equation*}
\partial \bm{V}^{\pi}_t = \sum_{i = t}^\infty \big(\prod_{j = t + 1}^i \bm{L}_j^\pi\big) {(\partial \bm{\Pi}^\pi_i) \bm{Q}^{\pi}_i} \ .
\end{equation*}
Moreover, by Equation \eqref{adv-nvmdp-mat} and Equation \eqref{piu-relp-nvmdp-mat},
\begin{equation*}
\begin{aligned}
(\partial \bm{\Pi}^\pi_i) \bm{U} = \bm{O} \Rightarrow (\partial \bm{\Pi}^\pi_i) \bm{Q}_i^\pi = (\partial \bm{\Pi}^\pi_i) \bm{A}_i^\pi + (\partial \bm{\Pi}^\pi_i) \bm{U} \bm{V}_i^\pi = (\partial \bm{\Pi}^\pi_i) \bm{A}_i^\pi \ ,
\end{aligned}
\end{equation*}
Thus,
\begin{equation}\phantomsection\label{svderiv-recurs-nvmdp-mat}
\partial \bm{V}^{\pi}_t = \sum_{i = t}^\infty \big(\prod_{j = t + 1}^i \bm{L}_j^\pi\big) {(\partial \bm{\Pi}^\pi_i) \bm{A}^{\pi}_i} \ .
\end{equation}

Let $r_i(s_i, a_i, s_{i + 1}) = \bar{r}_i(s_i, a_i) = r_i(s_i, a_i)$, then by Lemma \ref{thm:sv-av-avgrwd-nvmdp} and Equation \eqref{sv-recurs-nvmdp-mat},
\begin{equation*}
\Big(\sum_{i = t}^\infty \big(\prod_{j = t + 1}^i \bm{L}_j^\pi\big) {\bm{\Pi}^\pi_i \bm{r}_i}\Big)_s = \E^\pi [\sum_{i = t}^{\infty}\Gamma^\tau_{t, i} (\bm{r}_i)_{(s_i, a_i)} \mid s_t = s]
\end{equation*}
holds for all $\bm{r}_i$ s.t. $\|\bm{r}_i\|_\infty \le R_B$.
Furthermore, set $\bm{r}_u = \bm{0}$ for all $u \ge t, u \ne i$:
\begin{equation}\phantomsection\label{sv-recurs-nvmdp-mat2scar}
\Big(\big(\prod_{j = t + 1}^i \bm{L}_j^\pi\big) {\bm{\Pi}^\pi_i \bm{r}_i}\Big)_s = \E^\pi [\Gamma^\tau_{t, i} (\bm{r}_i)_{(s_i, a_i)} \mid s_t = s] \ .
\end{equation}

Let $\lambda$ denotes a sufficient small positive real number, and set
\begin{equation*}
(\bm{r}_i)_{(s_i, a_i)} = \lambda \sum_{a} \big(\partial \pi(a | s_i)\big) A^\pi_i(s_i, a) \ .
\end{equation*}
Then
\begin{equation*}
\bm{\Pi}^\pi_i \bm{r}_i = \lambda \sum_{s_i} \Big(\sum_{a} \big(\partial \pi(a | s_i)\big)  A^\pi_i(s_i, a) \Big)\bm{e}_{s_i} = \lambda \big((\partial \bm{\Pi}^\pi_i) \bm{A}^{\pi}_i\big)
\end{equation*}

Apply the above results to Equation \eqref{sv-recurs-nvmdp-mat2scar}:
\begin{equation*}
\begin{aligned}
\Big(\big(\prod_{j = t + 1}^i \bm{L}_j^\pi\big) \big((\partial \bm{\Pi}^\pi_i) \bm{A}^{\pi}_i\big) \Big)_s &= \E^{a_{i + 1}, ..., s_i \sim \pi} \Big[\Gamma^\tau_{t, i} \sum_{a} \big(\partial \pi(a | s_i)\big) A^\pi_i(s_i, a) \mid s_t = s\Big] \quad(\lambda \text{ is cancelled})\\
\Rightarrow \partial V^{\pi}_t(s) &= \sum_{i = t}^\infty \E^{a_{i + 1}, ..., s_i \sim \pi} \Big[\Gamma^\tau_{t, i} \sum_a \pi(a | s_i) A^\pi_i(s_i, a) {\partial \pi(a | s_i) \over \pi(a | s_i)} \mid s_t = s\Big] \quad(\text{by Eq. } \eqref{svderiv-recurs-nvmdp-mat}) \\
&= \sum_{i = t}^\infty \E^{a_{i + 1}, ..., s_i \sim \pi} \Big[\Gamma^\tau_{t, i} \E^{a_i \sim \pi}[ A^\pi_i(s_i, a_i) {\partial \pi(a_i | s_i) \over \pi(a_i | s_i)} \mid s_i] \mid s_t = s\Big] \\
&= \sum_{i = t}^\infty \E^\pi[\Gamma^\tau_{t, i} A^\pi_i(s_i, a_i) {\partial \pi(a_i | s_i) \over \pi(a_i | s_i)} \mid s_t = s] = \text{R.H.S of Equation } \eqref{pol-grad-nvmdp} \ .\ \blacksquare
\end{aligned}
\end{equation*}

\begin{manuallemma}{\ref{thm:prfm-diff-lemma-nvmdp}}
For a NVMDP and any two policies $\pi, \pi'$:
\begin{equation*}
V^{\pi'}_t(s) - V^\pi_t(s) = \E^{\pi'} [\sum_{i = t}^\infty \Gamma^\tau_{t, i} A^\pi_i(s_i, a_i) \mid s_t = s] \ .
\end{equation*}
\end{manuallemma}

\noindent\textbf{Proof 1:}
By Lemma \ref{thm:sv-av-avgrwd-nvmdp},
\begin{equation*}
V^{\pi'}_t(s) = \E^{\pi'} [\sum_{i = t}^{\infty} \Gamma^\tau_{t, i} \bar{r}_i(s_i, a_i) \mid s_t = s] \ .
\end{equation*}

Moreover,
\begin{equation*}
\begin{aligned}
V^\pi_t(s) &= \E^{\pi'} \Bigl[\sum_{i = t}^{\infty} \Bigl( \Gamma^\tau_{t, i} V^\pi_i(s_i) - \Gamma^\tau_{t, i + 1} V^\pi_{i + 1}(s_{i + 1}) \Bigr) \mid s_t = s\Bigr] \quad (\Gamma^\tau_{t, t} = 1) \\
&= \E^{\pi'} \Bigl[\sum_{i = t}^{\infty} \Gamma^\tau_{t, i} \Bigl(V^\pi_i(s_i) - \gamma_{i + 1}(s_i, a_i, s_{i + 1}) V^\pi_{i + 1}(s_{i + 1}) \Bigr) \mid s_t = s\Bigr] \ .
\end{aligned}
\end{equation*}

Thus
\begin{equation*}
\begin{aligned}
&V^{\pi'}_t(s) - V^\pi_t(s) = \E^{\pi'} \Bigl[\sum_{i = t}^{\infty} \Gamma^\tau_{t, i} \Bigl(\bar{r}_i(s_i, a_i) + \gamma_{i + 1}(s_i, a_i, s_{i + 1}) V^\pi_{i + 1}(s_{i + 1}) - V^\pi_i(s_i) \Bigr) \mid s_t = s\Bigr] \\
&= \sum_{i = t}^{\infty} \E^{\pi'} \Bigl[ \Gamma^\tau_{t, i} \Bigl(\bar{r}_i(s_i, a_i) + \gamma_{i + 1}(s_i, a_i, s_{i + 1}) V^\pi_{i + 1}(s_{i + 1}) - V^\pi_i(s_i) \Bigr) \mid s_t = s\Bigr] \\
&= \sum_{i = t}^{\infty} \E^{a_t, s_{t + 1}, ... , s_i, a_i \sim \pi'} \Bigl[ \E^{s_{i + 1}} [ \Gamma^\tau_{t, i} \Bigl(\bar{r}_i(s_i, a_i) + \gamma_{i + 1}(s_i, a_i, s_{i + 1}) V^\pi_{i + 1}(s_{i + 1}) - V^\pi_i(s_i) \Bigr) \mid a_i, s_i, ..., s_t] \mid s_t = s\Bigr] \\
&= \sum_{i = t}^{\infty} \E^{a_t, s_{t + 1}, ... , s_i, a_i \sim \pi'} \Bigl[\Gamma^\tau_{t, i} \Bigl(Q^\pi_i(s_i, a_i) - V^\pi_i(s_i) \Bigr) \mid s_t = s\Bigr] \quad (\text{by Eq. \eqref{av-iter-nvmdp-e}}) \\
&= \sum_{i = t}^{\infty} \E^{\pi'} [\Gamma^\tau_{t, i} A^\pi_i(s_i, a_i) \mid s_t = s] = \E^{\pi'} [\sum_{i = t}^{\infty} \Gamma^\tau_{t, i} A^\pi_i(s_i, a_i) \mid s_t = s] \ .\ \blacksquare
\end{aligned}
\end{equation*}

\noindent\textbf{Proof 2:}
Consider any two policy $\pi, \pi'$.
By Equation \eqref{sv-av-nvmdp-mat}, \eqref{adv-nvmdp-mat}, and \eqref{piu-relp-nvmdp-mat},
\begin{equation*}
\begin{aligned}
\bm{\Pi}^\pi_t \bm{A}_t^\pi = \bm{\Pi}^\pi_t \bm{Q}_t^\pi - \bm{\Pi}^\pi_t \bm{U} \bm{V}_t^\pi = \bm{\Pi}^\pi_t \bm{Q}_t^\pi - \bm{V}_t^\pi = \bm{0} \quad(\forall \pi) \ ,
\end{aligned}
\end{equation*}
and
\begin{equation*}
\begin{aligned}
\bm{U} \bm{V}_t^\pi &= \bm{Q}_t^\pi - \bm{A}_t^\pi \\
\bm{U} \bm{V}_t^{\pi'} &= \bm{Q}_t^{\pi'} - \bm{A}_t^{\pi'} \\
\Rightarrow \bm{U} (\bm{V}_t^{\pi'} - \bm{V}_t^\pi) &= \bm{A}_t^\pi - \bm{A}_t^{\pi'} + (\bm{Q}_t^{\pi'} - \bm{Q}_t^\pi) \\
\Rightarrow \bm{V}_t^{\pi'} - \bm{V}_t^\pi &= \bm{\Pi}^{\pi'}_t \bm{A}_t^\pi + \bm{\Pi}^{\pi'}_t (\bm{Q}_t^{\pi'} - \bm{Q}_t^\pi) \quad(\bm{\Pi}^{\pi'}_t \bm{U} = \bm{I}, \bm{\Pi}^{\pi'}_t \bm{A}_t^{\pi'} = \bm{0}) \ .
\end{aligned}
\end{equation*}
By Equation \eqref{av-iter-nvmdp-mat},
\begin{equation*}
\bm{Q}^{\pi'}_t - \bm{Q}^\pi_t = \bm{J}_{t + 1} (\bm{V}^{\pi'}_{t + 1} - \bm{V}^\pi_{t + 1}) \ ,
\end{equation*}
thus
\begin{equation}\phantomsection\label{svdiff-selfiter-nvmdp-mat}
\begin{aligned}[b]
\bm{V}_t^{\pi'} - \bm{V}_t^\pi &= \bm{\Pi}^{\pi'}_t \bm{A}_t^\pi + \bm{\Pi}^{\pi'}_t \bm{J}_{t + 1} (\bm{V}^{\pi'}_{t + 1} - \bm{V}^\pi_{t + 1}) \\
&= \bm{\Pi}^{\pi'}_t \bm{A}_t^\pi + \bm{L}_{t + 1}^{\pi'} (\bm{V}^{\pi'}_{t + 1} - \bm{V}^\pi_{t + 1}) \quad(\text{by Eq. } \eqref{l-def-nvmdp-mat})
\end{aligned}
\end{equation}

By Equation \eqref{svdiff-selfiter-nvmdp-mat},
\begin{equation}\phantomsection\label{svdiff-recurs-nvmdp-mat}
\begin{aligned}[b]
\bm{V}_t^{\pi'} - \bm{V}_t^\pi &= \bm{\Pi}^{\pi'}_t \bm{A}_t^\pi + \bm{L}_{t + 1}^{\pi'} \bm{\Pi}^{\pi'}_{t + 1} \bm{A}_{t + 1}^\pi + \bm{L}_{t + 1}^{\pi'} \bm{L}_{t + 2}^{\pi'} (\bm{V}^{\pi'}_{t + 2} - \bm{V}^\pi_{t + 2}) \\
&... \\
&=\sum_{i = t}^\infty \big(\prod_{j = t + 1}^i \bm{L}_j^{\pi'} \big) \bm{\Pi}^{\pi'}_i \bm{A}_i^\pi \ .
\end{aligned}
\end{equation}
Moreover, by Lemma \ref{thm:sv-av-avgrwd-nvmdp} and Equation \eqref{sv-recurs-nvmdp-mat},
\begin{equation}\phantomsection\label{sv-mat2scar-nvmdp}
(\bm{V}^{\pi'}_t)_s = \Big(\sum_{i = t}^\infty \big(\prod_{j = t + 1}^i \bm{L}_j^{\pi'} \big) \bm{\Pi}^{\pi'}_i \bm{r}_i\Big)_s = \E^{\pi'} [\sum_{i = t}^\infty \Gamma^\tau_{t, i} (\bm{r}_i)_{(s_i, a_i)} \mid s_t = s] \ .
\end{equation}
Equation \eqref{sv-mat2scar-nvmdp} holds for all $\bm{r}_i$ s.t. $\|\bm{r}_i\|_\infty \le R_B$.  
For sufficiently small $\lambda > 0$, substituting $\bm{r}_i$ with $\lambda \bm{A}_i^\pi$ in Equation \eqref{sv-mat2scar-nvmdp} and combining with Equation \eqref{svdiff-recurs-nvmdp-mat} proves the lemma.
$\blacksquare$

\begin{manuallemma}{\ref{thm:D-oper-trans}}
\begin{equation*}
D^{\pi, \pi'}_t(s) = \E^\pi [\sum_{i = t}^\infty \Gamma^\tau_{t, i} A^\pi_i(s_i, a_i) {\pi'(a_i | s_i) \over \pi(a_i | s_i)} \mid s_t = s] \ .
\end{equation*}
\end{manuallemma}

\noindent\textbf{Proof:}
By Equation \eqref{D-oper-def},
\begin{equation*}
\begin{aligned}
&D^{\pi, \pi'}_t(s) = \sum_{i = t}^\infty \E^{a_t, s_{t + 1}, ..., s_i \sim \pi} \bigl[ \E^{a_i \sim \pi'} [\Gamma^\tau_{t, i} A^\pi_i(s_i, a_i) \mid s_i, ..., a_t, s_t ] \mid s_t = s \bigr] \\
&= \sum_{i = t}^\infty \E^{a_t, s_{t + 1}, ..., s_i \sim \pi} [\Gamma^\tau_{t, i} \sum_{a_i} \pi(a_i | s_i) A^\pi_i(s_i, a_i) {\pi'(a_i | s_i) \over \pi(a_i | s_i)} \mid s_t = s] \\
&= \sum_{i = t}^\infty \E^{a_t, s_{t + 1}, ..., s_i \sim \pi} \bigl[\Gamma^\tau_{t, i} \E^{a_i \sim \pi} [A^\pi_i(s_i, a_i) {\pi'(a_i | s_i) \over \pi(a_i | s_i)} \mid s_i \bigr] \mid s_t = s] \\
&= \sum_{i = t}^\infty \E^{a_t, s_{t + 1}, ..., s_i \sim \pi, a_i \sim \pi} [\Gamma^\tau_{t, i} A^\pi_i(s_i, a_i) {\pi'(a_i | s_i) \over \pi(a_i | s_i)} \mid s_t = s] 
= \E^\pi [\sum_{i = t}^\infty \Gamma^\tau_{t, i} A^\pi_i(s_i, a_i) {\pi'(a_i | s_i) \over \pi(a_i | s_i)} \mid s_t = s] \ .\ \blacksquare
\end{aligned}
\end{equation*}

The following theorem builds on the work of Schulman et al. \cite{TRPO-2015}.
We extend their result to the NVMDP setting in our proof 1, clarifying several technical steps that are implicit or only briefly sketched in their original presentation.

\begin{manualtheorem}{\ref{thm:prfm-diff-absbnd}}
If a NVMDP satisfies Assumption \ref{asm:trpo-Gamma-bd}, then for any two policies $\pi, \pi'$ and state $s$:
\begin{equation}\phantomsection\label{prfm-diff-absbnd}
\bigl| V^{\pi'}_t(s) - V^\pi_t(s) - D^{\pi, \pi'}_t(s) \bigr| \le C \alpha_t^2(\pi, \pi') \ ,
\end{equation}
where $D^{\pi, \pi'}_t(s)$ is defined by Equation \eqref{D-oper-def},
C is some constant, and
\begin{equation}\phantomsection\label{tvdist-upb}
\alpha_t(\pi, \pi') = \max_{j \ge t, x \in S} D_{TV} \bigl( \pi_j(\cdot | x)\|\pi'_j(\cdot | x) \bigr) \ .
\end{equation}
($D_{TV}$ is the total variation distance.)
\end{manualtheorem}

\noindent\textbf{Proof 1:}
Consider $\E^{a_t, s_{t + 1}, ..., s_i \sim \pi', a_i \sim \pi'} [\Gamma^\tau_{t, i} A^\pi_i(s_i, a_i) \mid s_t = s]$,
where $t, i$ denote the start time and the current time, respectively.
For the trajectory $a_t, s_{t + 1}, ..., s_{i - 1}, a_{i - 1}$ following policy $\pi'$,
an indicator variable $X'$ is 0 if the same actions $a_t, a_{t + 1}, ... , a_{i - 1}$ are selected under state $s_t, s_{t + 1}, ... , s_{i - 1}$ by policy $\pi$, respectively.
$X' = 1$ otherwise.
Thus,
\begin{equation}\phantomsection\label{D-oper-item-pi'}
\begin{aligned}[b]
&\E^{a_t, s_{t + 1}, ..., s_i \sim \pi', a_i \sim \pi'} [\Gamma^\tau_{t, i} A^\pi_i(s_i, a_i) \mid s_t = s] \\
&= \E^{X'} \Bigl[ \E^{a_t, s_{t + 1}, ..., s_i \sim \pi', a_i \sim \pi'} [\Gamma^\tau_{t, i} A^\pi_i(s_i, a_i) \mid X', s_t = s] \mid s_t = s \Bigr] \\
&= P(X' = 0 | s_t = s) \E^{a_t, s_{t + 1}, ..., s_i \sim \pi', a_i \sim \pi'}  [\Gamma^\tau_{t, i} A^\pi_i(s_i, a_i) \mid X' = 0, s_t = s] \\
&+ P(X' = 1 | s_t = s) \E^{a_t, s_{t + 1}, ..., s_i \sim \pi', a_i \sim \pi'} [\Gamma^\tau_{t, i} A^\pi_i(s_i, a_i) \mid X' = 1, s_t = s] \ .
\end{aligned}
\end{equation}

Simliarly, for the trajectory $a_t, s_{t + 1}, ..., s_{i - 1}, a_{i - 1}$ following policy $\pi$,
an indicator variable $X$ is 0 if the same actions $a_t, a_{t + 1}, ... , a_{i - 1}$ are selected under state $s_t, s_{t + 1}, ... , s_{i - 1}$ by policy $\pi'$, respectively.
$X = 1$ otherwise.
Therefore,
\begin{equation}\phantomsection\label{D-oper-item-pi}
\begin{aligned}[b]
&\E^{a_t, s_{t + 1}, ..., s_i \sim \pi, a_i \sim \pi'} [\Gamma^\tau_{t, i} A^\pi_i(s_i, a_i) \mid s_t = s] \\
&= \E^{X} \Bigl[ \E^{a_t, s_{t + 1}, ..., s_i \sim \pi, a_i \sim \pi'} [\Gamma^\tau_{t, i} A^\pi_i(s_i, a_i) \mid X, s_t = s] \mid s_t = s \Bigr] \\
&= P(X = 0 | s_t = s) \E^{a_t, s_{t + 1}, ..., s_i \sim \pi, a_i \sim \pi'} [\Gamma^\tau_{t, i} A^\pi_i(s_i, a_i) \mid X = 0, s_t = s] \\
&+ P(X = 1 | s_t = s) \E^{a_t, s_{t + 1}, ..., s_i \sim \pi, a_i \sim \pi'} [\Gamma^\tau_{t, i} A^\pi_i(s_i, a_i) \mid X = 1, s_t = s] \ .
\end{aligned}
\end{equation}

For $P(X' = 0 | s_t = s)$ and $P(X = 0 | s_t = s)$,
\begin{equation}\phantomsection\label{PX'0-eq-PX0}
\begin{aligned}[b]
&P(X' = 0 | s_t = s) \\
&= \sum_{a_t, s_{t + 1}, ..., s_{i - 1}, a_{i - 1}} P(X' = 0, a^{\pi'}_t = a_t, s^\tau_{t + 1} = s_{t + 1}, ..., s^\tau_{i - 1} = s_{i - 1}, a^{\pi'}_{i - 1} = a_{i - 1} | s_t = s) \\
&= \sum_{a_t, s_{t + 1}, ..., s_{i - 1}, a_{i - 1}} P(a^{\pi'}_t = a^{\pi}_t = a_t, s^\tau_{t + 1} = s_{t + 1}, ..., s^\tau_{i - 1} = s_{i - 1}, a^{\pi'}_{i - 1} = a^{\pi}_{i - 1} = a_{i - 1} | s_t = s) \\
&= \sum_{a_t, s_{t + 1}, ..., s_{i - 1}, a_{i - 1}} P(X = 0, a^{\pi}_t = a_t, s^\tau_{t + 1} = s_{t + 1}, ..., s^\tau_{i - 1} = s_{i - 1}, a^{\pi}_{i - 1} = a_{i - 1} | s_t = s) \\
&= P(X = 0 | s_t = s) \ .
\end{aligned}
\end{equation}

Since
\begin{equation*}
\begin{aligned}
&\E^{a_t, s_{t + 1}, ..., s_i \sim \pi', a_i \sim \pi'}  [\Gamma^\tau_{t, i} A^\pi_i(s_i, a_i) \mid X' = 0, s_t = s] \\
&=\sum_{a_t, s_{t + 1}, ..., s_i, a_i} {\Gamma^\tau_{t, i} A^\pi_i(s_i, a_i) \pi'(a_i | s_i) P(a^{\pi'}_t = a_t, s^\tau_{t + 1} = s_{t + 1}, ..., a^{\pi'}_{i - 1} = a_{i - 1} , s^\tau_i = s_i | X' = 0, s_t = s)} \\
&=\sum_{a_t, s_{t + 1}, ..., s_i, a_i} {{\Gamma^\tau_{t, i} A^\pi_i(s_i, a_i) \pi'(a_i | s_i) P(a^{\pi'}_t = a_t, s^\tau_{t + 1} = s_{t + 1}, ..., a^{\pi'}_{i - 1} = a_{i - 1} , s^\tau_i = s_i, X' = 0 | s_t = s)} \over P(X' = 0 | s_t = s)} \\
&=\sum_{a_t, s_{t + 1}, ..., s_i, a_i} {{\Gamma^\tau_{t, i} A^\pi_i(s_i, a_i) \pi'(a_i | s_i) P(a^{\pi'}_t = a^{\pi}_t = a_t, s^\tau_{t + 1} = s_{t + 1}, ..., a^{\pi'}_{i - 1} = a^{\pi}_{i - 1} = a_{i - 1}, s^\tau_i = s_i | s_t = s)} \over P(X' = 0 | s_t = s)} \ ,
\end{aligned}
\end{equation*}
and similarly
\begin{equation*}
\begin{aligned}
&\E^{a_t, s_{t + 1}, ..., s_i \sim \pi, a_i \sim \pi'}  [\Gamma^\tau_{t, i} A^\pi_i(s_i, a_i) \mid X = 0, s_t = s] \\
&=\sum_{a_t, s_{t + 1}, ..., s_i, a_i} {{\Gamma^\tau_{t, i} A^\pi_i(s_i, a_i) \pi'(a_i | s_i) P(a^{\pi'}_t = a^{\pi}_t = a_t, s^\tau_{t + 1} = s_{t + 1}, ..., a^{\pi'}_{i - 1} = a^{\pi}_{i - 1} = a_{i - 1}, s^\tau_i = s_i | s_t = s)} \over P(X = 0 | s_t = s)} \ ,
\end{aligned}
\end{equation*}
we have:
\begin{equation}\phantomsection\label{Xeq0-items-equal}
\begin{aligned}
&\E^{a_t, s_{t + 1}, ..., s_i \sim \pi', a_i \sim \pi'}  [\Gamma^\tau_{t, i} A^\pi_i(s_i, a_i) \mid X' = 0, s_t = s] \\
&= \E^{a_t, s_{t + 1}, ..., s_i \sim \pi, a_i \sim \pi'}  [\Gamma^\tau_{t, i} A^\pi_i(s_i, a_i) \mid X = 0, s_t = s] \ .
\end{aligned}
\end{equation}

Combine the results of Equation \eqref{D-oper-item-pi'}, \eqref{D-oper-item-pi}, \eqref{PX'0-eq-PX0}, and \eqref{Xeq0-items-equal}:
\begin{equation}\phantomsection\label{prfm_diff_ti}
\begin{aligned}[b]
&\E^{a_t, s_{t + 1}, ..., s_i \sim \pi', a_i \sim \pi'} [\Gamma^\tau_{t, i} A^\pi_i(s_i, a_i) \mid s_t = s] - \E^{a_t, s_{t + 1}, ..., s_i \sim \pi, a_i \sim \pi'} [\Gamma^\tau_{t, i} A^\pi_i(s_i, a_i) \mid s_t = s] \\
&= P(X' = 1 | s_t = s) \Bigl(\E^{a_t, s_{t + 1}, ..., s_i \sim \pi', a_i \sim \pi'} [\Gamma^\tau_{t, i} A^\pi_i(s_i, a_i) \mid X' = 1, s_t = s] \\
&- \E^{a_t, s_{t + 1}, ..., s_i \sim \pi, a_i \sim \pi'} [\Gamma^\tau_{t, i} A^\pi_i(s_i, a_i) \mid X = 1, s_t = s]\Bigr) \ .
\end{aligned}
\end{equation}

By Equation \eqref{PX'0-eq-PX0}:
\begin{equation*}
\begin{aligned}
&P(X' = 0 | s_t = s) \\
&= \sum_{a_t, s_{t + 1}, ..., s_{i - 1}, a_{i - 1}} P(a^{\pi'}_t = a^{\pi}_t = a_t, s^\tau_{t + 1} = s_{t + 1}, ..., s^\tau_{i - 1} = s_{i - 1}, a^{\pi'}_{i - 1} = a^{\pi}_{i - 1} = a_{i - 1} | s_t = s) \\
&= \sum_{a_t, s_{t + 1}, ..., s_{i - 1}, a_{i - 1}} P(a^{\pi'}_{i - 1} = a^{\pi}_{i - 1} = a_{i - 1} | s_t = s, ..., s^\tau_{i - 1} = s_{i - 1}) P(s^\tau_{i - 1} = s_{i - 1} | s_t = s, ..., a^{\pi'}_{i - 2} = a^{\pi}_{i - 1} = a_{i - 2}) \\
& ... P(s^\tau_{t + 1} = s_{t + 1} | s_t = s, a^{\pi'}_t = a^{\pi}_t = a_t) P(a^{\pi'}_t = a^{\pi}_t = a_t | s_t = s) \\
&= \sum_{a_t, s_{t + 1}, ..., s_{i - 1}, a_{i - 1}} P(a^{\pi'}_{i - 1} = a^{\pi}_{i - 1} = a_{i - 1} | s_{i - 1}) p_{i - 2}(s_{i - 1} | s_{i - 2}, a_{i - 2}) ... p_t(s_{t + 1} | s_t, a_t) P(a^{\pi'}_t = a^{\pi}_t = a_t | s_t = s) \\
&= \sum_{a_t, s_{t + 1}, ..., s_{i - 1}} P(a^{\pi'}_{i - 1} = a^{\pi}_{i - 1} | s_{i - 1}) p_{i - 2}(s_{i - 1} | s_{i - 2}, a_{i - 2}) ... p_t(s_{t + 1} | s_t, a_t) P(a^{\pi'}_t = a^{\pi}_t = a_t | s_t = s)
\end{aligned}
\end{equation*}

Let $\alpha$ be an upper bound such that
\begin{equation*}
P(a^{\pi'}_j \neq a^{\pi}_j | s_j = x) \le \alpha \quad (\forall j \ge t, x \in S) \ .
\end{equation*}

Then
\begin{equation*}
\begin{aligned}
&P(X' = 0 | s_t = s) \\
&\ge (1 - \alpha) \sum_{a_t, s_{t + 1}, ..., s_{i - 1}} p_{i - 2}(s_{i - 1} | s_{i - 2}, a_{i - 2}) P(a^{\pi'}_{i - 2} = a^{\pi}_{i - 2} = a_{i - 2} | s_{i - 2}) ... p_t(s_{t + 1} | s_t, a_t) P(a^{\pi'}_t = a^{\pi}_t = a_t | s_t = s) \\
&= (1 - \alpha) \sum_{a_t, s_{t + 1}, ..., a_{i - 2}} P(a^{\pi'}_{i - 2} = a^{\pi}_{i - 2} = a_{i - 2} | s_{i - 2}) ... p_t(s_{t + 1} | s_t, a_t) P(a^{\pi'}_t = a^{\pi}_t = a_t | s_t = s) \\
&= (1 - \alpha) \sum_{a_t, s_{t + 1}, ..., s_{i - 2}} P(a^{\pi'}_{i - 2} = a^{\pi}_{i - 2} | s_{i - 2}) ... p_t(s_{t + 1} | s_t, a_t) P(a^{\pi'}_t = a^{\pi}_t = a_t | s_t = s) \\
&... \\
&\ge (1 - \alpha)^{i - t - 1} \sum_{a_t, s_{t + 1}} p_t(s_{t + 1} | s_t, a_t) P(a^{\pi'}_t = a^{\pi}_t = a_t | s_t) \\
&= (1 - \alpha)^{i - t - 1} \sum_{a_t} P(a^{\pi'}_t = a^{\pi}_t = a_t | s_t) \\
&= (1 - \alpha)^{i - t - 1} P(a^{\pi'}_t = a^{\pi}_t | s_t) \\
&\ge (1 - \alpha)^{i - t} \ .
\end{aligned}
\end{equation*}

Thus
\begin{equation*}
\begin{aligned}
P(X' = 1 | s_t = s) &= 1 - P(X' = 0 | s_t = s) \\
&\le 1 - (1 - \alpha)^{i - t} \\
&= \alpha (1 + (1 - \alpha) + ... + (1 - \alpha)^{i - t - 1}) \\
&\le (i - t)\alpha \ .
\end{aligned}
\end{equation*}

Since $\E^{a \sim \pi}[A^\pi_i(s_i, a) \mid s_i] = \E^{a \sim \pi}[Q^\pi_i(s_i, a) - V^\pi_i(s_i) \mid s_i] = 0$,
\begin{equation*}
\begin{aligned}
&\E^{a_t, s_{t + 1}, ..., s_i \sim \pi', a_i \sim \pi'} [\Gamma^\tau_{t, i} A^\pi_i(s_i, a_i) \mid X' = 1, s_t = s] \\
&= \E^{a_t, s_{t + 1}, ..., s_i \sim \pi', a_i \sim \pi'} [\Gamma^\tau_{t, i} A^\pi_i(s_i, a_i) \mid X' = 1, s_t = s] - \E^{a_t, s_{t + 1}, ..., s_i \sim \pi', \bar{a}_i \sim \pi} [\Gamma^\tau_{t, i} A^\pi_i(s_i, \bar{a}_i) \mid X' = 1, s_t = s] \\
&= \E^{a_t, s_{t + 1}, ..., s_i \sim \pi', a_i \sim \pi', \bar{a}_i \sim \pi} [\Gamma^\tau_{t, i} \bigl(A^\pi_i(s_i, a_i) - A^\pi_i(s_i, \bar{a}_i) \bigr) \mid X' = 1, s_t = s] \\
&= \E^{a_t, s_{t + 1}, ..., s_i \sim \pi'} \bigl[\E^{a_i \sim \pi', \bar{a}_i \sim \pi} [\Gamma^\tau_{t, i} \bigl(A^\pi_i(s_i, a_i) - A^\pi_i(s_i, \bar{a}_i) \bigr) \mid s_t, a_t, ..., s_i, X' = 1] \mid X' = 1, s_t = s \bigr] \\
&= \E^{a_t, s_{t + 1}, ..., s_i \sim \pi'} \bigl[\Gamma^\tau_{t, i} \E^{a_i \sim \pi', \bar{a}_i \sim \pi} [A^\pi_i(s_i, a_i) - A^\pi_i(s_i, \bar{a}_i) \mid s_i] \mid X' = 1, s_t = s \bigr] \\
&\Big| \E^{a_i \sim \pi', \bar{a}_i \sim \pi} [A^\pi_i(s_i, a_i) - A^\pi_i(s_i, \bar{a}_i) \mid s_i] \Big| \\
&= \Big| P(a_i \neq \bar{a}_i | s_i) \E^{a_i \sim \pi', \bar{a}_i \sim \pi} [A^\pi_i(s_i, a_i) - A^\pi_i(s_i, \bar{a}_i) \mid a_i \neq \bar{a}_i, s_i] \\
&+ P(a_i = \bar{a}_i | s_i) \E^{a_i \sim \pi', \bar{a}_i \sim \pi} [A^\pi_i(s_i, a_i) - A^\pi_i(s_i, \bar{a}_i) \mid a_i = \bar{a}_i, s_i] \Big| \\
&= \Big| P(a_i \neq \bar{a}_i | s_i) \E^{a_i \sim \pi', \bar{a}_i \sim \pi} [A^\pi_i(s_i, a_i) - A^\pi_i(s_i, \bar{a}_i) \mid a_i \neq \bar{a}_i, s_i] \Big| \le 2 \alpha A_B \\
&\Rightarrow \Big|\E^{a_t, s_{t + 1}, ..., s_i \sim \pi', a_i \sim \pi'} [\Gamma^\tau_{t, i} A^\pi_i(s_i, a_i) \mid X' = 1, s_t = s]\Big| \le 2 \alpha A_B \Gamma^M_{t, i} \ ,
\end{aligned}
\end{equation*}
where $A_B = sup_{j, s, a, \pi}\big|A^\pi_j(s, a) \big|$ and $\Gamma^M_{t, i} = \prod_{j = t + 1}^i \max_{s, a, s'} \gamma_j(s, a, s')$.

Similarly,
\begin{equation*}
\begin{aligned}
\Big|\E^{a_t, s_{t + 1}, ..., s_i \sim \pi, a_i \sim \pi'} [\Gamma^\tau_{t, i} A^\pi_i(s_i, a_i) \mid X = 1, s_t = s]\Big| \le 2 \alpha A_B \Gamma^M_{t, i} \ .
\end{aligned}
\end{equation*}

Applying the above bounds to Equation \eqref{prfm_diff_ti}:
\begin{equation*}
\begin{aligned}[b]
&\Big| \E^{a_t, s_{t + 1}, ..., s_i \sim \pi', a_i \sim \pi'} [\Gamma^\tau_{t, i} A^\pi_i(s_i, a_i) \mid s_t = s] - \E^{a_t, s_{t + 1}, ..., s_i \sim \pi, a_i \sim \pi'} [\Gamma^\tau_{t, i} A^\pi_i(s_i, a_i) \mid s_t = s] \Big| \\
&\le P(X' = 1 | s_t = s) \cdot 4 \alpha A_B \Gamma^M_{t, i} \le 4(i - t) \alpha^2 A_B \Gamma^M_{t, i} \ .
\end{aligned}
\end{equation*}

Combine Equation \eqref{prfm-diff-exp}, Equation \eqref{D-oper-def}, and the above bound,
\begin{equation}
\begin{aligned}[b]
&\bigl| V^{\pi'}_t(s) - V^\pi_t(s) - D^{\pi, \pi'}_t(s) \bigr| \\
&= \Bigl| \sum_{i = t}^\infty \Bigl( \E^{a_t, s_{t + 1}, ..., s_i \sim \pi', a_i \sim \pi'} [\Gamma^\tau_{t, i} A^\pi_i(s_i, a_i) \mid s_t = s] - \E^{a_t, s_{t + 1}, ..., s_i \sim \pi, a_i \sim \pi'} [\Gamma^\tau_{t, i} A^\pi_i(s_i, a_i) \mid s_t = s] \Bigr) \Bigr| \\
&\le \sum_{i = t}^\infty \bigl| \E^{a_t, s_{t + 1}, ..., s_i \sim \pi', a_i \sim \pi'} [\Gamma^\tau_{t, i} A^\pi_i(s_i, a_i) \mid s_t = s] - \E^{a_t, s_{t + 1}, ..., s_i \sim \pi, a_i \sim \pi'} [\Gamma^\tau_{t, i} A^\pi_i(s_i, a_i) \mid s_t = s] \bigr| \\
&\le 4A_B \alpha^2 \sum_{i = t}^\infty (i - t) \Gamma^M_{t, i} \ ,
\end{aligned}
\end{equation}
where $\sum_{i = t}^\infty (i - t) \Gamma^M_{t, i}$ is bounded by $M$ by Assumption \ref{asm:trpo-Gamma-bd}.

The total variation distance has the following property (Levin et al. \cite{MC-MixTime-2017}):
if the total variation distance between two distributions $Y$ and $Z$ is $c$,
then a joint distribution $(Y', Z')$ exists such that $Y', Z'$ have the same marginal distributions as $Y, Z$, respectively, and $P(Y' \neq Z') = c$.
For any two policies $\pi, \pi'$ and $\alpha_t(\pi, \pi')$ from Equation \eqref{tvdist-upb}, we can always choose policies $\bar{\pi}, \bar{\pi}'$ with the same marginals as $\pi, \pi'$ such that:
\begin{equation*}
P(a^{\bar{\pi}'}_j \neq a^{\bar{\pi}}_j | s_j = x) = D_{TV} \bigl( \pi'_j(\cdot | x)\|\pi_j(\cdot | x) \bigr) \le \alpha_t(\pi, \pi') \quad (\forall j \ge t, x \in S) \ ,
\end{equation*}
then
\begin{equation*}
\begin{aligned}
\bigl| V^{\pi'}_t(s) - V^\pi_t(s) - D^{\pi, \pi'}_t(s) \bigr| = \bigl| V^{\bar{\pi}'}_t(s) - V^{\bar{\pi}}_t(s) - D^{\bar{\pi}, \bar{\pi}'}_t(s) \bigr| \le 4A_B \alpha_t^2(\pi, \pi') \sum_{i = t}^\infty (i - t) \Gamma^M_{t, i} \ .
\end{aligned}
\end{equation*}
The theorem is thus proven since $4 A_B \sum_{i = t}^\infty (i - t) \Gamma^M_{t, i}$ is bounded.
$\blacksquare$

\noindent\textbf{Proof 2:}
Similarly to Proof 2 of Theorem \ref{thm:pol-grad-nvmdp}, let $r_i(s_i, a_i, s_{i + 1}) = \bar{r}_i(s_i, a_i) = r_i(s_i, a_i)$.
Then by Lemma \ref{thm:sv-av-avgrwd-nvmdp}, Equation \eqref{sv-recurs-nvmdp-mat}, and setting $\bm{r}_u = \bm{0}$ for all $u \ge t, u \ne i$:
\begin{equation*}
\Big(\big(\prod_{j = t + 1}^i \bm{L}_j^\pi\big) {\bm{\Pi}^\pi_i \bm{r}_i}\Big)_s = \E^\pi [\Gamma^\tau_{t, i} (\bm{r}_i)_{(s_i, a_i)} \mid s_t = s] \ ,
\end{equation*}
which holds for all $\bm{r}_i$ s.t. $\|\bm{r}_i\|_\infty \le R_B$.

Let $\lambda$ denotes a sufficient small positive real number, and set
\begin{equation*}
(\bm{r}_i)_{(s_i, a_i)} = \lambda \sum_{a} \pi'(a | s_i) A^\pi_i(s_i, a) \ .
\end{equation*}
Then
\begin{equation*}
\bm{\Pi}^\pi_i \bm{r}_i = \lambda \sum_{s_i} \Big(\sum_{a} \pi'(a | s_i) A^\pi_i(s_i, a)\Big)\bm{e}_{s_i} = \lambda \bm{\Pi}^{\pi'}_i \bm{A}^{\pi}_i
\end{equation*}

Combine the above results:
\begin{equation*}
\begin{aligned}
\Big(\big(\prod_{j = t + 1}^i \bm{L}_j^\pi\big) \bm{\Pi}^{\pi'}_i \bm{A}^{\pi}_i \Big)_s &= \E^{a_t, s_{t + 1}, ..., s_i \sim \pi} \Big[\Gamma^\tau_{t, i} \sum_{a} \pi'(a | s_i) A^\pi_i(s_i, a) \mid s_t = s\Big] \quad(\lambda \text{ is cancelled})\\
\Rightarrow \Big(\sum_{i = t}^\infty \big(\prod_{j = t + 1}^i \bm{L}_j^\pi\big) \bm{\Pi}^{\pi'}_i \bm{A}^{\pi}_i \Big)_s &= \sum_{i = t}^\infty \E^{a_t, s_{t + 1}, ..., s_i \sim \pi} \Big[\Gamma^\tau_{t, i} \sum_{a_i} \pi'(a_i | s_i) A^\pi_i(s_i, a_i) \mid s_t = s\Big] \\
&= \sum_{i = t}^\infty \E^{a_t, s_{t + 1}, ..., s_i \sim \pi} \Big[\Gamma^\tau_{t, i} \E^{a_i \sim \pi'}[A^\pi_i(s_i, a_i) \mid s_i, ..., s_t] \mid s_t = s\Big] \\
&= \sum_{i = t}^\infty \E^{a_t, s_{t + 1}, ..., s_i \sim \pi, a_i \sim \pi'} [\Gamma^\tau_{t, i} A^\pi_i(s_i, a_i) \mid s_t = s] \\
&= D^{\pi, \pi'}_t(s) \quad(\text{by Eq. } \eqref{D-oper-def})
\end{aligned}
\end{equation*}

Denote $\bm{D}^{\pi, \pi'}_t$ s.t. $(\bm{D}^{\pi, \pi'}_t)_s = D^{\pi, \pi'}_t(s)$, then
\begin{equation}\phantomsection\label{D-oper-nvmdp-mat}
\bm{D}^{\pi, \pi'}_t = \sum_{i = t}^\infty \big(\prod_{j = t + 1}^i \bm{L}_j^\pi\big) \bm{\Pi}^{\pi'}_i \bm{A}^{\pi}_i \ .
\end{equation}

Combine Equation \eqref{D-oper-nvmdp-mat} with Equation \eqref{svdiff-recurs-nvmdp-mat} (established in Proof 2 of Lemma \ref{thm:prfm-diff-lemma-nvmdp}):
\begin{equation}\phantomsection\label{prfm-diff-nvmdp-mat}
\bm{V}_t^{\pi'} - \bm{V}_t^\pi - \bm{D}^{\pi, \pi'}_t = \sum_{i = t + 1}^\infty \big(\prod_{j = t + 1}^i \bm{L}_j^{\pi'} - \prod_{j = t + 1}^i \bm{L}_j^\pi\big) \bm{\Pi}^{\pi'}_i \bm{A}_i^\pi \ .
\end{equation}
A notable fact about the R.H.S of Equation \eqref{prfm-diff-nvmdp-mat} is that the terms with $i = t$ cancel in the summation.

By the definition of $\alpha_t(\pi, \pi')$ in Equation \eqref{tvdist-upb},
\begin{equation*}\phantomsection\label{pi-diffupb-nvmdp-mat}
\begin{aligned}[b]
\|\bm{\Pi}^{\pi'}_i - \bm{\Pi}^{\pi}_i\|_\infty &= \max_{s} \sum_{a} |\pi'_i(a | s) - \pi_i(a | s)| \\
&= \max_{s} 2 D_{TV} \big(\pi_i(\cdot | s)\|\pi'_i(\cdot | s)\big) \\
&\le 2\alpha_t(\pi, \pi') \ .
\end{aligned}
\end{equation*}

Moreover, by Equation \eqref{sv-av-nvmdp-mat}, \eqref{adv-nvmdp-mat}, and \eqref{piu-relp-nvmdp-mat},
\begin{equation}\phantomsection\label{pi'a-inftynormupb-nvmdp}
\begin{aligned}[b]
\bm{\Pi}^\pi_i \bm{A}_i^\pi &= \bm{\Pi}^\pi_i \bm{Q}_i^\pi - \bm{\Pi}^\pi_i \bm{U} \bm{V}_i^\pi = \bm{\Pi}^\pi_i \bm{Q}_i^\pi - \bm{V}_i^\pi = \bm{0} \\
\Rightarrow \bm{\Pi}^{\pi'}_i \bm{A}_i^\pi &= \bm{\Pi}^{\pi'}_i \bm{A}_i^\pi - \bm{\Pi}^\pi_i \bm{A}_i^\pi = (\bm{\Pi}^{\pi'}_i - \bm{\Pi}^\pi_i) \bm{A}_i^\pi \\
\Rightarrow \|\bm{\Pi}^{\pi'}_i \bm{A}_i^\pi\|_\infty &\le \|\bm{\Pi}^{\pi'}_i - \bm{\Pi}^\pi_i\|_\infty \|\bm{A}_t^\pi\|_\infty \le 2A_B \alpha_t(\pi, \pi') \ ,
\end{aligned}
\end{equation}
where $A_B$ bounds all $A^\pi_i(\cdot)$ in absolute value.

Denote $\gamma^M_i = \max_{s, a, s'} \gamma_i(s, a, s')$ for all $i \ge t + 1$, then by Equation \eqref{l-def-nvmdp-mat},
\begin{equation}\phantomsection\label{l-diffupb-nvmdp-mat}
\begin{aligned}
\bm{L}_i^{\pi'} - \bm{L}_i^\pi &= (\bm{\Pi}^{\pi'}_{i - 1} - \bm{\Pi}^\pi_{i - 1}) \bm{J}_i \\
\Rightarrow \|\bm{L}_i^{\pi'} - \bm{L}_i^\pi\|_\infty &\le \|\bm{\Pi}^{\pi'}_{i - 1} - \bm{\Pi}^\pi_{i - 1}\|_\infty \|\bm{J}_i\|_\infty \\
&\le 2\alpha_t(\pi, \pi') \gamma^M_i \quad(\forall i \ge t + 1) \ .
\end{aligned}
\end{equation}

Thus the following inequality holds when $k = 1$:
\begin{equation}\phantomsection\label{l-proddiffupb-nvmdp-mat}
\big\|\prod_{j = t + 1}^{t + k} \bm{L}_j^{\pi'} - \prod_{j = t + 1}^{t + k} \bm{L}_j^\pi\big\|_\infty \le 2k \alpha_t(\pi, \pi') \prod_{j = t + 1}^{t + k} \gamma^M_j \ .
\end{equation}
Assume Inequality \eqref{l-proddiffupb-nvmdp-mat} holds for $k$.
Since Equation \eqref{l-def-nvmdp-mat} implies
\begin{equation}\phantomsection\label{l-inftynormupb-nvmdp}
\|\bm{L}_i^\pi\|_\infty \le \|\bm{\Pi}^\pi_{i - 1}\|_\infty \|\bm{J}_i\|_\infty \le \gamma^M_i \quad(\forall i \ge t + 1) \ ,
\end{equation}
we have
\begin{equation*}
\begin{aligned}
\prod_{j = t + 1}^{t + k + 1} \bm{L}_j^{\pi'} - \prod_{j = t + 1}^{t + k + 1} \bm{L}_j^\pi &= \big(\prod_{j = t + 1}^{t + k} \bm{L}_j^{\pi'} - \prod_{j = t + 1}^{t + k} \bm{L}_j^\pi\big) \bm{L}_{t + k + 1}^{\pi'} + \big(\prod_{j = t + 1}^{t + k} \bm{L}_j^\pi\big) \big(\bm{L}_{t + k + 1}^{\pi'} - \bm{L}_{t + k + 1}^\pi\big) \\
\Rightarrow \big\| \prod_{j = t + 1}^{t + k + 1} \bm{L}_j^{\pi'} - \prod_{j = t + 1}^{t + k + 1} \bm{L}_j^\pi \big\|_\infty &\le \big(2k \alpha_t(\pi, \pi') \prod_{j = t + 1}^{t + k} \gamma^M_j\big) \gamma^M_{t + k + 1} + \big(\prod_{j = t + 1}^{t + k} \gamma^M_j\big) \cdot 2\alpha_t(\pi, \pi') \gamma^M_{t + k + 1} \\
&= 2(k + 1)\alpha_t(\pi, \pi') \prod_{j = t + 1}^{t + k + 1} \gamma^M_j \ ,
\end{aligned}
\end{equation*}
where the second line applies Inequality \eqref{l-diffupb-nvmdp-mat} and Inequality \eqref{l-inftynormupb-nvmdp}, together with the inductive assumption that \eqref{l-proddiffupb-nvmdp-mat} holds for $k$.
Therefore, Inequality \eqref{l-proddiffupb-nvmdp-mat} is established.

Combine Equation \eqref{prfm-diff-nvmdp-mat} with Inequality \eqref{pi'a-inftynormupb-nvmdp} and Inequality \eqref{l-proddiffupb-nvmdp-mat}:
\begin{equation*}
\|\bm{V}_t^{\pi'} - \bm{V}_t^\pi - \bm{D}^{\pi, \pi'}_t\|_\infty \le \sum_{i = t + 1}^\infty 4A_B (i - t) \alpha^2_t(\pi, \pi') \prod_{j = t + 1}^i \gamma^M_j \ .
\end{equation*}
The theorem is thus proven, and the bound is the same as Proof 1.
$\blacksquare$


\subsection{Performance of model-free methods in Tricky Gridworld}\label{sec:apndx-long-table}

Table \ref{tbl:trickgwd-q-learn-res} summarizes the experimental results of classic Q-learning, NVMDP-Q-learning (Section \ref{sec:q-learn}), and generalized Q-learning variants (Section \ref{sec:geq-learn}) in Tricky Gridworld, which are described in Section \ref{sec:tricky-gwd-mfres}.
Classic Q-learning is tested with dr-0 under both r-lvn and r-svn.
Other algorithms are evaluated with dr-0 to dr-3 under r-lvn and dr-1 to dr-2 under r-svn.
For generalized Q-learning variants, we consider four estimate track sizes: $(n, l) \in \{(6,1), (3,2), (2,3), (1,6)\}$.
In addition, $\lambda = 0.5$ and $\eta = 0.7$ are used in WtAvg-Q.
Each algorithm–configuration pair is run for 100 trials, each lasting 50,000 episodes.
Across the 100 trials, the average number of episodes, training steps, and return standard deviation preceding convergence to optimality are recorded.
The total number of final trajectories passing through cells (3,1) or (4,2) is also measured, where smaller values indicate stronger avoidance effects.

\begin{longtable}{lcccrrrr}
\captionsetup{width=0.8\textwidth, justification=centering}
\caption{Model-free Algorithm Performance in Tricky Gridworld}\label{tbl:trickgwd-q-learn-res} \\
\toprule
\textbf{Algorithm} & \textbf{Parameters} & \textbf{Rwd} & \textbf{Disc} & \textbf{Episodes} & \textbf{Steps(\(\mathbf{10^3}\))} & \textbf{std} & \textbf{Count} \\
\midrule
\endfirsthead

\toprule
\textbf{Algorithm} & \textbf{Parameters} & \textbf{Rwd} & \textbf{Disc} & \textbf{Episodes} & \textbf{Steps(\(\mathbf{10^3}\))} & \textbf{std} & \textbf{Count} \\
\midrule
\endhead

\midrule
\multicolumn{8}{r}{\textit{Continued on next page}} \\
\endfoot

\bottomrule
\endlastfoot

Classic-Q & -- & r-lvn & dr-0 & -- & -- & 39.04 & 71 \\
\hline
NVMDP-Q & -- & r-lvn & dr-0 & 6140 & 1219.65 & 656.99 & 88 \\
\hline
Maxmin-Q & n = 6, l = 1 & r-lvn & dr-0 & 28980 & 3204.62 & 721.24 & 92 \\
\hline
Maxmin-Q & n = 3, l = 2 & r-lvn & dr-0 & 15720 & 2202.43 & 707.26 & 90 \\
\hline
Maxmin-Q & n = 2, l = 3 & r-lvn & dr-0 & 11295 & 1757.79 & 683.62 & 91 \\
\hline
Maxmin-Q & n = 1, l = 6 & r-lvn & dr-0 & 6465 & 1231.58 & 652.13 & 88 \\
\hline
PTMxm-Q & n = 6, l = 1 & r-lvn & dr-0 & 39745 & 4048.48 & 704.30 & 82 \\
\hline
PTMxm-Q & n = 3, l = 2 & r-lvn & dr-0 & 19670 & 2628.08 & 694.34 & 86 \\
\hline
PTMxm-Q & n = 2, l = 3 & r-lvn & dr-0 & 12850 & 1980.48 & 672.00 & 85 \\
\hline
PTMxm-Q & n = 1, l = 6 & r-lvn & dr-0 & 6410 & 1230.64 & 658.31 & 87 \\
\hline
Averaged-Q & n = 6, l = 1 & r-lvn & dr-0 & 37115 & 3940.78 & 694.38 & 87 \\
\hline
Averaged-Q & n = 3, l = 2 & r-lvn & dr-0 & 19275 & 2691.21 & 660.49 & 87 \\
\hline
Averaged-Q & n = 2, l = 3 & r-lvn & dr-0 & 13575 & 2127.47 & 647.92 & 90 \\
\hline
Averaged-Q & n = 1, l = 6 & r-lvn & dr-0 & 7925 & 1483.41 & 627.06 & 83 \\
\hline
WtAvg-Q & n = 6, l = 1 & r-lvn & dr-0 & 37410 & 3974.91 & 691.86 & 90 \\
\hline
WtAvg-Q & n = 3, l = 2 & r-lvn & dr-0 & 19090 & 2651.39 & 668.06 & 91 \\
\hline
WtAvg-Q & n = 2, l = 3 & r-lvn & dr-0 & 12935 & 2041.89 & 654.57 & 87 \\
\hline
WtAvg-Q & n = 1, l = 6 & r-lvn & dr-0 & 6765 & 1313.31 & 613.91 & 88 \\
\hline
NVMDP-Q & -- & r-lvn & dr-1 & 6185 & 1216.53 & 963.11 & 23 \\
\hline
Maxmin-Q & n = 6, l = 1 & r-lvn & dr-1 & 28820 & 3186.08 & 1014.07 & 6 \\
\hline
Maxmin-Q & n = 3, l = 2 & r-lvn & dr-1 & 15670 & 2189.68 & 992.31 & 10 \\
\hline
Maxmin-Q & n = 2, l = 3 & r-lvn & dr-1 & 11105 & 1754.82 & 982.66 & 12 \\
\hline
Maxmin-Q & n = 1, l = 6 & r-lvn & dr-1 & 6460 & 1229.09 & 965.25 & 28 \\
\hline
PTMxm-Q & n = 6, l = 1 & r-lvn & dr-1 & 38590 & 4039.96 & 979.81 & 32 \\
\hline
PTMxm-Q & n = 3, l = 2 & r-lvn & dr-1 & 19125 & 2625.65 & 966.46 & 24 \\
\hline
PTMxm-Q & n = 2, l = 3 & r-lvn & dr-1 & 12640 & 1973.52 & 950.26 & 27 \\
\hline
PTMxm-Q & n = 1, l = 6 & r-lvn & dr-1 & 6430 & 1227.79 & 948.06 & 28 \\
\hline
Averaged-Q & n = 6, l = 1 & r-lvn & dr-1 & 36085 & 3935.52 & 962.83 & 1 \\
\hline
Averaged-Q & n = 3, l = 2 & r-lvn & dr-1 & 19380 & 2679.33 & 961.71 & 6 \\
\hline
Averaged-Q & n = 2, l = 3 & r-lvn & dr-1 & 13670 & 2119.13 & 922.24 & 7 \\
\hline
Averaged-Q & n = 1, l = 6 & r-lvn & dr-1 & 7960 & 1481.08 & 880.52 & 28 \\
\hline
WtAvg-Q & n = 6, l = 1 & r-lvn & dr-1 & 37030 & 3969.53 & 969.75 & 9 \\
\hline
WtAvg-Q & n = 3, l = 2 & r-lvn & dr-1 & 19120 & 2641.39 & 965.06 & 10 \\
\hline
WtAvg-Q & n = 2, l = 3 & r-lvn & dr-1 & 13045 & 2031.90 & 942.64 & 17 \\
\hline
WtAvg-Q & n = 1, l = 6 & r-lvn & dr-1 & 7065 & 1310.81 & 891.54 & 34 \\
\hline
NVMDP-Q & -- & r-lvn & dr-2 & 6495 & 1217.90 & 759.47 & 19 \\
\hline
Maxmin-Q & n = 6, l = 1 & r-lvn & dr-2 & 28835 & 3200.29 & 835.53 & 9 \\
\hline
Maxmin-Q & n = 3, l = 2 & r-lvn & dr-2 & 16280 & 2191.97 & 814.29 & 9 \\
\hline
Maxmin-Q & n = 2, l = 3 & r-lvn & dr-2 & 11050 & 1754.62 & 793.62 & 15 \\
\hline
Maxmin-Q & n = 1, l = 6 & r-lvn & dr-2 & 6435 & 1231.54 & 780.15 & 33 \\
\hline
PTMxm-Q & n = 6, l = 1 & r-lvn & dr-2 & 38375 & 4039.88 & 795.59 & 32 \\
\hline
PTMxm-Q & n = 3, l = 2 & r-lvn & dr-2 & 19170 & 2629.44 & 780.70 & 23 \\
\hline
PTMxm-Q & n = 2, l = 3 & r-lvn & dr-2 & 12605 & 1977.64 & 774.05 & 20 \\
\hline
PTMxm-Q & n = 1, l = 6 & r-lvn & dr-2 & 6465 & 1232.42 & 780.24 & 23 \\
\hline
Averaged-Q & n = 6, l = 1 & r-lvn & dr-2 & 36355 & 3937.35 & 798.63 & 5 \\
\hline
Averaged-Q & n = 3, l = 2 & r-lvn & dr-2 & 18980 & 2681.02 & 765.31 & 4 \\
\hline
Averaged-Q & n = 2, l = 3 & r-lvn & dr-2 & 13405 & 2119.70 & 760.69 & 12 \\
\hline
Averaged-Q & n = 1, l = 6 & r-lvn & dr-2 & 7815 & 1482.84 & 719.79 & 26 \\
\hline
WtAvg-Q & n = 6, l = 1 & r-lvn & dr-2 & 36800 & 3971.04 & 792.53 & 10 \\
\hline
WtAvg-Q & n = 3, l = 2 & r-lvn & dr-2 & 18940 & 2647.73 & 767.21 & 10 \\
\hline
WtAvg-Q & n = 2, l = 3 & r-lvn & dr-2 & 12770 & 2034.30 & 753.78 & 19 \\
\hline
WtAvg-Q & n = 1, l = 6 & r-lvn & dr-2 & 6950 & 1315.25 & 749.27 & 24 \\
\hline
NVMDP-Q & -- & r-lvn & dr-3 & 6195 & 1216.48 & 960.29 & 2 \\
\hline
Maxmin-Q & n = 6, l = 1 & r-lvn & dr-3 & 28415 & 3188.54 & 1035.68 & 0 \\
\hline
Maxmin-Q & n = 3, l = 2 & r-lvn & dr-3 & 15455 & 2194.08 & 1006.39 & 0 \\
\hline
Maxmin-Q & n = 2, l = 3 & r-lvn & dr-3 & 11025 & 1752.47 & 992.40 & 0 \\
\hline
Maxmin-Q & n = 1, l = 6 & r-lvn & dr-3 & 6410 & 1224.13 & 982.18 & 0 \\
\hline
PTMxm-Q & n = 6, l = 1 & r-lvn & dr-3 & 37800 & 4037.09 & 987.09 & 2 \\
\hline
PTMxm-Q & n = 3, l = 2 & r-lvn & dr-3 & 18670 & 2624.74 & 992.54 & 4 \\
\hline
PTMxm-Q & n = 2, l = 3 & r-lvn & dr-3 & 12645 & 1971.59 & 976.28 & 0 \\
\hline
PTMxm-Q & n = 1, l = 6 & r-lvn & dr-3 & 6250 & 1224.15 & 932.99 & 0 \\
\hline
Averaged-Q & n = 6, l = 1 & r-lvn & dr-3 & 35605 & 3926.93 & 983.16 & 0 \\
\hline
Averaged-Q & n = 3, l = 2 & r-lvn & dr-3 & 18970 & 2672.70 & 982.48 & 0 \\
\hline
Averaged-Q & n = 2, l = 3 & r-lvn & dr-3 & 13315 & 2119.75 & 943.07 & 0 \\
\hline
Averaged-Q & n = 1, l = 6 & r-lvn & dr-3 & 7660 & 1480.71 & 885.34 & 0 \\
\hline
WtAvg-Q & n = 6, l = 1 & r-lvn & dr-3 & 36275 & 3962.55 & 991.53 & 0 \\
\hline
WtAvg-Q & n = 3, l = 2 & r-lvn & dr-3 & 18545 & 2638.48 & 965.10 & 0 \\
\hline
WtAvg-Q & n = 2, l = 3 & r-lvn & dr-3 & 12670 & 2028.72 & 953.24 & 0 \\
\hline
WtAvg-Q & n = 1, l = 6 & r-lvn & dr-3 & 6735 & 1310.70 & 924.19 & 0 \\
\hline
Classic-Q & -- & r-svn & dr-0 & -- & -- & 7.73 & 79 \\
\hline
NVMDP-Q & -- & r-svn & dr-1 & 5990 & 1206.98 & 887.86 & 0 \\
\hline
Maxmin-Q & n = 6, l = 1 & r-svn & dr-1 & 26165 & 3107.75 & 992.66 & 0 \\
\hline
Maxmin-Q & n = 3, l = 2 & r-svn & dr-1 & 14345 & 2135.28 & 957.33 & 0 \\
\hline
Maxmin-Q & n = 2, l = 3 & r-svn & dr-1 & 10385 & 1708.70 & 955.39 & 0 \\
\hline
Maxmin-Q & n = 1, l = 6 & r-svn & dr-1 & 6050 & 1207.74 & 925.91 & 0 \\
\hline
PTMxm-Q & n = 6, l = 1 & r-svn & dr-1 & 36685 & 4020.67 & 952.08 & 0 \\
\hline
PTMxm-Q & n = 3, l = 2 & r-svn & dr-1 & 18020 & 2596.51 & 939.34 & 0 \\
\hline
PTMxm-Q & n = 2, l = 3 & r-svn & dr-1 & 12075 & 1949.42 & 937.86 & 0 \\
\hline
PTMxm-Q & n = 1, l = 6 & r-svn & dr-1 & 6025 & 1209.21 & 946.14 & 0 \\
\hline
Averaged-Q & n = 6, l = 1 & r-svn & dr-1 & 35670 & 3926.07 & 962.71 & 0 \\
\hline
Averaged-Q & n = 3, l = 2 & r-svn & dr-1 & 18645 & 2671.52 & 928.77 & 0 \\
\hline
Averaged-Q & n = 2, l = 3 & r-svn & dr-1 & 13210 & 2111.09 & 931.35 & 0 \\
\hline
Averaged-Q & n = 1, l = 6 & r-svn & dr-1 & 7610 & 1471.03 & 865.76 & 0 \\
\hline
WtAvg-Q & n = 6, l = 1 & r-svn & dr-1 & 35885 & 3951.95 & 964.51 & 0 \\
\hline
WtAvg-Q & n = 3, l = 2 & r-svn & dr-1 & 18255 & 2630.20 & 931.80 & 0 \\
\hline
WtAvg-Q & n = 2, l = 3 & r-svn & dr-1 & 12385 & 2022.64 & 915.48 & 0 \\
\hline
WtAvg-Q & n = 1, l = 6 & r-svn & dr-1 & 6565 & 1306.22 & 888.65 & 0 \\
\hline
NVMDP-Q & -- & r-svn & dr-2 & 5920 & 1210.17 & 734.38 & 0 \\
\hline
Maxmin-Q & n = 6, l = 1 & r-svn & dr-2 & 26025 & 3116.96 & 817.16 & 0 \\
\hline
Maxmin-Q & n = 3, l = 2 & r-svn & dr-2 & 14450 & 2135.07 & 793.60 & 0 \\
\hline
Maxmin-Q & n = 2, l = 3 & r-svn & dr-2 & 10285 & 1715.46 & 773.18 & 0 \\
\hline
Maxmin-Q & n = 1, l = 6 & r-svn & dr-2 & 6015 & 1209.27 & 742.52 & 0 \\
\hline
PTMxm-Q & n = 6, l = 1 & r-svn & dr-2 & 37075 & 4022.96 & 779.20 & 0 \\
\hline
PTMxm-Q & n = 3, l = 2 & r-svn & dr-2 & 18255 & 2602.30 & 760.75 & 0 \\
\hline
PTMxm-Q & n = 2, l = 3 & r-svn & dr-2 & 11975 & 1949.77 & 757.76 & 0 \\
\hline
PTMxm-Q & n = 1, l = 6 & r-svn & dr-2 & 5945 & 1208.81 & 706.97 & 0 \\
\hline
Averaged-Q & n = 6, l = 1 & r-svn & dr-2 & 35410 & 3929.59 & 776.08 & 0 \\
\hline
Averaged-Q & n = 3, l = 2 & r-svn & dr-2 & 18680 & 2674.10 & 759.47 & 0 \\
\hline
Averaged-Q & n = 2, l = 3 & r-svn & dr-2 & 13200 & 2115.80 & 734.64 & 0 \\
\hline
Averaged-Q & n = 1, l = 6 & r-svn & dr-2 & 7655 & 1472.18 & 690.32 & 0 \\
\hline
WtAvg-Q & n = 6, l = 1 & r-svn & dr-2 & 36110 & 3953.00 & 791.13 & 0 \\
\hline
WtAvg-Q & n = 3, l = 2 & r-svn & dr-2 & 18410 & 2634.88 & 765.82 & 0 \\
\hline
WtAvg-Q & n = 2, l = 3 & r-svn & dr-2 & 12435 & 2026.25 & 728.84 & 0 \\
\hline
WtAvg-Q & n = 1, l = 6 & r-svn & dr-2 & 6590 & 1305.03 & 724.86 & 0 
\end{longtable}


\end{document}